\documentclass[journal]{IEEEtran}
\usepackage{cite}

\usepackage[pdftex]{graphicx}
\graphicspath{{Figures/}}
\DeclareGraphicsExtensions{.pdf,.jpeg,.png}

\usepackage[cmex10]{amsmath}
\usepackage{amssymb} 
\usepackage{algorithmic}

\usepackage{array}

\ifCLASSOPTIONcompsoc
  \usepackage[caption=false,font=normalsize,labelfont=sf,textfont=sf]{subfig}
\else
  \usepackage[caption=false,font=footnotesize]{subfig}
\fi
\usepackage{url}

\usepackage{graphics} 
\usepackage{amsmath} 
\usepackage{amssymb}  
\usepackage{cite}
\usepackage[ruled,vlined]{algorithm2e}
\usepackage{tikz}
\usetikzlibrary{arrows,shapes,snakes,automata,backgrounds,petri}
\usepackage{textcomp}

\newcommand{\rd}{{\mathrm d}}

\newcommand{\rtr}{{\mathrm{tr}}}

\newcommand{\rT}{{\mathrm{T}}}

\newcommand{\va}{{\bf a}}

\newcommand{\vx}{{\bf x}}

\newcommand{\vp}{{\bf p}}

\newcommand{\vq}{{\bf q}}
\newcommand{\vt}{{\bf t}}

\newcommand{\vf}{{\bf f}}

\newcommand{\vG}{{\bf G}}
\newcommand{\vu}{{\bf u}}
\newcommand{\vv}{{\bf  v}}

\newcommand{\vk}{{\bf k}}

\newcommand{\vw}{{\bf w}}

\newcommand{\vR}{{\bf R}}

\newcommand{\vF}{{\bf F}}

\newcommand{\vB}{{\bf B}}

\newcommand{\argmax}{\operatornamewithlimits{argmax}}
\newcommand{\argmin}{\operatornamewithlimits{argmin}}

\newcommand{\E}{\mathbb{E}}

\newcommand{\Qb}{\mathbb{Q}}
\newcommand{\Pb}{\mathbb{P}}
\newcommand{\Fb}{\mathbb{F}}
\newcommand{\Hb}{\mathbb{H}}
\newcommand{\Rb}{\mathbb{R}}

\newcommand{\KL}[2]{\mathbb{D}_{\mathrm{KL}}\left({#1}\parallel {#2}\right)}

\newcommand{\ExP}[2]{\E_{{#1}}{\left[#2\right]}}

\newcommand{\pluseq}{\mathrel{+}=}

\usepackage{xcolor,colortbl}

\definecolor{Gray}{gray}{0.65}
\newcolumntype{M}{>{\centering\arraybackslash}m{\dimexpr.225\linewidth-2\tabcolsep}}
\newcolumntype{G}{>{\columncolor{Gray}}M}
\newcolumntype{N}{>{\centering\arraybackslash}m{\dimexpr.1\linewidth-2\tabcolsep}}

\begin{document}
%
\title{Information Theoretic Model Predictive Control: Theory and Applications to Autonomous Driving}
%
%

\author{Grady~Williams, Paul~Drews, Brian~Goldfain, 
		James~M.~Rehg, and Evangelos~A.~Theodorou
\thanks{Grady Williams, Brian Goldfain, and James M. Rehg are with the School
of Interactive Computing in the College of Computing at the Georgia Institute of Technology, Atlanta, GA, 30332, USA.
email:~gradyrw@gatech.edu,~bgoldfain3@gatech.edu,~rehg@gatech.edu}
\thanks{Paul Drews is with the School of Electrical and Computer Engineering in the College of Engineering
at the Georgia Institute of Technology.
email:~pdrews3@gatech.edu}
\thanks{Evangelos A. Theodorou is with the Daniel Guggenheim School of Aerospace Engineering in the
College of Engineering at the Georgia Institute of Technology. 
email:~evangelos.theodorou@gatech.edu.}
\thanks{Manuscript received July 5, 2017.}}

%
%

\markboth{Manuscript}%
{Shell \MakeLowercase{\textit{et al.}}: AutoRally T-RO Manuscript}
%



\maketitle

\begin{abstract}
We present an information theoretic approach to stochastic optimal control problems that can be used to derive general sampling based optimization schemes. This new mathematical method is used to develop a sampling based model predictive control algorithm. We apply this information theoretic model predictive control (IT-MPC) scheme to the task of aggressive autonomous driving around a dirt test track, and compare its performance to a model predictive control version of the cross-entropy method.
\end{abstract}


%
\IEEEpeerreviewmaketitle

\section{Introduction} \label{introduction}

Autonomous vehicles have the potential to revolutionize transportation by drastically reducing traffic injuries and fatalities, freeing commute time for more productive activities, and enabling more efficient infrastructure utilization \cite{fagnant2015preparing,thrun2010toward}. A key step in the design of an autonomous vehicle is the control methodology used to convert the vehicle state and world representation into physical actuation \cite{katrakazas2015real}. Existing control methodologies have proven to be effective for many standard vehicle tasks such as lane keeping, turning, and parking. However, there is an important frontier of control at the limits of vehicle performance that has not been fully addressed by prior work. Autonomous racing and the mitigation of risk during collision avoidance are examples of \emph{aggressive driving} domains in which success requires vehicles to operate near their dynamic performance limits.

The control problem for aggressive autonomous driving, and for autonomous driving generally, can naturally be phrased in the language of stochastic optimal control theory. In this framework, a cost function depending on the state and control input is specified, and the goal is to minimize the expected accumulated cost subject to the stochastic dynamical constraints of the vehicle. The advantage of stochastic optimal control over alternative methods is that it directly takes into account the noise characteristics and dynamics of the vehicle during optimization. This is particularly important for control regimes in which the dynamics of the vehicle-terrain interaction play a critical role. Stochastic optimal control therefore combines planning and execution into a single step, providing an elegant theoretical formulation for the control of an autonomous vehicle.

Despite the mathematical appeal of the problem formulation admitted by optimal control theory, it traditionally has not been utilized in the context of autonomous driving. The most popular current methods for controlling autonomous vehicles have their roots in the DARPA Grand and Urban Challenges, where the winners used a hierarchical approach that split the control problem into two sub-problems: path planning and path tracking using a feedback control law \cite{paden2016survey}. In these methods, a path satisfying some driving-related constraints is first planned, and then this path is used as the input to a low-level feedback control law which computes the steering and throttle commands to be used. 
\begin{figure}[t]
\centering
\includegraphics[width=\columnwidth]{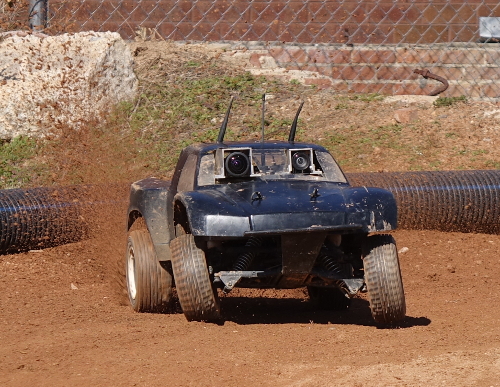}
\caption{Aggressive autonomous driving with information theoretic model predictive control (IT-MPC).}
\label{Fig:powerslide}
\vspace{-4mm}
\end{figure}
While the hierarchical approach makes the control problem tractable, and has many successful applications to autonomous vehicles \cite{thrun2006stanley, urmson2008autonomous, de1998feedback, bacha2008odin,leonard2008perception}, the decomposition into planning and execution phases introduces inherent limitations. In particular, the path planner typically has very coarse knowledge of the underlying system dynamics, usually only utilizing kinematic constraints \cite{dolgov2008practical, fraichard1998path, pepy2006safe,kuwata2009real}. This means that performing maneuvers in aggressive regimes is problematic, since a planned path may not be dynamically feasible \cite{pepy2006path}. Conversely, a path planner may eliminate an aggressive yet feasible trajectory if it is limited to considering paths only in some known safe region. 

Traditionally, the barrier to directly applying optimal control methods to the full autonomous driving problem has been tractability. The state space in autonomous driving is too high dimensional for global methods like solving the Hamilton-Jacobi-Bellman equation to apply, and it involves non-linear dynamics and non-convex objectives which makes applying local methods difficult. There have been a number of methods which analyzed the problem from an optimal control perspective off-line. Examples of this line of research include the work in \cite{velenis2007modeling} where cornering is posed as a minimum time problem and analyzed offline. In \cite{gerdts2009generating} an optimal open loop control sequence is computed  offline, and an LQR controller is used to stabilize the vehicle about the open loop trajectory. Additionally, an approach for performing aggressive sliding maneuvers in order to avoid collisions is developed in \cite{tsiotras2014real}, where optimal trajectories are generated offline for a variety of initial conditions and then a feedback controller is synthesized using Gaussian Process regression. However, given the complexity and sheer number of situations involved in autonomous driving it is clear that the general autonomous driving problem cannot be tackled by generating policies offline. One method which does perform simultaneous planning and tracking online with optimal control is \cite{keivan2013realtime}, where a planner solves a boundary value problem to interpolate between way-points in order to generate a feasible trajectory for a model predictive controller. This method is capable of producing impressive acrobatic maneuvers, however, it relies on a dense series of waypoints to reduce the cost function to a quadratic optimization objective, and introducing additional non-quadratic terms or constraints would be non-trivial. 
%
%
%

In this paper, we develop a new type of control framework based on an information theoretic interpretation of optimal control, and we demonstrate that it is able to overcome the tractability issues associated with the autonomous driving problem. This framework results in a theoretically-sound method for creating sampling based optimization methods, and by utilizing recent advances in computing with graphics processing units (GPUs), we can create a highly parallel sampling algorithm which can operate in a model predictive control (receding horizon) manner in a fast control loop (40 Hz). Additionally, since our method is derivative-free it can handle discontinuous cost functions, which are useful for considering the hard constraints needed for keeping the vehicle on the track, even when operating at high speeds. The contribution of this paper is to develop this new control framework in detail and demonstrate its effectiveness for autonomous vehicle control in aggressive driving regimes. In particular we make the following contributions:
\begin{enumerate}
\item In section \ref{it-control} we derive a highly parallelizable control update law using an information theoretic interpretation of stochastic optimal control, and we show how it can be used to create a flexible model predictive control algorithm.
\item In sections \ref{soc} and \ref{relateds} we provide a detailed discussion of the relationship between the information theoretic approach, classical stochastic optimal control theory, and other stochastic optimization methods popular in robotics.
\item In section \ref{experimental-setup} we report rigorous test results, consisting of over 100 kilometers of autonomous driving data, applying the control algorithm to an aggressive autonomous driving scenario on a 1:5 scale vehicle. We provide experimental comparisons to a baseline sampling-based approach, the cross-entropy method, applied to the same task.  
\end{enumerate}

This paper extends our previous work on aggressive driving with sampling based control \cite{williams2016aggressive, Williams-ICRA-17} both theoretically and technically. In terms of theory, we provide a unified view of sampling-based control by relating our information theoretic framework from~\cite{williams2016aggressive} to stochastic optimal control. In terms of technical contributions, we present a modified version of the algorithm presented in \cite{Williams-ICRA-17} which is more flexible and takes into account practical considerations such as smoothing. Additionally, we present substantially more comprehensive experimental validation of our method, a critical task given the stochastic nature of the algorithms.
In~\cite{williams2016aggressive} we presented results based on 20 laps of driving around our test track, and in \cite{Williams-ICRA-17} this was increased to almost 100. In this article, we present findings from over 1700 total laps, corresponding to over 100 kilometers of driving data. 


\section{Preliminaries} \label{prelims}

While our primary focus is on aggressive autonomous driving,  the model predictive control algorithm that we develop is applicable to many other tasks. We present a general derivation of our control formulation and apply it to the problem of ground vehicle control in section~\ref{experimental-setup}. Consider a general discrete time, continuous state-action dynamical system of the form:
\begin{equation}\label{Equation:Dynamics}
\vx_{t+1} = \vF(\vx_t, \vv_t), 
\end{equation}
where $\vx_t \in \Rb^n$ is the state of the system at time $t$, $\vv_t \in \Rb^m$ is the input to the system at time $t$,
and $\vF$ denotes the, usually non-linear, state-transition function of the system. We will assume that $\vF$ is time-invariant and that we have a finite time-horizon $t \in \{0, 1, 2, \dots T-1 \}$ where the unit of time is determined by the control frequency of the system. The assumption that $\vF$ in \eqref{Equation:Dynamics} is time-invariant is not strictly necessary but covers most cases of interest for model predictive control \cite{mayne2014model} and simplifies our notation. We assume that we do not have direct control over the input variable, $\vv_t$, but rather that $\vv_t$ is a random vector generated by a white noise process with density function:
%
%
\begin{equation}
\vv_t \sim \mathcal{N}(\vu_t, \Sigma), \nonumber
\end{equation}
and that we have direct control over the mean $\vu_t$. This is a reasonable noise assumption for many robotic systems where the commanded input has to pass through a lower level of control before reaching the actual system. In this work, for instance, the controller outputs the steering and throttle inputs for a fifth-scale vehicle, and these in turn are used as set-point targets for servomotor controllers. In this case, our assumption translates to the low level controller achieving the set-point with some error that satisfies a Gaussian distribution. This is much more reasonable than assuming that the low level controller perfectly hits its target every time. Additional artificial sources of noise could be inserted into the system in order to foster exploration. Next, suppose that we are given a sequence of inputs:
\begin{equation}
\left( \vv_0, \vv_1, \dots \vv_{T-1} \right) = V \in \Rb^{m \times T},  \nonumber
\end{equation} 
and a corresponding sequence of mean input variables:
\begin{equation}
\left( \vu_0, \vu_1, \dots \vu_{T-1} \right) = U \in \Rb^{m \times T}. \nonumber
\end{equation}
We can then define the probability density functions for $V$ as:
\begin{align}
&q(V|U, \Sigma) = \prod_{t=0}^{T-1} Z^{-1}\exp\left( -\frac{1}{2}(\vv_t - \vu_t)^\rT \Sigma^{-1} (\vv_t - \vu_t)\right) \nonumber \\
\label{Equation:TrajectoryPDF}
&= Z^{-T} \exp \left( -\frac{1}{2} \sum_{t=0}^{T-1} (\vv_t - \vu_t)^\rT \Sigma^{-1} (\vv_t - \vu_t)\right),   
%
%
\end{align}
where $Z = \left( (2\pi)^m |\Sigma| \right)^{\frac{1}{2}}$. Throughout this text we will denote probability density functions with a lowercase letter, and the probability distribution (measure) corresponding to the density will be denoted by the same letter in uppercase blackboard boldface. So the density $q(V|U,\Sigma)$ corresponds to the distribution $\Qb_{U,\Sigma}$.
%
%

Given a running cost function, $\mathcal{L}(\vx_t, \vu_t)$, and a terminal cost $\phi(\vx_{T})$, we can define the discrete time optimal control problem as:
\begin{equation}
\label{Equation:OptimalControl}
U^* = \argmin_{U \in \mathcal{U}} \ExP{\Qb_{U,\Sigma}}{\phi(\vx_{T}) + \sum_{t=0}^{T-1}\mathcal{L}(\vx_t, \vu_t)}. 
\end{equation}
Where $\mathcal{U}$ is the set of admissible command sequences. We assume that the running cost can be split into an arbitrary state-dependent running cost, and a control cost which is a quadratic function of the system noise:
%
%
\begin{equation}\nonumber
\mathcal{L}(\vx_t, \vu_t) = c(\vx_t) + \frac{\lambda}{2} \left( \vu_t^\rT \Sigma^{-1} \vu_t + \beta_t^\rT \vu_t \right).
\end{equation}
The affine term $\beta$ allows for the location of the minimum control cost to be moved away from zero (although $\beta = 0$ is the standard case). Next denote $C(\vx_0, \vx_1, \dots \vx_T)$ as the portion of the cost of a trajectory that only depends on the state:
%
%
\begin{equation} \nonumber
C(\vx_0, \vx_, \dots \vx_T) = \phi(\vx_T) + \sum_{t=0}^{T-1} c(\vx_t).
\end{equation}
In the following, it will be necessary to refer to the state-cost of an input sequence $V$, along with an initial condition. For this we define the operator $\mathcal{H}$ which transforms an input sequence (along with an initial condition) into a resulting trajectory:
\begin{equation} \nonumber
\mathcal{H}(V; \vx_0) = \left( \vx_0, ~ \vF(\vx_0, \vv_0), ~ \vF(\vF(\vx_0, \vv_0), \vv_1), \dots \right).
\end{equation}
Then the state-cost of an input sequence is defined as the functional composition:
\begin{equation}
\label{Equation:State-Cost}
S(V; \vx_0) = C( \mathcal{H}(V; \vx_0) ).
\end{equation}
For notational compactness we will drop the dependence on the initial condition and simply refer to this as $S(V)$, unless it is ambiguous as to what the initial condition is. Lastly we will need to define two quantities from information theory that are required for our derivation. First we define the \emph{Free-Energy} of a control system as:
\begin{equation}
\label{Equation:FreeEnergy}
\mathcal{F}\left(S, p, \vx_0, \lambda \right) = \log \left( \ExP{\Pb}{\exp \left( -\frac{1}{\lambda} S(V) \right)} \right).
\end{equation}
Here $\lambda \in \Rb^+$ is called the inverse temperature, $\Pb$ is some probability density over input sequences which we will refer to as the base probability, and $ p $ is the corresponding density. The base distribution is roughly analogous to a bayesian prior, usually it is defined as the uncontrolled dynamics of the system (i.e. $\Pb = \Qb_{{\bf{0}}, \Sigma)}$) but this need not always be the case. Next let $\Fb$ and $\Hb$ be two probability distributions that are absolutely continuous\footnote{Absolute continuity between $\vq$ and $\vp$ means that if one density is zero so is the other (i.e $(\vp(V) = 0) \leftrightarrow (\vq(V) = 0$) ).} with each other. Then the KL-Divergence between $\Fb$ and $\Hb$ is:

\begin{equation}
\label{Equation:KL}
\KL{\Fb}{\Hb} = \ExP{\Fb}{\log \left(  \frac{f(V)}{h(V)} \right)}. 
\end{equation}
The KL-Divergence provides a method for comparing distances\footnote{Although the KL-Divergence is not technically a distance metric as it is not symmetric.} between probability distributions and is therefore useful for defining optimization objectives.

\section{Information-Theoretic Model Predictive Control} \label{it-control}

In this section we show how the definition of the free energy from \eqref{Equation:FreeEnergy} can be used to derive a lower
bound for the optimal control problem that we defined in \eqref{Equation:OptimalControl}. This lower bound is subsequently used to create a sampling based model predictive control algorithm. 

Consider a base distribution $\Pb$ and a distribution induced by an open-loop control sequence: $\Qb_{U,\Sigma}$, and suppose that these distributions are absolutely continuous with each other. We start by making the following observation:
\begin{align}
&\mathcal{F}\left(S, p, \vx_0, \lambda \right) = \log \left( \ExP{\Pb}{\exp \left( -\frac{1}{\lambda} S(V) \right)} \right) \nonumber \\
&= \log \left( \ExP{\Qb_{U,\Sigma}}{\exp \left( -\frac{1}{\lambda} S(V) \right) \frac{p(V)}{q(V|U,\Sigma)}} \right), \nonumber
\end{align}
where the last equality follows from switching the expectation by using the standard importance sampling trick of multiplying by $1 = \frac{q(V|U,\Sigma)}{q(V|U, \Sigma)}$. Using the concavity of the logarithm, we can apply Jensen's inequality and obtain:
\begin{equation}
\label{Equation:Jensens}
\mathcal{F}\left(S, p, \vx_0, \lambda \right) \ge \ExP{Q_{U,\Sigma}}{\log\left( \exp \left(-\frac{1}{\lambda} S(V) \right) \frac{p(V)}{q(V|U,\Sigma)} \right)}.
\end{equation}
The right-hand side of this inequality can be simplified, using basic properties of the logarithm and the definition of the KL-Divergence, as:
\begin{align}
&= -\frac{1}{\lambda}\ExP{\Qb_{U,\Sigma}}{S(V) + \lambda \log \left( \frac{q(V|U,\Sigma)}{p(V)} \right)} \nonumber \\
\label{Equation:Inequality}
&= -\frac{1}{\lambda}\Big( \ExP{\Qb_{U,\Sigma}}{S(V)} + \lambda \KL{\Qb_{U,\Sigma}}{\Pb} \Big).
\end{align}
Substituting \eqref{Equation:Inequality} back into \eqref{Equation:Jensens}, and then multiplying each side by $-\lambda$ results in the following free energy lower bound:
\begin{equation}
\label{Equation:Free-Energy-Bound}
-\lambda \mathcal{F}\left(S, p, \vx_0, \lambda \right) \le \ExP{\Qb_{U,\Sigma}}{S(V)} + \lambda \KL{\Qb_{U,\Sigma}}{\Pb}.
\end{equation}
On the left side of this equation we have the negative inverse temperature multiplying the free energy of the system, and on the right side the state-cost for an optimal control problem followed by the KL-Divergence between the base and controlled distribution. The KL-Divergence measures the difference between two probability distributions, so it intuitively acts as a type of control cost by penalizing deviations of the controlled distribution from the base distribution. More concretely, suppose we assign the base distribution as:
\begin{align}
\label{Equation:BaseMeasure}
p(V) = q(V|\tilde{U}, \Sigma),
\end{align}
where $\tilde{U}$ represents some nominal control input applied to the system. Then the KL-Divergence between $\Qb_{U,\Sigma}$ and $\Qb_{\tilde{U},\Sigma}$ is 
\begin{align}
&\KL{\Qb_{U,\Sigma}}{\Qb_{\tilde{U},\Sigma}}= \frac{1}{2}\sum_{t=0}^{T-1} \left(\vu_t - \tilde{\vu}_t\right)^\rT \Sigma^{-1} \left(\vu_t - \tilde{\vu}_t \right)\nonumber \\
&= \frac{1}{2}\sum_{t=0}^{T-1} \left( \vu_t \Sigma^{-1} \vu_t - \tilde{\vu}_t^\rT \Sigma^{-1}\vu_t + \tilde{\vu}_t\Sigma^{-1}\tilde{\vu}_t \right), \nonumber
\end{align}
which  if we set $\beta^\rT = -\tilde{\vu}^\rT \Sigma^{-1}$ and $c = \tilde{\vu}^\rT \Sigma^{-1} \tilde{\vu}$ we get:
\begin{equation}
\label{Equation:Controls-Cost}
\KL{\Qb_{U,\Sigma}}{\Qb_{\tilde{U},\Sigma}} = \frac{1}{2}\sum_{t=0}^{T-1} \left( \vu_t^\rT \Sigma^{-1} \vu_t + \beta_t^\rT \vu_t + c_t \right).
\end{equation}
Which is the type of quadratic control cost that we are interested in minimizing.\footnote{The constant term $c$ is irrelevant for the purpose of optimizing with respect to $U$} Usually $\tilde{U} = {\bf{0}}$, so that the base probability distribution corresponds to the distribution induced by the uncontrolled system dynamics. In that case $\beta = {\bf{0}}$ and $c = 0$.

Substituting (\ref{Equation:Controls-Cost}) and (\ref{Equation:State-Cost}) into the RHS of (\ref{Equation:Free-Energy-Bound}) and expanding we obtain:
\begin{equation}
\label{Equation:Combined-Cost}
\ExP{\Qb_{U,\Sigma}}{\phi(\vx_{T}) + \sum_{t=0}^{T-1} c(\vx_t) + \frac{\lambda}{2}\left(\vu_t + \beta_t^\rT \vu_t + c_t \right)}.
\end{equation}
This is clearly equivalent to the cost function in \eqref{Equation:OptimalControl}, allowing us to conclude:
\begin{equation}
\label{Equation:Objective-Bound}
-\lambda \mathcal{F}\left(S, p, \vx_0, \lambda \right) \le 
\ExP{\Qb_{U,\Sigma}}{\phi(\vx_{T}) + \sum_{t=0}^{T-1}\mathcal{L}(\vx_t, \vu_t)}.
\end{equation}
We have thus established that the negative free energy provides a lower bound on the standard optimal control objective. Note that \eqref{Equation:Objective-Bound} and \eqref{Equation:Free-Energy-Bound} are in fact a family of lower bounds indexed by the choice of nominal trajectory $\tilde{U}$ in defining the base measure in \eqref{Equation:BaseMeasure}. The choice $\tilde{U}=0$ uses the uncontrolled dynamics as the base measure and corresponds to a control cost with $\beta = 0$.  

\subsection{Optimal Distribution}

In the previous subsection, we demonstrated that the free energy of the system provides a lower bound on the cost of an optimal control problem. We now establish a further equivalence between optimizing the control objective by selecting a control trajectory, and achieving the lower bound in (\ref{Equation:Objective-Bound}) by choosing an optimal distribution for the controls. 
Define the optimal control density function $\Qb^*$ as follows:
\begin{align}
q^*(V) &= \frac{1}{\eta}\exp\left(-\frac{1}{\lambda}S(V) \right) p(V)  \label{Equation:Optimal-Density} \\
\eta &= \int_{\Rb^{m \times T}} \exp\left(-\frac{1}{\lambda}S(V) \right) p(V) \rd V. \label{Equation:eta}
\end{align} 
We will now show that this particular choice of distribution achieves the lower bound. Substituting $\Qb^*$ into the KL-Divergence term from the RHS of \eqref{Equation:Free-Energy-Bound} yields
\begin{align}
\KL{\Qb^*}{\Pb} &= \ExP{\Qb^*}{ \log \left(\frac{\frac{1}{\eta}\exp\left(-\frac{1}{\lambda}S(V) \right)p(V)}{p(V)} \right) \rd V}  \nonumber \\
&= -\frac{1}{\lambda}\ExP{\Qb^*}{S(V)} - \log \left( \eta \right).  \nonumber 
\end{align}
Substituting this divergence into \eqref{Equation:Free-Energy-Bound} results in:
\begin{equation}
-\lambda \mathcal{F} \le \ExP{\Qb^*}{S(V)} - \ExP{\Qb^*}{S(V)} - \lambda \log \left( \eta \right).  \nonumber
\end{equation}
Simplifying the RHS and substituting \eqref{Equation:eta} we obtain
\begin{equation}
-\lambda \mathcal{F} \le -\lambda \log \left( \ExP{\Pb}{\exp\left(-\frac{1}{\lambda}S(V) \right)} \right). \nonumber
\end{equation}
Since the RHS is precisely the definition of $-\lambda \mathcal{F}\left(S, p, \vx_0, \lambda \right)$, the inequality reduces to an equality and we have established the optimality of $q^*(V)$. Note that the key to the construction of the optimal distribution in \eqref{Equation:Optimal-Density} is the augmentation of the base measure with the cost of the state trajectory. As a consequence, control inputs drawn from the optimal distribution achieve a lower cost, in expectation, than any other control distribution.

\begin{figure}
\includegraphics[width=.95\columnwidth]{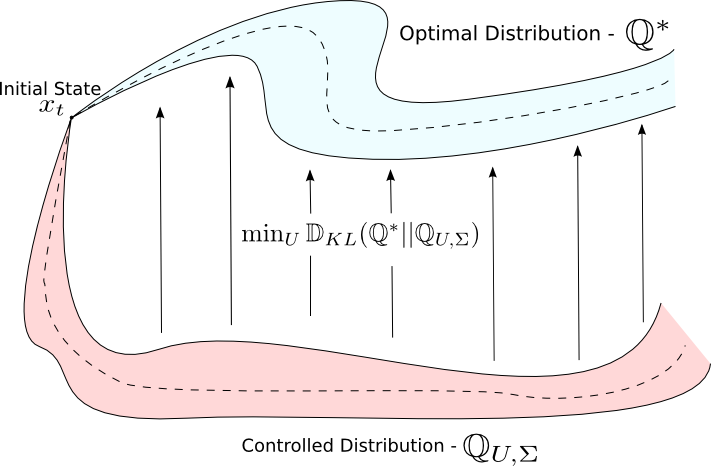}
\caption{Visualization of the information theoretic control objective of ``pushing" the controlled distribution close to the optimal one.}
\label{Fig:ObjectivePushing}
\end{figure}

We have demonstrated an equivalence between optimizing a control trajectory and sampling from an optimal control distribution. We can exploit this equivalence to develop a novel scheme for optimal control: instead of directly minimizing \eqref{Equation:OptimalControl}, we can ``push" the controlled distribution $\Qb_{U,\Sigma}$ as close as possible to the optimal distribution $\Qb^*$ (see Fig. \ref{Fig:ObjectivePushing}). If $\Qb_{U,\Sigma}$ is aligned with $\Qb^*$, then sampling from $\Qb_{U,\Sigma}$ by applying the resulting control input will result in low cost trajectories. 

\subsection{KL-Divergence Minimization}

The goal of aligning the controlled distribution $\Qb_{U,\Sigma}$ with the optimal distribution $\Qb^*$ can be achieved by minimizing the KL-Divergence:
\begin{equation}
\label{Equation:KL_Objective}
U^* = \argmin_{U \in \mathcal{U}}\left[\KL{\Qb^*}{\Qb_{U,\Sigma}}\right].
\end{equation}
Expanding the objective we obtain:
\begin{align}
&U^* = \argmin_{U \in \mathcal{U}}\left[\ExP{\Qb^*}{\log\left(\frac{q^*(V)}{q(V|U, \Sigma)} \right) } \right] \nonumber \\
\label{Equation:Full_KL}
&= \argmin_{U \in \mathcal{U}}\Big[ \ExP{\Qb^*}{\log(q^*(V))} - \ExP{\Qb^*}{\log\left(q(V|U, \Sigma) \right)} \Big]  \\
\label{Equation:Simplified_KL}
&= \argmax_{U \in \mathcal{U}}\Big[\ExP{\Qb^*}{\log\left( q(V|U, \Sigma) \right)} \Big].
\end{align}
The step from \eqref{Equation:Full_KL} to \eqref{Equation:Simplified_KL} follows because the optimal distribution is invariant to the particular control input that we apply to the system. Substituting the definition of $q(V|U, \Sigma)$ from \eqref{Equation:TrajectoryPDF} into the objective from \eqref{Equation:Simplified_KL} yields
\begin{equation}
\mathbb{E}_{\Qb^*}\Bigg[ \log \left( Z^{-T} \right)  
-\frac{1}{2}\sum_{t=0}^{T-1}\left(\vv_t - \vu_t\right)^\rT \Sigma^{-1}\left(\vv_t - \vu_t \right) \Bigg]. \nonumber
\end{equation}
Removing the constant, we obtain the following quadratic minimization problem:
\begin{equation}
\label{Equation:QuadraticMin}
U^* = \argmin_{U \in \mathcal{U}}\left( \ExP{\Qb^*}{\sum_{t=0}^{T-1} \left(\vv_t - \vu_t\right)^\rT \Sigma^{-1}\left(\vv_t - \vu_t \right)} \right).
\end{equation}
In the unconstrained case ($\mathcal{U} = \mathbb{R}^m$), we can solve for $\vu_t$ to yield the optimal solution:
\begin{equation}
\label{Equation:QuadraticOptimization}
\vu_t^* = \ExP{\Qb^*}{\vv_t} ~ ~ \forall t \in \{0, 1, \dots T-1 \}.
\end{equation}
In the case of a general $\mathcal{U}$, the solution of \eqref{Equation:QuadraticMin} requires the solution of a quadratic program. In section \ref{sec:control-constraints} we demonstrate how to convert a problem with control constraints into an unconstrained one, and therefore we do not further consider this scenario. 

We see that the optimal open-loop control sequence is the expected value of control trajectories sampled from the optimal distribution. This expression is not useful by itself, since we have no method for directly sampling from the optimal distribution. However, we will demonstrate in section \ref{sec:Importance-Sampling} that \eqref{Equation:QuadraticOptimization} can be used to develop an approximate iterative method for computing $U^*$.

It is worth noting that, based on the asymmetry of the KL Divergence, an alternative formulation of (\ref{Equation:KL_Objective}) is to minimize $\KL{\Qb_{U,\Sigma}}{\Qb^*}$. It can be shown that this results in a non-convex optimization problem which can be used to obtain a gradient equation\footnote{The resulting gradient equation is equivalent to the well known policy gradient theorem from reinforcement learning.}. However, since we are interested in real-time model predictive control, the convex optimization problem in (\ref{Equation:QuadraticMin}) is preferable, since gradient step-sizes are difficult to tune in a real-time control framework.

\subsection{Importance Sampling}
\label{sec:Importance-Sampling}

We can use the technique of importance sampling~\cite{doucet2001} to construct a set of samples that provide an unbiased estimate of the optimal control solution given a current control distribution. Given an initial estimate of the controls, denoted by $\hat{U}$, we have:
%
%
\begin{align}
\ExP{\Qb^*}{\vv_t} &= \int q^*(V) \vv_t \rd V \nonumber \\
&= \int \underbrace{\frac{q^*(V)}{q\left( V|\hat{U},\Sigma \right)}}_{w(V)} q\left(V|\hat{U},\Sigma\right) \vv_t \rd V \nonumber
\end{align}
This integral expression can be expressed as the following expectation:
\begin{equation}
\ExP{\Qb_{U,\Sigma}}{w(V) \vv_t}, ~~ w(V) = \frac{q^*(V)}{q\left(V|\hat{U},\Sigma\right)}. \nonumber
\end{equation}
The weighting term, $w(V)$, 
%
%
is the importance sampling weight which allows us to compute expectations with respect to $\Qb^*$ by sampling trajectories from $\Qb_{\hat{U},\Sigma}$. This weighting term can be split into a two terms: one depending on the state cost of a trajectory, and the other the control cost. This is done by using the base distribution $p(V)$ as follows:
\begin{align}
w(V) &= \left( \frac{q^*(V)}{p(V)} \right) \left( \frac{p(V)}{q\left(V|\hat{U},\Sigma\right)} \right), \nonumber \\ 
     &= \frac{1}{\eta}\exp\left( -\frac{1}{\lambda}S(V) \right) \left( \frac{p(V)}{q\left(V|\hat{U},\Sigma\right)} \right),
\end{align}
In the case that the base distribution takes the form as in \eqref{Equation:BaseMeasure}, we have the following:
\begin{align}
\frac{p(V)}{q\left(V|\hat{U},\Sigma\right)} &= \exp\left(-\frac{1}{2} \sum_{t=0}^{T-1} \mathcal{D} + 2(\hat{\vu}_t - \tilde{\vu}_t)^\rT \Sigma^{-1} \vv_t \right), \nonumber \\
\mathcal{D} &= \tilde{\vu}_t^\rT \Sigma^{-1} \tilde{\vu}_t-\hat{\vu}_t^\rT \Sigma^{-1} \hat{\vu}_t , \nonumber
\end{align}
Notice that the term $\mathcal{D}$ does not depend on $\vv_t$, so its possible to factor it outside of the integral and cancel it with the corresponding term appearing in the importance sampling estimate of $\eta$. This leaves us with the weight:
\begin{equation}
\label{Equation:Control-Importance-Weight}
\frac{p(V)}{q\left(V|\hat{U},\Sigma\right)} \propto \exp\left(-\sum_{t=0}^{T-1} (\hat{\vu}_t - \tilde{\vu}_t)^\rT \Sigma^{-1} \vv_t \right). \nonumber
\end{equation}
This term encourages samples to move in the direction of the base distribution, $\tilde{\vu}_t - \hat{\vu}_t$. We then have:
\begin{align}
\label{Equation:IT-Controls}
w(V) &= \frac{1}{\eta}\exp\left( -\frac{1}{\lambda}\left(S(V) + \lambda\sum_{t=0}^{T-1}(\hat{\vu}_t - \tilde{\vu}_t)^\rT \Sigma^{-1} \vv_t  \right) \right), \nonumber \\
\vu_t &= \ExP{\Qb_{\hat{U}, \Sigma}}{w(V) \vv_t}. 
\end{align}
Equation \eqref{Equation:IT-Controls} describes the \emph{optimal information theoretic control law} for a given base distribution. Note that the control obtained from \eqref{Equation:IT-Controls} is globally optimal (in an information theoretic sense) under the condition that the expectation can be perfectly evaluated. In practice, it must be estimated using a Monte-Carlo approximation, which can create the appearance of ``local optimums'' due to insufficient sampling of the state space.

\subsection{Practical Issues}

Equation \eqref{Equation:IT-Controls} forms the basis for our sampling based control methodology. However, there are a few practical issues to address before describing the full information theoretic model predictive control algorithm. These are (1) Shifting the range of the trajectory costs, (2) Decoupling the control cost and temperature, (3) Handling control constraints, (4) Smoothing the outputted solution, and (5) Sampling trajectories fast enough for online optimization. In this subsection we explain effective solutions to these problems which keep the theoretical basis for the algorithm intact.

\subsubsection{Shifting the range of the trajectory costs}

The negative exponentiation required by the importance sampling weight is numerically sensitive to the range of the input values. If the costs are too high then the negative exponentiation results in values numerically equal to zero, and if the costs are not bounded from below then the negative exponentiation can lead to overflow errors. For this reason we shift the range of the costs so that the best trajectory sampled has a value of $0$. This simultaneously bounds the costs from below and ensures that at least one trajectory has an importance sampling weight which is not numerically zero. This is done as follows: first expand out the normalizing term $\eta$ in \eqref{Equation:IT-Controls} so that the importance sampling weight is: 
\begin{equation}
= \frac{\exp\left( -\frac{1}{\lambda}\left(S(V) + \lambda\sum_{t=0}^{T-1}(\hat{\vu}_t - \tilde{\vu}_t)^\rT \Sigma^{-1} \vv_t  \right) \right)}{
\int \exp\left( -\frac{1}{\lambda}\left(S(V) + \lambda\sum_{t=0}^{T-1}(\hat{\vu}_t - \tilde{\vu}_t)^\rT \Sigma^{-1} \vv_t  \right) \right) \rd V
}. \nonumber
\end{equation}
Now define $\rho$ as the minimum cost (in the Monte-Carlo approximation it is the minimum \emph{sampled} cost). We then multiply by:
\begin{equation}
1 = \frac{\exp\left(\frac{1}{\lambda}\rho \right)}{\exp\left(\frac{1}{\lambda}\rho\right)}. \nonumber
\end{equation}
which results in:
\begin{align*}
&\resizebox{.98 \columnwidth}{!}{$
w(V) = \frac{1}{\tilde{\eta}}\exp\left( -\frac{1}{\lambda}\left(S(V) + \lambda\sum_{t=0}^{T-1}(\hat{\vu}_t - \tilde{\vu}_t)^\rT \Sigma^{-1} \vv_t  - \rho \right) \right)$}, \\
&\resizebox{.98 \columnwidth}{!}{$
\tilde{\eta} =  \int \exp\left( -\frac{1}{\lambda}\left(S(V) + \lambda\sum_{t=0}^{T-1}(\hat{\vu}_t - \tilde{\vu}_t)^\rT \Sigma^{-1} \vv_t  - \rho \right) \right) $}.
\end{align*}
which bounds the cost from below by 0. Since we have only multiplied by $1$, this procedure does not change the optimality of the approach in any way, besides better conditioning it numerically.

\subsubsection{Decoupling control cost and temperature}

Consider the form of the importance sampling weight from \eqref{Equation:IT-Controls} when we take the uncontrolled dynamics of the system as the base distribution:\footnote{This is a natural choice from an optimization perspective, since the minimum control cost is achieved with $U \equiv {\bf{0}}$.}
\begin{equation}
w(V) = \frac{1}{\tilde{\eta}} \exp\left( -\frac{1}{\lambda} \left( S(V) + \lambda \sum_{t=0}^{T-1} \hat{\vu}_t^\rT \Sigma^{-1} \vv_t \right) \right). \nonumber
\end{equation}
The challenge with this formulation is that changing the inverse temperature $\lambda$, also changes the relative control cost and vice versa. The inverse temperature determines how tightly peaked the optimal distribution is, as $\lambda \rightarrow 0$, the optimal distribution places all of its mass on a single trajectory, whereas as $\lambda \rightarrow \infty$ all points in the state space have equal weight. Figure \ref{Fig:Lambda} shows the probability weights corresponding to trajectory costs for varying values of $\lambda$. 

In \eqref{Equation:IT-Controls} lowering the inverse temperature also lowers the control cost, this is sensible from a theoretical point of view: if we are allowed more control authority over the system, then we should be able to more tightly maintain a given trajectory. Unfortunately, raising the temperature too high results in numerical instability since most trajectories are rejected (have weight numerically equal to zero), at which point the importance sampling oscillates between solutions instead of converging. Our solution is to change the base distribution which defines the control cost. Let $\hat{U}$ be the current planned control sequence, and define the new base distribution as:

\begin{figure}
\includegraphics[width=.95\columnwidth, trim = 0mm 0mm 0mm 0mm, clip=true]{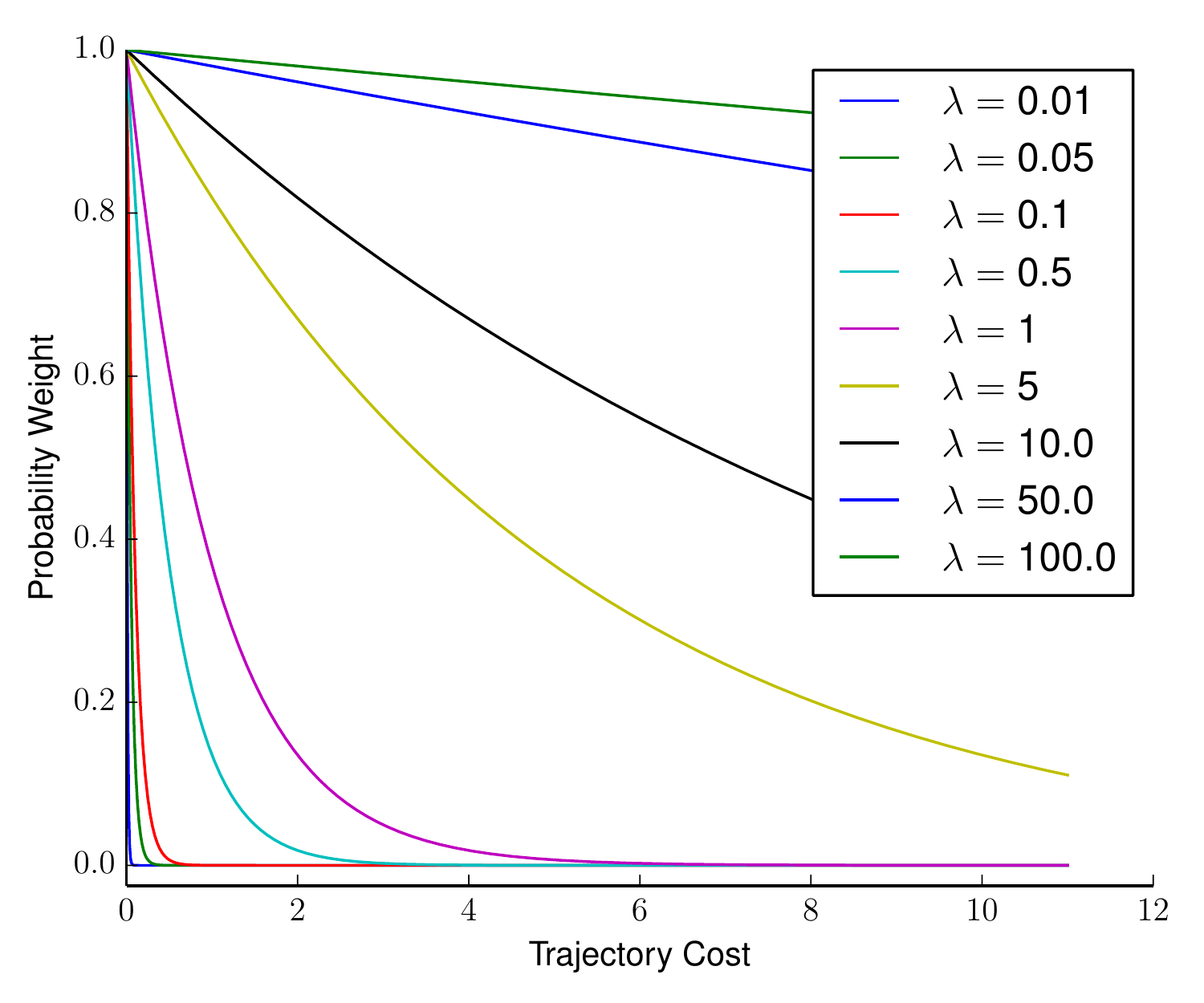}
\caption{Effect of changing $\lambda$ on the probability weight corresponding to a trajectory cost. Low values of $\lambda$ result in many trajectories being rejected, high values of $\lambda$ take close to an un-weighted average.}
\label{Fig:Lambda}
\vspace{-3mm}
\end{figure}

\begin{equation}
\tilde{p}(V) = p(V|\alpha\hat{U}, \Sigma), \nonumber
\end{equation}
where $0 < \alpha < 1$. With $\alpha=0$, the base distribution reverts back to the uncontrolled dynamics and pushes $U$ to zero. And with $\alpha = 1$, the base distribution is the distribution corresponding to the current planned control law, which keeps $U$ near the distribution corresponding to $\hat{U}$. The case of $\alpha=1$ can be interpreted as placing a cost on how much the new open loop control law is allowed to deviate from the previous one, which is useful for creating smooth motions. A value in-between zero and one balances the two requirements of low energy and smoothness.

The construction of the optimal distribution and the corresponding control law are the same under this new base distribution. However, the control cost portion of the importance sampling weight now becomes:
\begin{equation}
\sum_{t=0}^{T-1} (1 - \alpha)\hat{\vu}^\rT \Sigma^{-1} \vv_t.\nonumber
\end{equation}
We then have $\gamma = \lambda(1 - \alpha)$ as the new control cost parameter, resulting in
\begin{equation}
w(V) = \frac{1}{\tilde{\eta}}\exp\left(-\frac{1}{\lambda}\left(S(V) + \gamma \sum_{t=0}^{T-1} \hat{\vu}_t^\rT \Sigma^{-1} \vv_t \right)\right),\nonumber
\end{equation}
as the final probability weighting for the algorithm.

\subsubsection{Handling Control Constraints} \label{sec:control-constraints}

Most interesting control systems, including the autonomous vehicle we consider here, have actuator limits that the controller must take into account. There are several ways to do this in our information theoretic framework: rejection sampling for trajectories that violate the control constraints, or by formulating the optimization problem in \eqref{Equation:QuadraticMin} as a quadratic program and solving the constrained optimization problem. Although both of these methods could work in theory, they both have significant drawbacks in practice. The first approach completely removes trajectory samples from the optimization, and the second one requires solving a quadratic program online. A simpler solution is to make the problem unconstrained by pushing the control constraints into the system dynamics: 
%
%
\begin{equation}
\vx_{t+1} = \vF(\vx_t, g(\vv_t)), \nonumber
\end{equation}
where $g(\vv_t)$ is a clamping function that restricts $\vv_t$ to remain within an allowable input region. Since the sampling based update law does not require computing gradients or linearizing the dynamics, adding this additional non-linearity (and non-smooth) component into the dynamics is trivial to implement, and it works well in practice. 

\subsubsection{Control Smoothing}

The stochastic nature of the sampling procedure can lead to significant chattering in the resulting control, which can be removed by smoothing the output control sequence. One very effective method for smoothing is by fitting local polynomial approximations to the control sequence. Consider the quadratic objective \eqref{Equation:QuadraticMin}:
\begin{equation}
\label{Equation:StandardOpt}
\vu_t^* = \argmin\left( \ExP{\Qb^*}{\left(\vv_t - \vu_t\right)^\rT \Sigma^{-1}\left(\vv_t - \vu_t \right)} \right).
\end{equation}
And now consider fitting a local polynomial approximation (at every timestep) so that $\vu_t = \va_0 + \va_1t + \va_2t^2 + \dots \va_k t^k = A \vt$, where $A = (\va_0, \va_1, \dots \va_k)$ and $\vt = (1, t, t^2, \dots t^k)$. Our goal is to then find the optimal set of coefficients at each timestep. The optimal coefficients, at timestep $j$, can be found through the following optimization:
\begin{equation}
A_j^* = \argmin\left( \ExP{\Qb^*}{\sum_{t = (j-k)}^{j+k} \left(\vv_t - A\vt \right)^\rT \Sigma^{-1}\left(\vv_t - A \vt \right)} \right).\nonumber \\
\end{equation}
Note how the optimization now spans multiple timesteps into the past and future in order to compute a smoother control input. This optimization problem is equivalent to optimizing the objective:
\begin{align}
&\ExP{\Qb^*}{\sum_{t = (j-k)}^{j+k} \vv_t^\rT \Sigma^{-1} A \vt + \vt^\rT A^\rT \Sigma^{-1} A \vt} \nonumber \\
&= \sum_{t = (j-k)}^{j+k} \ExP{\Qb^*}{\vv_t}^\rT \Sigma^{-1}  A \vt + \vt^\rT A^\rT \Sigma^{-1} A \vt. \nonumber
\end{align}
This in turn is equivalent to the minimization:
\begin{align}
\label{Equation:Smoothing}
\small{A_t = \argmin\left( \sum_{t = j-k}^{j+k} \left(\ExP{\Qb^*}{\vv_t} - A\vt \right)^\rT \Sigma^{-1}\left(\ExP{\vq^*}{\vv_t} - A \vt \right) \right)} \nonumber
\end{align}
This is a convenient expression because it means that we can first compute the weighted average over trajectories, and then perform a local polynomial approximation in order to smooth the resulting control sequence. In other words, we do not have to handle any polynomial expressions inside the expectation. The naive method for computing the controls is to then compute $\ExP{\Qb^*}{\vv_t}$ using a Monte-Carlo approximation, solve for each $A_t$, and lastly compute the smoothed control inputs $\vu_t$. However, a simpler method which achieves the same result is to use a Savitsky-Galoy filter \cite{savitzky1964smoothing} which implements local polynomial smoothing using a specific set of convolution coefficients. Using a Savitsky-Galoy filter, we simply compute $U' = \ExP{\Qb^*}{V}$ and then compute the smoothed control sequence, $U$, by passing $U'$ through the convolutional filter.

\subsubsection{GPU-Based Trajectory Sampling}

The key requirement for applying our information theoretic framework in a model predictive control setting is the ability to generate and evaluate a large number of samples in real time. As in our prior sampling-based MPC methods \cite{williams2017model}, we perform sampling in parallel on a graphics processing unit (GPU) with Nvidia\textquotesingle s CUDA architecture. In our implementation, all of the trajectory samples are processed individually in parallel. In addition to the sample level parallelism, each individual sample uses between 4 and 16 CUDA threads depending on the dynamics model. This is done in order to take advantage of the parallel nature of the linear algebra routines that our vehicle dynamics models rely on. Depending upon the model and cost function, our implementation can achieve control loops from 40-60 HZ using a few thousand samples of 2-3 second long trajectories. Note that sampling 1200, 2.5 second long trajectories at 40 Hz corresponds to making approximately 4.8 million queries to the full non-linear dynamics of the vehicle every second. For the complex vehicle dynamics that we consider, this is only possible using a modern GPU. 

\subsection{Real-Time MPC Algorithm}


With our information theoretic control update law, as well as methods for handling control constraints, smoothing, and real-time sampling, we are now ready to describe the full information theoretic model predictive control (IT-MPC) algorithm. The algorithm (Alg. \ref{Algorithm:sb_mpc}) starts by taking in the current state from an external state estimator, and then produces $K$ trajectory samples in parallel on the GPU. Each sample is generated by randomly sampling a sequence of control perturbations, each drawn from $q(V| \hat{U}, \Sigma)$, then the dynamics are simulated forward, and the cost is computed for each trajectory.

Once the costs for each perturbation sequence are computed, they are converted to probability weights, this is done using the method in Alg. \ref{Algorithm:mppi}. After the probability weights have been computed, the un-smoothed control update is computed via a probability weighted average over all the perturbation sequences. Lastly, this update is smoothed by passing it through a convolutional filter with the Savitsky-Galoy coefficients. The first control is then sent to the actuators, and the remaining sequence of length $T-1$ is slid down and used to warm-start the optimization at the next time instance.

\begin{algorithm}
\SetKwInOut{Input}{Given}
\Input{$\vF, g$: Transition Model\;
       $K$: Number of samples\;
       $T$: Number of timesteps\;
       $(\vu_0, \vu_1, ... \vu_{T-1})$: Initial control sequence\;
       $\Sigma, \phi, c, \gamma, \alpha$: Cost functions/parameters\;
       $\text{SGF}$: Savitsky-Galoy convolutional filter\;}        
\While{task not completed}{
$\vx_0 \leftarrow \text{GetStateEstimate()}$\;
\For{$k \leftarrow 0$ \KwTo $K-1$}{
  $\vx \leftarrow \vx_0$\;
  Sample $\mathcal{E}^k = \left( \epsilon_0^k \dots \epsilon_{T-1}^k \right), ~\epsilon_t^k \in \mathcal{N}(0, \Sigma)$\;
  \For{$t \leftarrow 1$ \KwTo $T$}{
    \If{$k < (1 - \alpha) K$}{
    	$\vv_{t-1} = \vu_{t-1} + \epsilon_{t-1}^k$\;
    }\Else{
    	$\vv_{t-1} = \epsilon_{t-1}^k$\;
    }
    $\vx \leftarrow \vF(\vx, g(\vv_{t-1}))$\;
  	$S_k \pluseq  c(\vx) + \gamma \vu_{t-1}^\rT \Sigma^{-1} \vv_t$\;
  }
  $S_k \pluseq \phi(\vx)$\;
}

$\text{ComputeWeights}(S_0, S_1, \dots S_k)$\;
\For{$t \leftarrow 0$ \KwTo $T-1$}{
$U \leftarrow U + \text{SGF}*\left(\sum_{k=1}^K w_k \mathcal{E}^k\right)$\;
}

$\text{SendToActuators}(\vu_0)$\;

\For{$t \leftarrow 1$ \KwTo $T-1$}{ 
	$\vu_{t-1} \leftarrow \vu_t$\;  
}
$\vu_{T-1} \leftarrow \text{Intialize}(\vu_{T-1})$\;
}
\caption{Sampling Based MPC \label{Algorithm:sb_mpc}}
\end{algorithm}

\begin{algorithm}
\SetKwInOut{Input}{Given}
\Input{$S_1, S_2, \dots S_K$: Trajectory costs\;
$\lambda$: Inverse Temperature\;}
$\rho \leftarrow \min_k[ S_k) ]$\;
$\tilde{\eta} \leftarrow \sum_{k=1}^{K} \exp\left( -\frac{1}{\lambda} (S_k - \rho) \right)$\;
\For{$k \leftarrow 1$ \KwTo $K$}{
  $w_k \leftarrow \frac{1}{\eta}\exp\left( -\frac{1}{\lambda} (S_k - \rho) \right)$\;
}
\Return $\{w_1, w_2, \dots w_K \}$
\caption{Information Theoretic Weight Computation} 
\label{Algorithm:mppi}
\end{algorithm}


The iterative importance sampling procedure, where trajectories are sampled using the un-executed portion from the previously computed sequence, is key to achieving a high level of performance with the algorithm. However, in the presence of strong disturbances the importance sampling procedure can be problematic. This is because an exceptionally strong disturbance (see Fig.~\ref{Fig:Disturbances}) can push the entire spray of trajectories into low cost regions where it may be impossible to recover. As a solution, we maintain a very small (less than one percent, denoted by $\alpha$ in Alg. \ref{Algorithm:sb_mpc}) number of trajectories which do not use any importance sampler (i.e they are just Gaussian perturbations around zero). This enables the IT-MPC algorithm to reset itself if a disturbance destroys the effectiveness of the previously computed control sequence, and as long as the proper importance sampling weight is included it does not bias the monte-carlo estimate.

\section{Relation to Stochastic Optimal Control} \label{soc}

The iterative sampling based procedure described in the previous section minimizes the KL-Divergence between the controlled and optimal distribution. This notion of optimality differs from the usual notion of optimality in stochastic optimal control. The goal of this section is to illuminate  the differences between these two notions  and identify the special cases where the two notions of optimality coincide. The tool most appropriate for this task is path integral control theory \cite{kappen2005path, theodorou2010generalized}. We outline the derivation of the path integral control law below, for a full description see \cite{williams2017model}. In the path integral control framework we consider a control-affine, stochastic differential equation of the form:
\begin{equation}   \nonumber
\rd \vx_t = \left( \vf(\vx_t) + \vG(\vx_t)\vu_t \right) \rd t + \vB(\vx_t) \rd \vw,
\end{equation}
where $\vB$ defines the covariance of the system. The cost function is then assumed to take the form:
\begin{equation} \nonumber
\mathcal{L}(\vx_t, \vu_t) = c(\vx_t) + \frac{1}{2}\vu_t^\rT \vR(\vx_t) \vu_t,
\end{equation}
where the control cost matrix $R$ is positive definite and satisfies the condition:
\begin{equation} \nonumber
\vB \vB^\rT = \lambda \vG \vR^{-1} \vG^\rT, ~ \forall \vx.
\end{equation}
The optimal controls then take the form:
\begin{equation}
\label{Equation:Soc}
\vu(\vx_t)^* = -\vR(\vx_t)^{-1}\vG(\vx_t)V_\vx,
\end{equation}
where $V_\vx$ is the gradient of the value function with respect to the current state. The value function in the stochastic optimal control framework is defined as:
\begin{equation} \nonumber
V(\vx) = \min_{\vu} \ExP{\mathcal{Q}}{\phi(\vx_T) + \int_{0}^T \mathcal{L}(\vx_s, \vu_s) \rd s},
\end{equation}
where $\mathcal{Q}$ denotes the controlled distribution induced by the \emph{continuous system dynamics}. The value function will satisfy the following stochastic Hamilton-Jacobi-Bellman partial differential equation:
\begin{equation}\nonumber
-\partial_t V = c(\vx) + \vf^\rT - \frac{1}{2}V_\vx^\rT \vG \vR^{-1} \vG^\rT V_\vx + \frac{1}{2}\rtr\left(\vB \vB^\rT V_{\vx \vx} \right).
\end{equation}
If one could solve the partial differential equation (PDE) and obtain the derivative $V_\vx$, then the problem would be solved. However, due to the curse of dimensionality, directly solving the PDE using numerical methods is tractable only for systems with a very small number of dimensions. The path integral approach is based on the insight that the PDE can be transformed into a path integral, which is an expectation over all possible system trajectories. This transformation is obtained by making an exponential transformation of the value function:
\begin{equation} \nonumber
V(\vx) = -\lambda \log \left(\Psi(\vx) \right),
\end{equation}
which, combined with the assumption on the control cost, enables the stochastic HJB-PDE to be transformed into the linear PDE:
\begin{equation} \nonumber
\partial \Psi_t = \frac{\Psi}{\lambda}c(\vx) - \vf^\rT \Psi_\vx - \frac{1}{2}\rtr\left(\vB \vB^\rT \right)\Psi_{\vx \vx}.
\end{equation}
This, in turn, enables the Feynman-Kac lemma \cite{kac1949distributions} to be applied. This expresses the solution of $\Psi$ in terms of the path integral:
\begin{equation}
\label{Equation:fc_result}
\Psi(\vx) = \ExP{\mathcal{P}}{\exp\left(-\frac{1}{\lambda}S(\tau)\right)\Psi(\vx_T)},
\end{equation}
where $\mathcal{P}$ is the distribution induced by the uncontrolled continuous system dynamics and $S(\tau)$ is the state-dependent portion of the cost:
\begin{equation}\nonumber
S(\tau) = \phi(\vx_T) + \int_0^T c(\vx_t) \rd t.
\end{equation}
The optimal control is then obtained by differentiating $-\lambda \log(\Psi)$, and substituting the result into \eqref{Equation:Soc}. The final product is the path integral form of the optimal controls given by:
\begin{equation}\nonumber
\vu^* \rd t = \vR^{-1} \vG^\rT \left(\vG \vR^{-1} \vG^\rT \right)^{-1} \vB \frac{\ExP{\mathcal{P}}{\exp\left(-\frac{1}{\lambda} S(\tau) \right) \rd \vw}}{\ExP{\mathcal{P}}{\exp\left(-\frac{1}{\lambda} S(\tau) \right)}},  
\end{equation}
which can be approximated in discrete time as:
\begin{align} \nonumber
\vu^* &= \vR^{-1} \vG^\rT \left(\vG \vR^{-1} \vG^\rT \right)^{-1} \vB \frac{\ExP{\mathcal{P}}{\exp\left(-\frac{1}{\lambda} S(\tau) \right) \frac{\epsilon}{\sqrt{\Delta t}}}}{\ExP{\mathcal{P}}{\exp\left(-\frac{1}{\lambda} S(\tau) \right)}},  \nonumber \\
\epsilon &\sim \mathcal{N}(0, I_{m \times m}). \nonumber
\end{align}
Note that in the case that $\vG$ and $\vB$ don't have full rank we can decompose the system into indirectly and directly actuated parts:
\begin{equation} \nonumber
\vx = \begin{pmatrix}\vx_a \\ \vx_c \end{pmatrix}, ~~
\vG = \begin{pmatrix}\vG_a \\ \vG_c \end{pmatrix}, ~~ 
\vB = \begin{pmatrix}\vB_a \\ \vB_c \end{pmatrix},
\end{equation}
and then express these equations in terms of the directly actuated components ($\vx_c$, $\vB_c$, $\vG_c$) of the system. In comparing a discrete time system with a continuous time one, the choice of unit of time is arbitrary. So, without loss of generality, take $\Delta t = 1$. We then obtain two expressions for the optimal controls:
\begin{align}
\label{Equation:UPI}
\vu_{\text{PI}}^* &= \vR^{-1} \vG^\rT \left(\vG \vR^{-1} \vG^\rT \right)^{-1} \vB \frac{\ExP{\mathcal{P}}{\exp\left(-\frac{1}{\lambda} S(\tau) \right) \epsilon}}{\ExP{\mathcal{P}}{\exp\left(-\frac{1}{\lambda} S(\tau) \right)}} \\
\label{Equation:UIT}
\vu_{\text{IT}}^* &= \frac{\ExP{\Pb}{\exp(-\frac{1}{\lambda}S(V))\vv_t}}{\ExP{\Pb}{\exp(-\frac{1}{\lambda}S(V))}}, ~~ \vv \sim \mathcal{N}(0, \Sigma)
\end{align}
The path integral form of the optimal controls is given in \eqref{Equation:UPI}, and its optimality is based on the analysis of the classical stochastic HJB-PDE. In contrast, \eqref{Equation:UIT} gives the form of the optimal controls obtained through the information theoretic framework from section~\ref{sec:Importance-Sampling}. Both of these equations rely on a path integral which computes a negative exponentiated cost-weighted average over trajectories. The difference between the two is the space in which sampling takes place: in the information theoretic case the sampling takes place directly in control space, whereas in the path integral case the sampling takes place in trajectory space and it therefore requires the projection operator $\vR^{-1}\vG^\rT \left(\vG \vR^{-1} \vG^\rT \right)^{-1}$ to be applied. If we make the additional assumption in the path integral case that the noise enters the system through the control input $\vB = \vG \sqrt{\Sigma}$, with $\Sigma$ the noise profile for the control only, then we obtain:
%
%
\begin{align}\begin{split}
\vu_{\text{PI}}^* &= \frac{\ExP{\mathcal{P}}{\exp\left(-\frac{1}{\lambda} S(\tau) \right) \sqrt{\Sigma} \epsilon}}{\ExP{\mathcal{P}}{\exp\left(-\frac{1}{\lambda} S(\tau) \right)}}
= \vu_{\text{IT}}^*, \\
\vR &= \lambda \Sigma^{-1}.\end{split} \label{Equation:PI_Control}
\end{align}

Note that \eqref{Equation:PI_Control} can be directly obtained from \eqref{Equation:IT-Controls} by setting the base and importance sampling controls to the uncontrolled system dynamics and plugging in the definition of $w(V)$. Although similar, the two optimal control expressions in and \eqref{Equation:IT-Controls} and \eqref{Equation:PI_Control}  differ in two significant ways, namely the output of the optimization process, and the assumptions required for the derivation.

\subsubsection{Output of the Optimization} In the case of minimizing the KL-Divergence, we get an entire open loop control plan, whereas in the path integral case the equations only provide an update law for the current time-step. Getting an entire open loop plan is helpful because it provides a principled way to iteratively improve the importance sampling (this is the warm starting procedure in Alg. \ref{Algorithm:mppi}). In contrast, prior work on model predictive path integral control \cite{williams2017model} is based on dynamic programming, and importance sampling is incorporated in a heuristic fashion.
%
%

\subsubsection{Assumptions Required} The most significant difference between the two approaches is that the path integral control derivation requires the dynamics to be affine in the control input, whereas in the information theoretic setting the dynamics can be represented by an arbitrary non-linear function. Although the control-affine assumption covers a large class of systems, notably absent are ground vehicle dynamics and many function approximators popular in machine learning. (e.g. neural networks). Additionally, for the stochastic Hamilton-Jacobi-Bellman equation to be valid the cost and dynamics need to satisfy certain regularity conditions. In the information theoretic setting the cost and dynamics only need to be measurable functions.

\begin{figure}[t]
\centering
\includegraphics[width=.95\columnwidth]{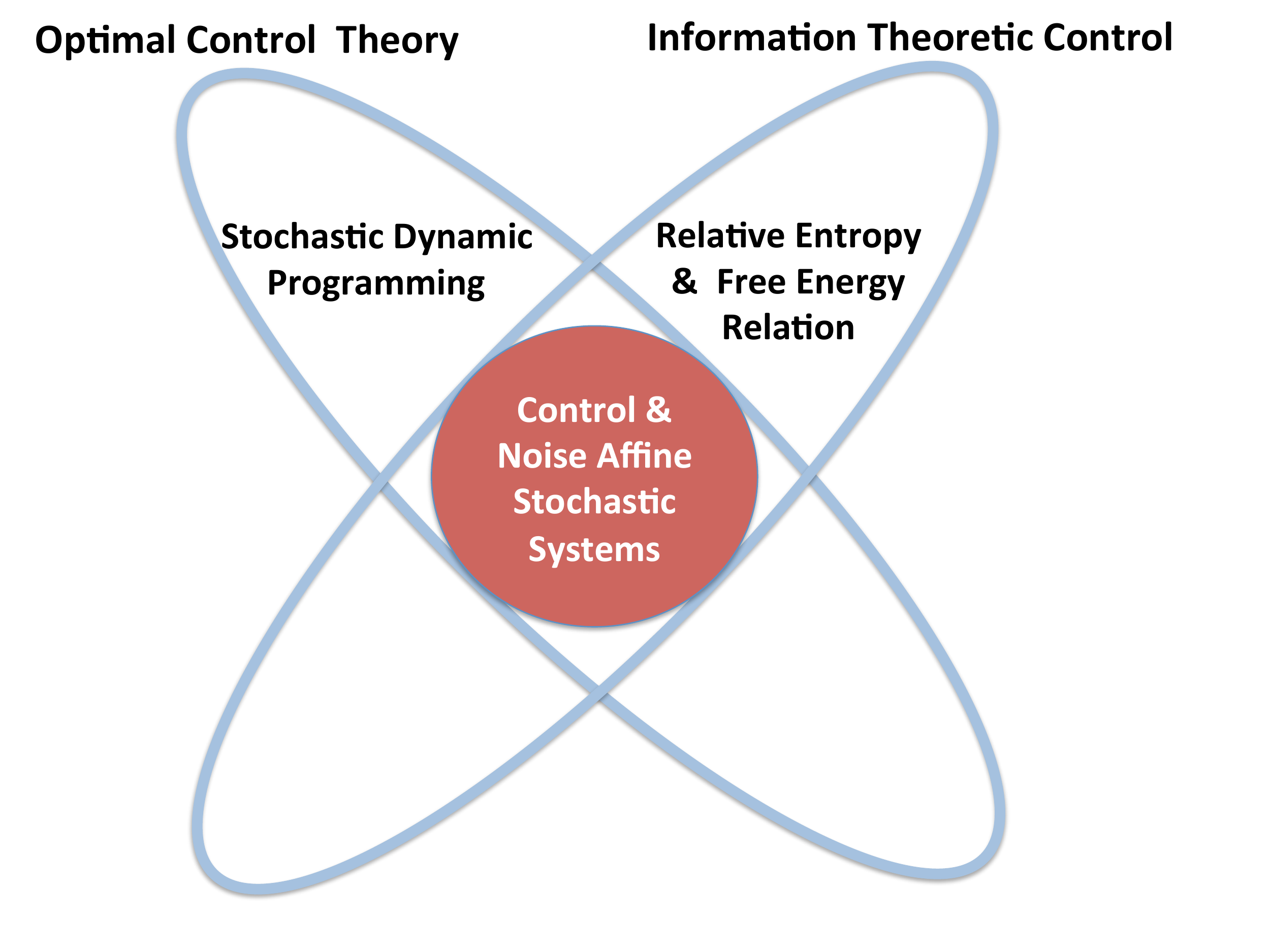}
\caption{Connection between stochastic optimal control theory and information theoretic control. This connection is exact for the case of control and noise affine stochastic systems.}
\label{fig:soc}
\vspace{-4mm}
\end{figure}

The information theoretic approach can exactly recover the path integral optimal control law when  control and noise affine dynamics are considered. In this case the information theoretic quantities  of  free energy  and relative entropy are  expressed in the space of state trajectories. Moreover,  the free energy  becomes a value function since it  satisfies the HJB equation from which the corresponding optimal control can be derived \cite{Theodorou_Entropy2015,Theodorou_DCD_2012}. The  equivalence between the two approaches in the control affine case  relies on the fact that the Feynman-Kac lemma holds in both directions. In particular, given a  backward and linear PDE  there exists an expectation of a cost function which,  when evaluated on sampled trajectories generated from the corresponding stochastic differential equation, provides the probabilistic representation of the solution of the PDE. And vice versa, for a pair of an expectation and a stochastic differential equation used to generate trajectories to evaluate the expectation,  there exist  a backward  and linear PDE with solution  equal to the  aforementioned expectation. It is therefore the Feynman-Kac lemma that creates the connection between the relative entropy-free energy relation  and dynamic programming for the case of control affine dynamics and solidifies our information theoretic framework by creating connections with traditional stochastic control methods and notions of optimality.
 
\section{Related Work on Sampling Based Control} \label{relateds}

Sampling based optimization has a long history in robotics, especially in the reinforcement learning domain. As such, there are a number of alternative approaches that could be used to derive a similar update law to \eqref{Equation:IT-Controls}. Both the policy gradient theorem \cite{peters2006policy} and reward weighted regression \cite{dayan1997using, peters2007reinforcement} could be used, by choosing an appropriate control parameterization and cost transformation, to create a sampling based update law similar to \eqref{Equation:IT-Controls}. However, this would be an ad-hoc approach since the choice of an exponential transformation of the cost function is theoretically unjustified in those frameworks, whereas in our framework it naturally emerges due to the form of the optimal distribution \eqref{Equation:Optimal-Density}. 
%
%
Moreover, those methods are designed for iterative optimization of the parameters of feedback policies, as opposed to open loop control laws, and therefore only guarantee convergence to a local minimum. In our framework, the solution is the global minimum of the KL-Divergence optimization objective, as long as the sampling procedure sufficiently explores the state space. This is important, since fast convergence is crucial and step sizes are hard to tune while running in an online setting. 
%
%
%
%

\subsection{Cross-Entropy for Motion Planning}

The cross-entropy method for motion planning \cite{kobilarov2012cross} is the previous work which is the closest mathematically to our approach. As in our case, in the cross-entropy method the objective function has the form:
\begin{equation}
\theta^* = \argmin_U \KL{\tilde{\Qb}^*}{\Qb},\nonumber
\end{equation}
where $\tilde{\Qb}^*$ is an optimal distribution and $\Qb$ is the distribution induced by the control parameters $\theta$. However, instead of using the free-energy lower bound as we do, in the cross-entropy method the density of the optimal distribution is defined as follows:
\begin{equation}
\tilde{q}^*(V) = I\left({ \{C(V) \le \gamma \} } \right), \nonumber
\end{equation}
where $I$ is the indicator function, $C$ is the cost-to-go function, and $\gamma$ is a constant upper-bound on the trajectory cost that we would like to enforce. In order to optimize this objective, the following iterative procedure is proposed:
\begin{enumerate}
\item Sample parameters $\{\theta_1, \theta_2, \dots \theta_K \}$ from a given proposal distribution $P^i(\theta)$ (usually a Gaussian or a Gaussian mixture model).
\item Determine the elite parameter set threshold: $\gamma_i = C(V, \theta_j)$ where $j$ is the index of the $L_{th}$ best trajectory sample. $L < K$.
\item Compute the elite parameter set: $E_s = \{\theta_k | C(V;\theta_k) \le \gamma_i \}$
\item Update the parameters using expectation maximization over the elite set: $P^{i+1} \leftarrow \text{EM}(E_s)$
\item If converged end, otherwise repeat.
\end{enumerate}
In the case of optimizing the mean of a Gaussian distribution, the cross-entropy method described here is identical to the information-theoretic approach, except that the cross-entropy method takes an \emph{un-weighted} average over the top $L$ sampled parameters. In contrast, the information-theoretic approach takes a weighted average over all the parameter samples. This is an important difference when planning trajectories since the information theoretic approach has more discriminative power over rejecting (assigning very low weight) samples, whereas cross-entropy must assign the same weight to the top $L$ samples, even if those samples have very different cost values. 

Since cross-entropy is the closest related work to our information theoretic approach, we have developed a model predictive controller based on it, and we provide an experimental comparison in section~\ref{results}.  It should be noted that, although the cross-entropy method is a popular stochastic optimization technique in robotics, it has not previously been applied in a model predictive control framework using massively parallel sampling with a GPU.  
%
%

One important modification to the cross-entropy method that we make is that we \emph{do not} update the sampling covariance in our model predictive control algorithm. This is because the sampling covariance rapidly shrinks once it converges on a good trajectory. This is desirable in the case of open loop trajectory optimization or parameterized policy learning \cite{stulp2012path}, but in the model predictive control case it is problematic since the environment changes at every step, which means that a good policy can turn catastrophic in a few time-steps. Thus, in a receding horizon setting, the sampling covariance needs to be able to both shrink and grow adaptively. We experimented with a number of simple methods for growing the covariance, but none proved satisfactory in a general setting so we elected to keep the covariance constant. The cross-entropy MPC algorithm that we implement uses the same sampling based MPC method from Alg. \ref{Algorithm:mppi}, except that the computation of the trajectory weights is different. The weight computation used for cross-entropy is given by Alg. \ref{Algorithm:cem}.

\begin{algorithm}
\SetKwInOut{Input}{Given}
\Input{$S_1, S_2, \dots S_K$: Trajectory costs\;
$\delta$: Eliteness Threshold\;}
$\zeta \leftarrow \text{round}(K(1 - \delta))$\;
$Z \leftarrow S_\zeta$\;
\For{$k \leftarrow 1$ \KwTo $K$}{
  $w_k \leftarrow \frac{1}{\zeta}I\{S_k < Z\}$\;
}
\Return $\{w_1, w_2, \dots w_K \}$
\caption{Compute Weights (Cross Entropy) \label{Algorithm:cem}}
\end{algorithm}

%
%


\section{Experimental Setup}\label{experimental-setup}

We applied the IT-MPC algorithm to the task of driving a 1:5 scale autonomous vehicle around a dirt test track.  In prior work \cite{Williams-ICRA-17}, we demonstrated the capability of an earlier version of the IT-MPC algorithm to control our vehicle platform and several other dynamical systems in simulation. In this work, we focus on the most challenging task, aggressive autonomous driving with a real-world platform, allowing us to focus on a thorough evaluation by collecting an order of magnitude more data. Autonomous driving is currently one of the most important application areas for robotics control, and our experiments were designed to probe the strengths and limitations of the IT-MPC algorithm at this difficult task. There are two characteristics of the autonomous driving problem that suggest that the sampling based optimization underlying the IT-MPC algorithm can be uniquely capable at the autonomous driving task, these are:
\begin{enumerate}
\item Most of the dynamic regimes of the vehicle system are highly non-linear, but not unstable. This is beneficial because, unlike most methods, non-linear dynamics are not an issue for IT-MPC. However, highly unstable systems are difficult to sample useful trajectories from, so the fact that vehicle dynamics are usually not unstable makes it feasible and beneficial to use our algorithm.
\item In autonomous driving, there are a number of constraints that are difficult to classify as either soft or hard. For example, avoiding a barrier might seem at first to be a hard constraint. However, if collision with a barrier is unavoidable, it is still important to continue controlling the vehicle in order to get out of collision. In this way, a barrier also has the properties of a soft-constraint. The IT-MPC controller can handle this by including an impulse-like cost for collisions, which enforces collision avoidance but retains the ability of the vehicle to navigate in the vicinity of the barrier.
\end{enumerate}
Our experiments were designed to determine whether these hypothesis are accurate, and to test how the algorithm performed as the requested speed was increased far beyond the friction limits of the vehicle-track system. As a baseline, we compare the IT-MPC controller against an MPC implementation of the cross-entropy method (CEM-MPC).
%
%

\subsection{AutoRally Vehicle Testbed}
\begin{figure}
\centering
\includegraphics[width=.95\columnwidth]{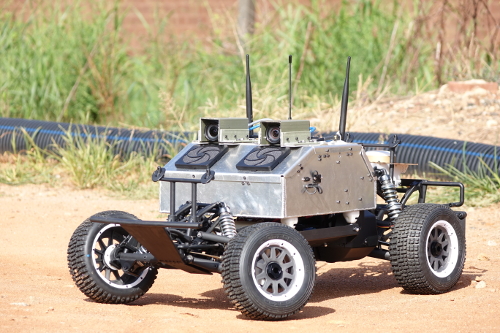}
\caption{1/5 scale AutoRally vehicle. Computing hardware is located inside the aluminum box, and the GPS is located at the rear of the vehicle.}
\label{Fig:AutoRally}
\vspace{-3mm}
\end{figure}

The AutoRally robot (Fig. \ref{Fig:AutoRally}) is an electric autonomous vehicle testbed 1/5 the size of a full scale car that is designed to be robust, safe, and easy to use. AutoRally is approximately 22 kg, measures 0.9 m long, and has a top speed of 113 kph. Fully autonomous driving is possible using only the onboard sensing and computing, and the software interface is built with the Robot Operating System (ROS) on Ubuntu. The onboard computer consists of a Mini-ITX motherboard, an Intel quad-core i7 processor, 16 GB RAM, 2 SSDs, an Nvidia GTX-750ti graphics card, and a 222 Wh battery. The rugged aluminum compute box enclosure is designed to withstand violent vehicle rollovers without damaging internal components. The sensor package includes 2 forward facing cameras, a Lord Microstrain 3DM-GX4-25 IMU, an RTK corrected GPS receiver, and Hall effect wheel speed sensors. The GPS is housed in a separate protective enclosure at the rear of the vehicle to provide the best signal, avoid interference from other electronics, and remain inside the protective plastic body cover. 

AutoRally has a three layer safety system to ensure that the vehicle can be remotely stopped at any time. The first is a software runstop which disables throttle control during autonomous operation. The second layer is a switch on the RC transmitter that allows seamless switching between manual and autonomous control. The final safety system layer is a live-man relay that physically breaks the throttle signal to disable all motion. This relay is operated by a button on the grip of the RC transmitter and also triggered automatically in the event of a power failure on the robot.

Model predictive control algorithms require accurate state feedback in order to operate, these include the vehicle pose and velocity. The state estimate is computed by combining the IMU and GPS measurements in an optimization framework that operates on factor graphs. The factor graph is constructed from the asynchronous sensor data, and iteratively optimized using the software package GTSAM and iSAM2\cite{kaess2012isam2}. To keep computational loads low and maintain high accuracy, this graph optimizes for state nodes at 10Hz, corresponding to GPS measurements \cite{Forster-RSS-15}. A 200 Hz state estimated is generated by integrating the IMU measurements to interpolate the state between the 10Hz GPS positions.

\subsection{Dynamics Models}
In order to deploy the IT-MPC algorithm we require an approximate model of the system dynamics. In the control literature there exist several types of models for full-scale vehicles \cite{Velenis2010,Lundahl2011,Burhaumudin2012}, as well as simplified ``bicycle'' vehicle models. However, there are a number of challenges in applying these models to the AutoRally system. Most notable among these are the dirt track which makes applying friction models meant for pavement difficult, and the significant roll dynamics of the vehicle which makes applying simplified models inaccurate. To circumvent these problems, we applied two machine learning approaches: one hybrid-physics based approach, and a pure machine learning approach using a fully-connected feed-forward neural network.
%
%

Both models have the same state-space description of the AutoRally vehicle with seven state variables: x-position, y-position, heading, roll, longitudenal (body-frame forward) velocity, lateral (body-frame sideways) velocity, and heading rate. These are denoted as $\left(p_x, p_y, \theta, r, v_x, v_y, \dot{\theta}\right)$ respectively. The two control variables are the steering and throttle inputs, which are denoted by $u_1$ and $u_2$.

A certain subset of the equations of motion are kinematically trivial given the state space representation. We therefore partition the state space into kinematic state variables, $\vx_k$, and dynamic state variables, $\vx_d$, such that:
\begin{equation}
\vx = 
\begin{pmatrix}
\vx_k \\  
\vx_d
\end{pmatrix}. \nonumber
\end{equation}  
Let $\vx_k = (p_x, p_y, \theta)^\rT$, then we can write the equations of motion for the kinematic variables as:
\begin{equation}
\vx_k(t+1) = \vx_k(t) + \vk(\vx)\Delta t, \nonumber
\end{equation}
where the function $\vk(\vx)$ is defined as:
\begin{align}
\vk(\vx) = 
\begin{pmatrix}
\cos(\theta)v_x - \sin(\theta)v_y\\
\sin(\theta)v_x + \cos(\theta)v_y \\
\dot{\theta}
\end{pmatrix}. \nonumber
\end{align}
Given these kinematic updates, the dynamics model only has to determine the update equations for the dynamic state variables:
\begin{equation}
\vx_d = \left(r, v_x, v_y, \dot{\theta} \right)^\rT . \nonumber
\end{equation}
The dynamics of these variables do not depend on the global coordinate frame, and therefore are not functions of the kinematic state variables. Therefore the equations of motions for the dynamic state variables can be written as:
\begin{equation}
\vx_d(t+1) = \vx_d(t) + \vf(\vx_d, \vv(t)) \Delta t, \nonumber 
\end{equation}
Where $\vv(t) = (u_1(t) + \epsilon_1(t), u_2(t) + \epsilon(t))$ is the randomly perturbed control input. The full equations of motion are then:
\begin{equation}
\vx(t+1) = 
\begin{pmatrix}
\vx_k(t) \\
\vx_d(t)
\end{pmatrix}
+ 
\begin{pmatrix}
\vk(\vx(t)) \\
\vf(\vx_d(t), \vv(t))
\end{pmatrix}
\Delta t. \nonumber
\end{equation}
Given these equations, the challenge is to determine the function $\vf$. Both methods fit their parameters using a system identification dataset collected by a human pilot executing a series of choreographed maneuvers. These maneuvers were:
\begin{enumerate}
\item Slow driving (3 - 6 m/s) around the track.
\item Zig-Zag maneuvers at slow speeds (3 - 6 m/s).
\item High acceleration maneuvers by applying full throttle at the beginning of a straight and applying full brake before entering the
next turn.
\item Sliding maneuvers where the pilot attempts to slide as much as possible.
\item High speed driving where the pilot simply attempts to drive around the track as
fast as possible.
\end{enumerate}
Each maneuver was executed for 3 minutes going counter-clockwise and 3 minutes going clockwise for a total of 30 minutes worth of driving data.

\subsubsection{Basis Function Model}
The basis function model has the form:
\begin{equation}
\vf(\vx_d) = \Theta^\rT \phi(\vx_d), \nonumber
\end{equation}
where $\Theta \in \Rb^{b \times 4}$ and $\phi(\vx) = (\phi_1(\vx), \phi_2(\vx), \dots \phi_b(\vx))^\rT \in \Rb^b$ is a matrix of coefficients and a vector of non-linear basis functions respectively. The term $b$ denotes the number of basis functions in the model. Given this model form, there are two challenges: determining an appropriate set of basis functions, and computing the coefficient matrix $\Theta$. For determining an appropriate set of basis function we analyzed the non-linear bicycle model of vehicle dynamics from \cite{hindiyeh2013dynamics}, and extracted out all of the non-linear functions that appeared in the algebraic equations. This led to a set of 21 basis functions, and then, using trial and error, we added 4 more basis functions to account for the roll dynamics and the non-linear throttle calibration. The vehicle model and the basis functions extracted from it are described in appendix \ref{appendix:bfm}. 

Given a set of basis functions and some data collected from the system, determining the coefficient matrix $\Theta$ is an unconstrained linear regression problem which is easy to solve. We used linear regression with Tikhonov regularization to solve for $\Theta$ given the basis functions and the system identification dataset. Even though we are interested in simulating entire trajectories forward in time, we train the model to minimize the one-step prediction error (i.e. given $(\vx_d(t), \vu(t))$ predict $\vx_d(t+1)$) as opposed to the multi-step prediction error. Minimizing multi-step prediction error is technically the correct objective, but considerably more difficult as the problem becomes non-convex \cite{venkatraman2015improving}. An important detail to note is that performing standard linear regression (without Tikhonov regularization) does not work for multi-step prediction as the weights tend to be very large which results in unstable forward simulation, even if the one-step prediction error is lower than the regularized method.

\subsubsection{Neural Network Model}
The second model that we trained to approximate the dynamics function, $\vf(\vx_d)$, was a multi-layer neural network model. We use a two hidden layer, fully connected model with hyperbolic tangent non-linearities. Each hidden layer had 32 neurons for a total of 1412 parameters. The neural network model is trained using the same 30 minute system identification dataset as the basis function model, and again we minimize one-step prediction error. The model is trained with mini-batch gradient descent using the RMSProp optimizer \cite{hintonlecture} and L2 regularization.

The neural network significantly outperformed the basis function model on a validation. The coefficient of determination, mean squared error, and mean absolute error for the two models on the validation set are shown in Table \ref{Table:DynamicsPerformance}. Despite the inferiority of the basis function model on these testing metrics, we still tested both models in order to determine whether the physics based features provided superior generality to the purely black-box neural network. Both models suffer from significant inaccuracies which reflect the effect of hidden variables, like track condition and battery voltage, that have a significant effect on the dynamics but which are not represented in the state space representation of the system. 

\begin{table}
\caption{Errors for Basis Function and Neural Network}
\begin{tabular}{c|c|c} 
         & Basis Function & Neural Network \\ \hline
R2 Score & .68 & .78 \\
Mean Squared Error & 2.07 & 1.39  \\
Mean Absolute Error & .93 & .76 \\    
\end{tabular}
\label{Table:DynamicsPerformance}
\end{table}

\subsection{Cost Function and Algorithmic Parameters}
There are a number of free parameters in the IT-MPC and CEM-MPC algorithms. We used simulation experiments to initially determine these parameters, and then used a small number of real-world experiments in order to fine tune them. The same cost function and algorithmic parameters were used across all the experimental settings (except for the speed target which modulates how fast the vehicle goes). Table \ref{Table:params} lists the parameter values used during the experiments. 

\begin{table}[htb!]
\caption{IT-MPC and CEM-MPC Parameters}
\begin{tabular}{c|c}
Parameter & Value \\ \hline
Control Frequency & 40 Hz \\
Time Horizon & 2 seconds \\ 
$\lambda$ &  12.5 \\ 
$\gamma$ & 0.1 \\
$\Sigma$ & ${\bf{\text{Diag}}}(0.0306, 0.0506)$ \\
${\bf{\text{Initialize}(\vx_{T-1})}}$ & $(0, 0)$ \\
Eliteness threshold (Cross-Entropy only) & $> 0.8$ percentile 
\end{tabular}
\label{Table:params}
\end{table}

Since the IT-MPC and CEM-MPC are both sampling based methods, we did not design separate cost functions for the two algorithms. This would not be the case in comparing with a gradient based method, where smoothness would have to be enforced. The state-dependent cost function that we used was of the form:
\begin{equation}
\alpha_1{\bf{\text{Track}}}(\vx) + \alpha_2{\bf{\text{Speed}}}(\vx) + \alpha_3{\bf{\text{Stabilizing}}}(\vx)
\end{equation}
The three components of the cost function are as follows:
\subsubsection{Track Cost} For the track cost we require a map representation of the track which gives an indication of how close to the edge the vehicles position is. There are a variety of ways to create such a map, our approach was to take a GPS survey of the boundaries of the track, then a cubic 2-dimensional spline was used to regress a cost map with points on the outer boundary set to 1 and points on the inner boundary set to -1. The absolute value of this map was taken to produce the overall cost map. Lastly, the total cost was capped at 2.5 in order to avoid regression artifacts far away from the track.
The cost-map is stored in CUDA texture memory which enables fast lookups for data exhibiting 2-d locality, it also automatically interpolates the grid so that look-ups with continuous positions are efficient. Letting $h(p_x, p_y)$ denote the value returned by the cost map, the overall track cost can be written as:
\begin{equation}
{\bf{\text{Track}}}(\vx) = h(p_x, p_y) + .9^t \left( 10000I(\{ h(p_x, p_y) > .99\}) \right). \nonumber
\end{equation}
In the second term $t$ is the timestep and $I$ is an indicator function. This is a time-decaying impulse penalty for being located outside the track boundaries. It is necessary to include the time-decay because of disturbances and errors in the dynamics, not using a time-decaying penalty is effective in simulation with perfect dynamics, but fails on the actual system. This is because a strong disturbance can push the importance sampling trajectory far off the track, which results in most samples receiving the impulse cost and being rejected, this destabilizes the optimization. Including the time-decay term enables trajectories which stay on the track until the very end of the horizon to play a role in the optimization, while still enforcing a hard constraint like objective by rejecting trajectories that are immediately about to exit the track. 

\subsubsection{Speed Cost}

The speed cost is a simple quadratic cost for achieving a desired forward speed:
\begin{equation}
{\bf{\text{Speed}}}(\vx) = \left(v_x - v_\text{des}  \right)^2, \nonumber
\end{equation}
where $v_x$ is the longitudinal velocity in body-frame. 

\subsubsection{Stabilizing Cost}

The stabilizing cost penalizes samples which exhibit extreme maneuvers that are known to result in undesirable behaviors (e.g. rollovers and spin-outs). This cost follows the track cost pattern where there is both a soft and hard cost. The stabilizing cost is:
\begin{align}
{\bf{\text{Stabilizing}}}(\vx) &= \zeta^2 + 10000I\left( \left\{ \left| \zeta \right| > .75 \right\} \right) \\
\zeta &= -\arctan \left(\frac{v_y}{\|v_x\|} \right), \nonumber
\end{align}
the term $\zeta$ is known as the side slip angle of the vehicle and measures the difference between the velocity vector of the vehicle and heading angle. Under normal driving conditions the side slip angle of the vehicle is zero. The stabilizing cost function provides a quadratic penalty for slip angles up to .75 radians (approximately 42 degrees), and then rejects any trajectories with a slip angle greater than 0.75 radians.

\section{Results} \label{results}

\begin{figure}
\centering
\includegraphics[width=\columnwidth]{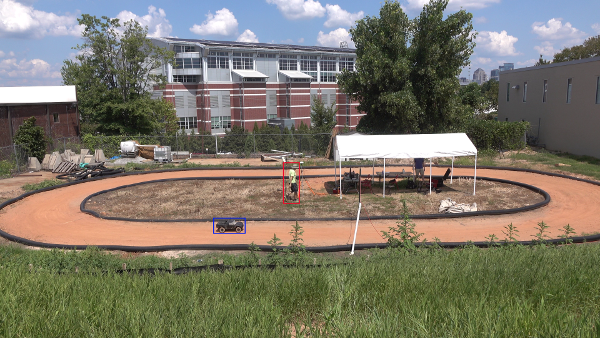}
\caption{Experimental setup at the Georgia Tech Autonomous Racing Facility. All of the state estimation and control software is run on-board the robot itself, making it fully-autonomous and self-contained.}
\label{Fig:TrackPic}
\end{figure}

All experiments were conducted at the Georgia Tech Autonomous Racing facility. The facility consists of a roughly elliptical dirt track which is 30 meters across at its widest point. An image of the track with the robot is shown in Fig. \ref{Fig:TrackPic}. A ground station is set up in the center of the track which consists of an operating control system (OCS) laptop, runstop, and a base station GPS module to provide RTK corrections to the GPS module on-board the robot. The OCS laptop is used to remotely communicate with the robot and monitor its status over WiFi. However, all of the software required for autonomous operation runs on the vehicle's on-board computer. We want to emphasize that \emph{all computations used for driving were performed on-board}. In our experiments, we tested 3 different speed targets (6 m/s, 8.5 m/s, and 11 m/s) for each of the two control methods with each of the two different dynamics models. Each setting was tested by maneuvering the vehicle clockwise and counter-clockwise around the track for 100 laps. Out of the 24 different scenarios, we were able to successfully collect 100 laps for 17 of the test scenarios, for a total of over 1700 laps around the track. This is equivalent to over 100 kilometers of driving data. The other 7 settings resulted in controllers that were too reckless or unstable, so we were unable to complete those trials in their entirety.

Each lap was classified as either a success, a failure, or invalid if the cause of a failure was external to the controller. The controller is not the only part of the system that can cause a failure, much more common are state estimator errors due to loss of the GPS signal. In addition, the first lap in each batch of data was discarded. This is because the starting lap has slightly different statistics than the other laps, due to the vehicle accelerating up from zero velocity. Note that the total number of starts depends on a number of variables outside the scope of the controller. For example, if there was a poor GPS connection it could take 5-10 batches to collect 100 laps, whereas if the controller and state estimator were perfect, 100 laps could be collected in a single run.

\subsection{Overall Performance}

Table \ref{Table:StatTable} shows lap time, success rate, and speed statistics for each of the tested settings, and Table \ref{Table:TrajTraces} shows the raw trajectory traces overlayed onto the track for all of the successful laps at each setting. The vehicle's behavior differed significantly depending upon the choice of algorithm (IT-MPC or CEM-MPC), the dynamics model (basis function or neural network), and the speed target (6m/s, 8.5m/s, or 11m/s). 
\begin{table}
\caption{IT-MPC and CEM-MPC Performance Statistics}
\begin{tabular}{c|c|c|c|}
Method & Success \% & Lap Time & Speed m/s \\ \hline &&& \\
IT-MPC-NN 6 m/s CC       &  100\%       & $16.98 \pm .32$         & $1.99 - 4.96$   \\ \hline &&& \\
CEM-NN 6 m/s CC        &  100\%       & $12.61 \pm .26$         & $2.86 - 6.63$   \\ \hline &&& \\
IT-MPC-BF 6 m/s CC       &  100\%       & $18.42 \pm .21$         & $1.28 - 5.65$   \\ \hline &&& \\
CEM-BF 6 m/s CC        &  83.16\%     & $11.82 \pm .43$         & $0.39 - 6.90$   \\ \hline &&& \\

IT-MPC-NN 6 m/s C       & 100\%        & $16.03 \pm .22$         & $2.58 - 4.78$     \\ \hline &&& \\
CEM-NN 6 m/s C        & 100\%        & $12.58 \pm .25$         & $2.33 - 6.28$    \\ \hline &&& \\
IT-MPC-BF 6 m/s C       & 100\%        & $16.00 \pm .37$        & $1.97 - 5.77$    \\ \hline &&& \\
CEM-BF 6 m/s C        & 97.30\%      & $12.14 \pm .31$         & $1.85 - 7.17$    \\ \hline &&& \\

IT-MPC-NN 8.5 m/s CC       & 100\%     & $11.78 \pm .26$         & $1.84 - 7.5$    \\ \hline &&& \\
CEM-NN 8.5 m/s CC        & 91.20\%   & $10.74 \pm .41$         & $1.60 - 8.39$    \\ \hline &&& \\
IT-MPC-BF 8.5 m/s CC       & 89.10\%   & $11.30 \pm .70$         & $1.16 - 7.71$         \\ \hline  &&& \\
CEM-BF 8.5 m/s CC        & $<$50\%   & N/A                     & N/A              \\ \hline \rowcolor{lightgray} &&& \\

\rowcolor{lightgray} IT-MPC-NN 8.5 m/s C       & 100\%      & $12.16 \pm .33$         & $2.09 - 7.46$              \\ \hline &&& \\
CEM-NN 8.5 m/s C        & 85.42\%    & $10.83 \pm .55$         & N/A              \\ \hline &&& \\
IT-MPC-BF 8.5 m/s C       & 89.00\%    & $9.81 \pm .31$          & $4.22 - 9.72$              \\ \hline &&& \\
CEM-BF 8.5 m/s C        & $<$50\%    & N/A                     & N/A              \\ \hline \rowcolor{lightgray} &&& \\

\rowcolor{lightgray} IT-MPC-NN 11 m/s CC       & 100\%      & $9.27 \pm .30$         & $3.46 - 9.06$              \\ \hline &&& \\
CEM-NN 11 m/s CC        & 66.32\%    & $8.42 \pm .23$         & $4.00 - 10.01$              \\ \hline &&& \\
IT-MPC-BF 11 m/s CC       & $<$50\%      & N/A                  & N/A                       \\ \hline &&& \\

IT-MPC-NN 11 m/s C       & 76.00\%      & $10.09 \pm .35$       & $1.47 - 9.37$                  \\ \hline &&& \\
CEM-NN 11 m/s C        & $<$50\%         &  N/A               & N/A                     \\ \hline &&& \\
IT-MPC-BF 11 m/s C       & $<$50\%         &  N/A               & N/A                      \\ \hline
\end{tabular}
\label{Table:StatTable}
\end{table}

\begin{table*}
\caption{Trajectory traces of successful IT-MPC and CEM-MPC testing runs.}
\begin{tabular}{N|MM|MM}
& IT-MPC-NN & CEM-NN & IT-MPC-BF & CEM-BF \\ \hline 
 & & &\\
CC-6ms & \includegraphics[width=.2\textwidth]{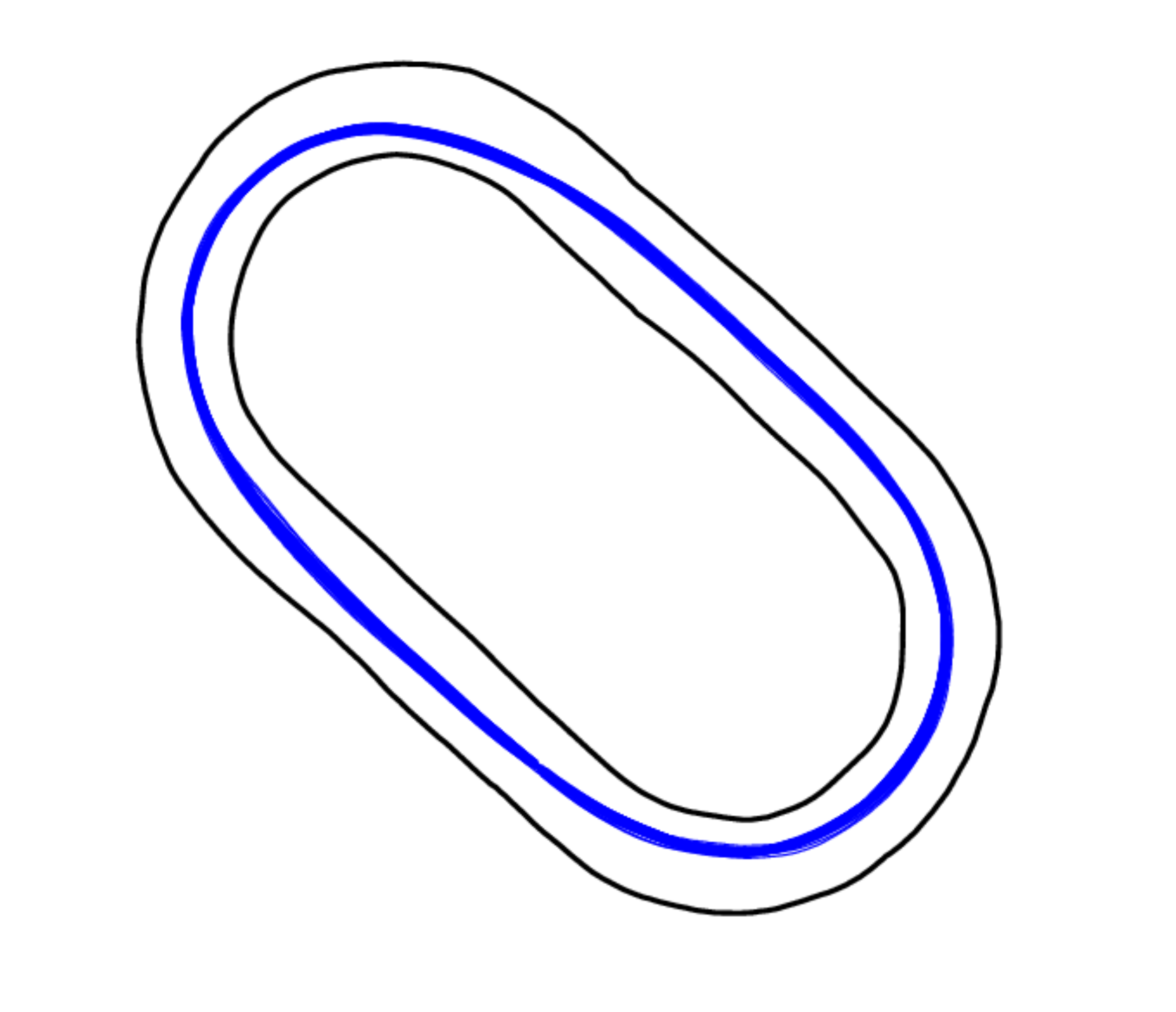} & 
\includegraphics[width=.2\textwidth]{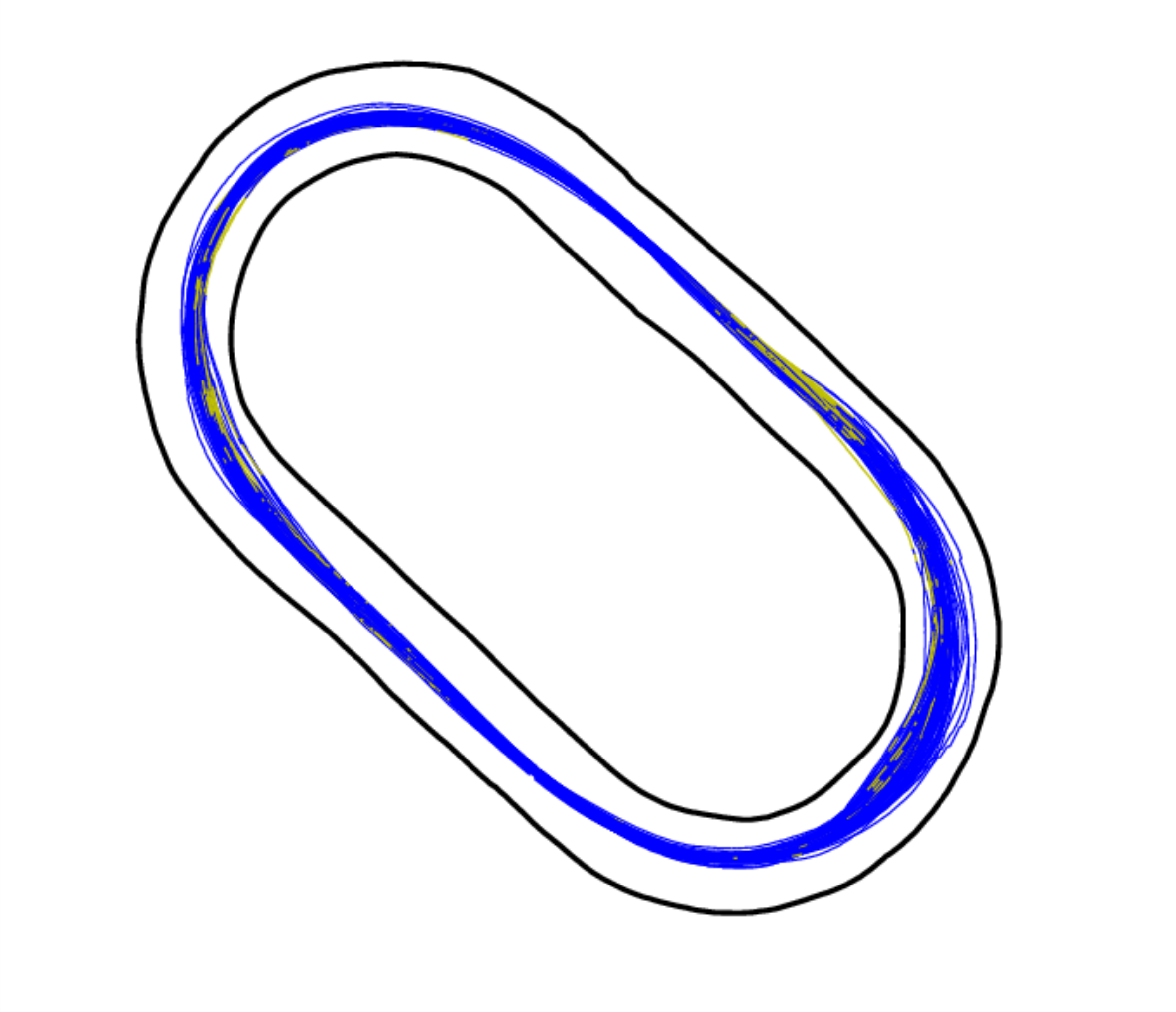} &
\includegraphics[width=.2\textwidth]{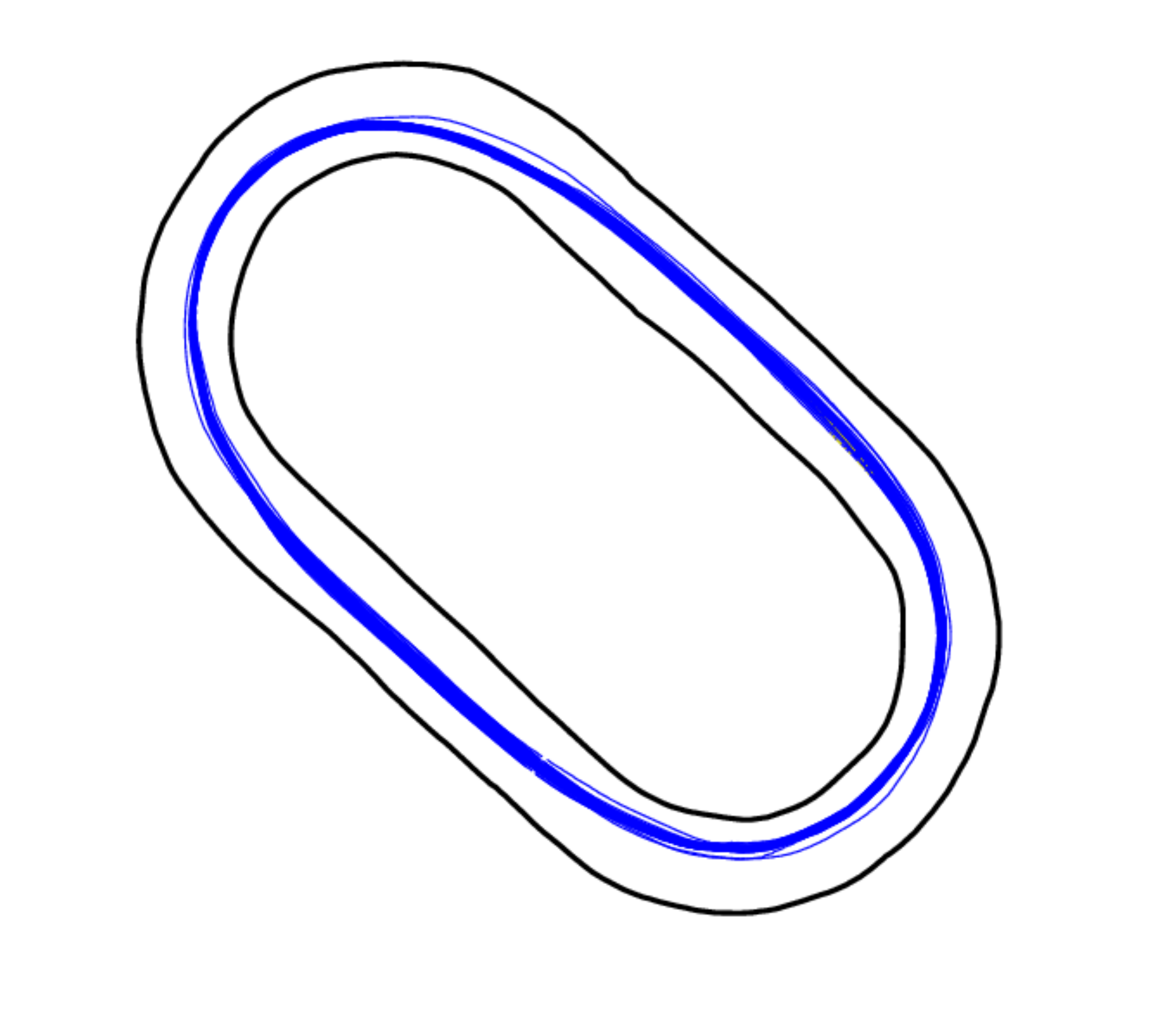} &
\includegraphics[width=.2\textwidth]{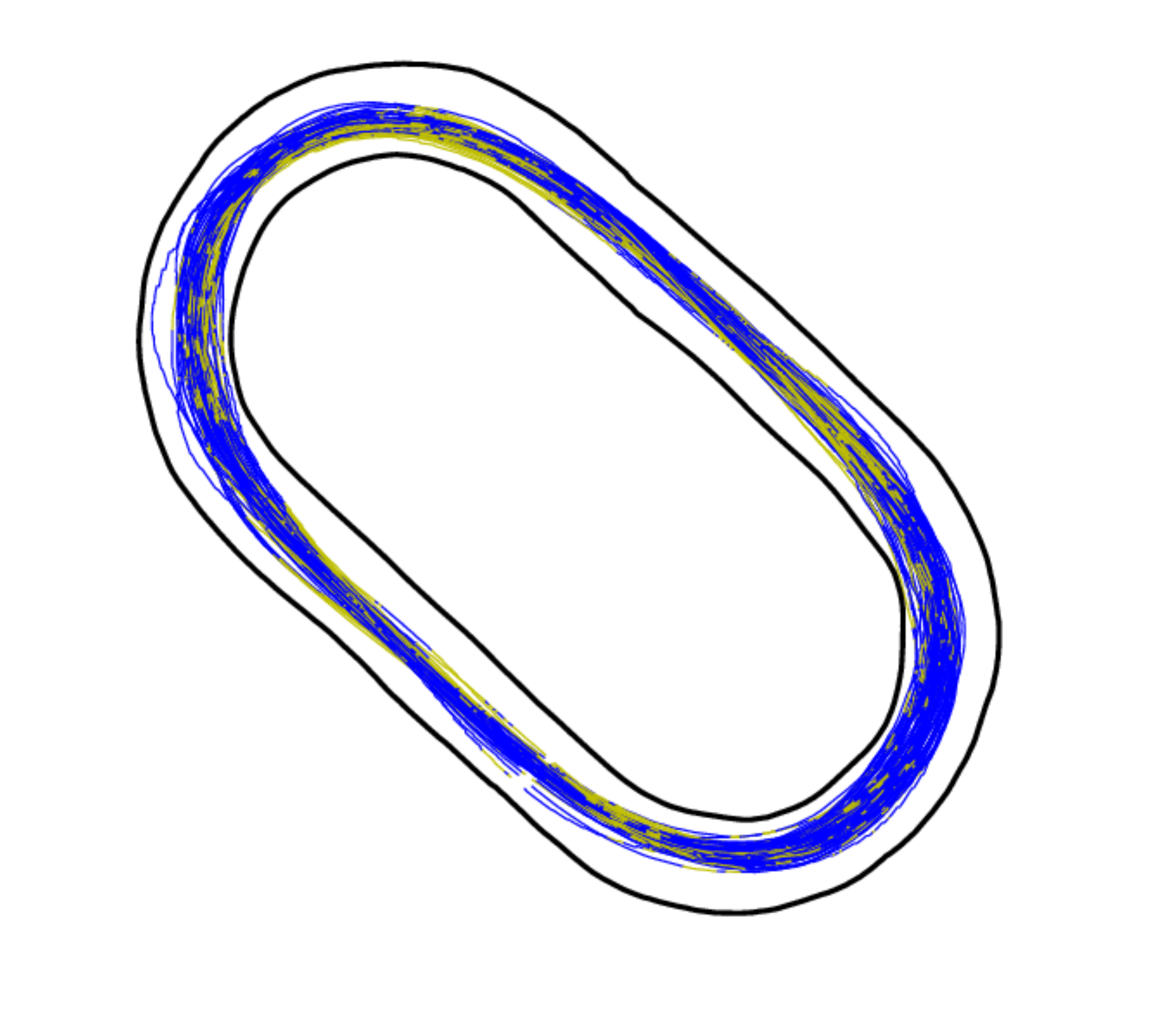}\\
C-6ms & 
\includegraphics[width=.2\textwidth]{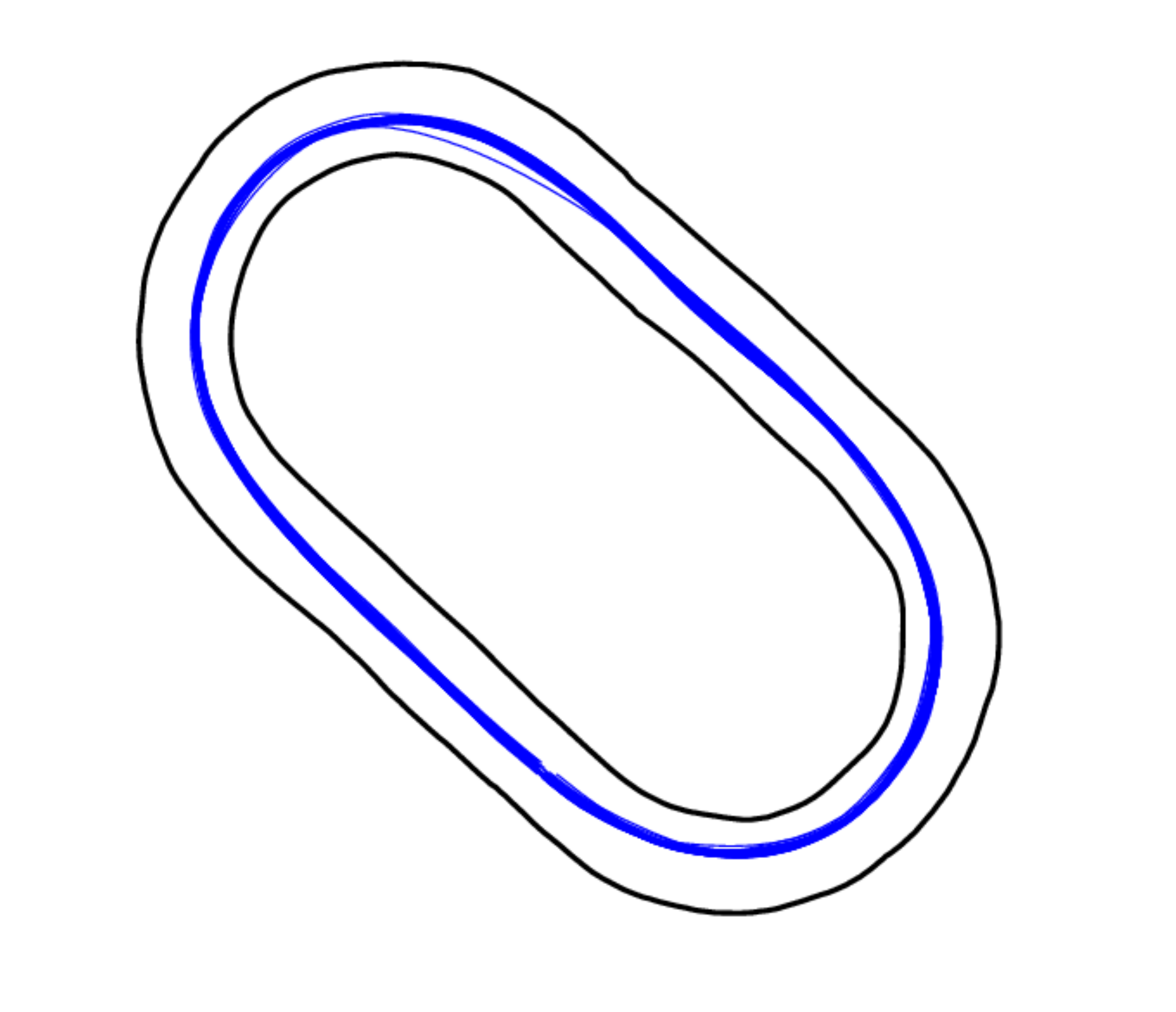} & 
\includegraphics[width=.2\textwidth]{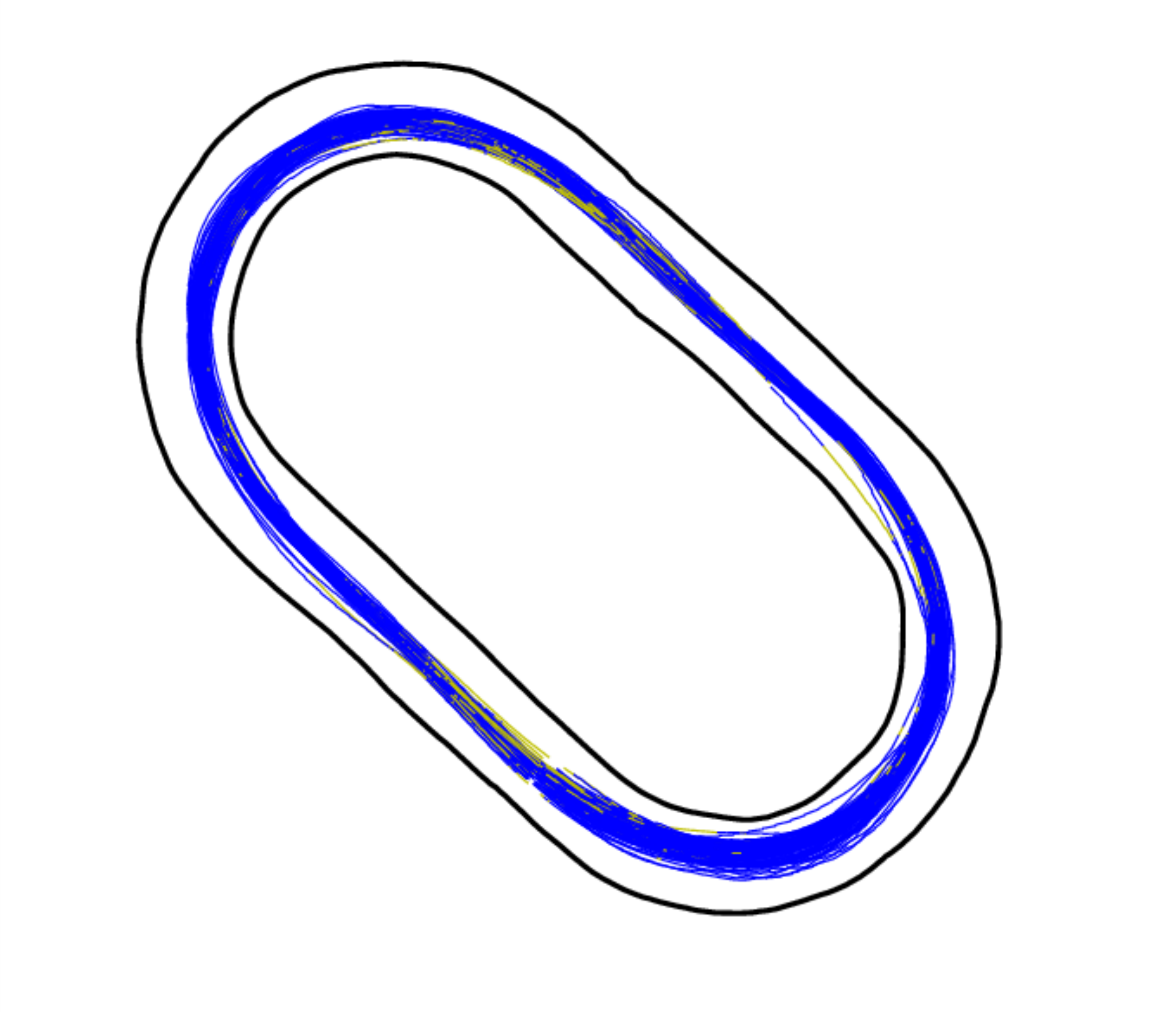} &
\includegraphics[width=.2\textwidth]{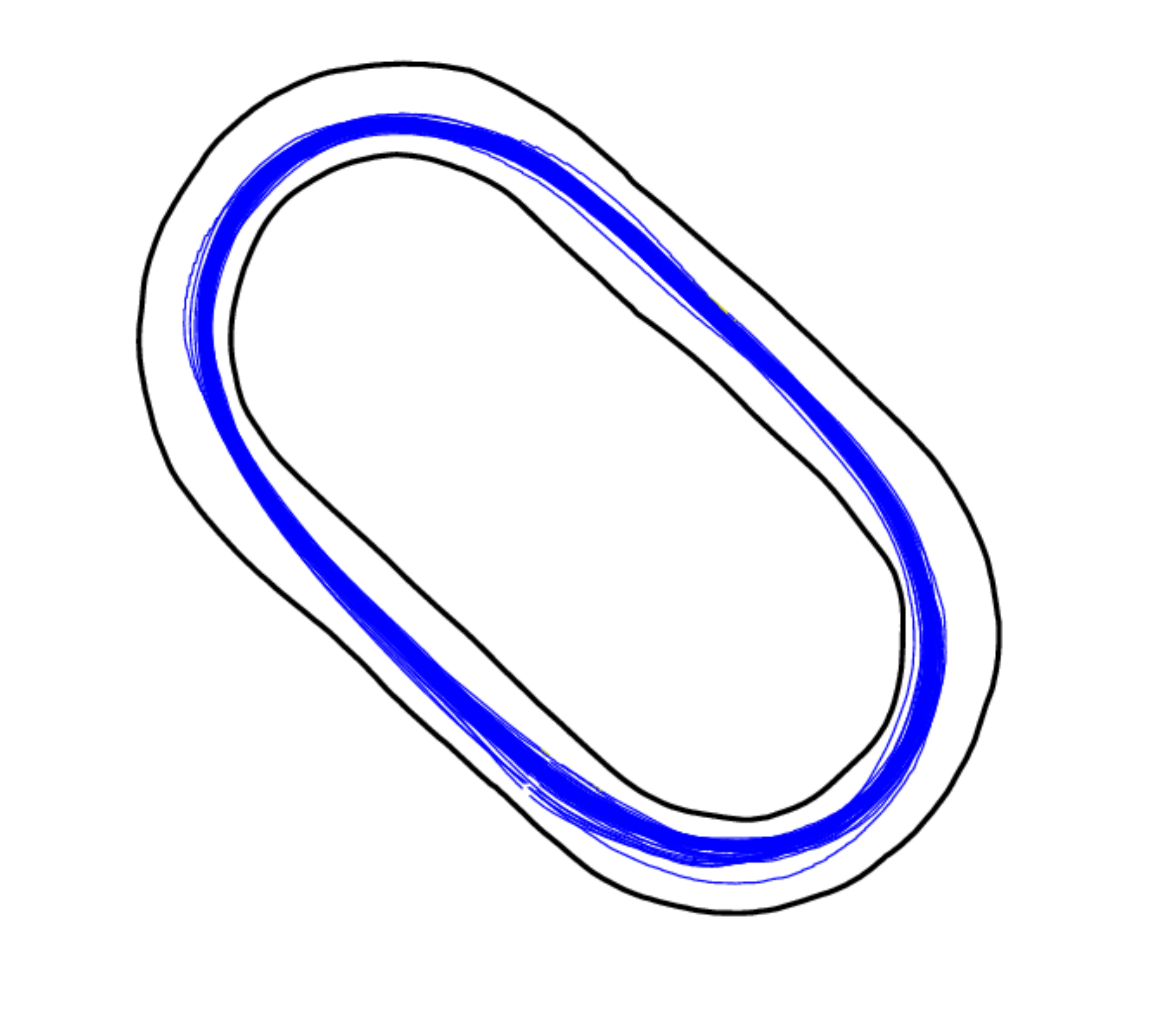} &
\includegraphics[width=.2\textwidth]{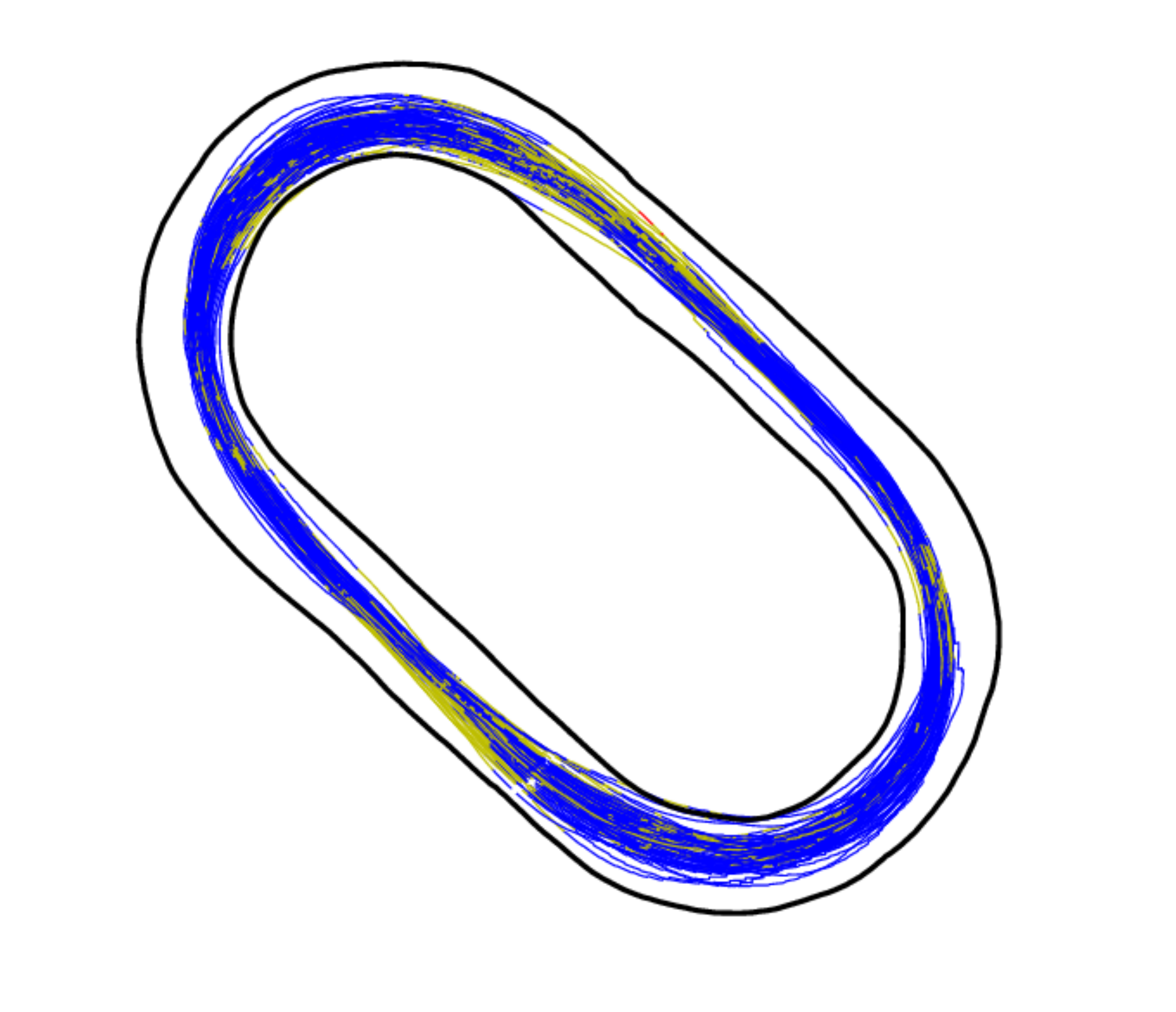}\\ \hline 
 & & &\\
CC-8.5ms & 
\includegraphics[width=.2\textwidth]{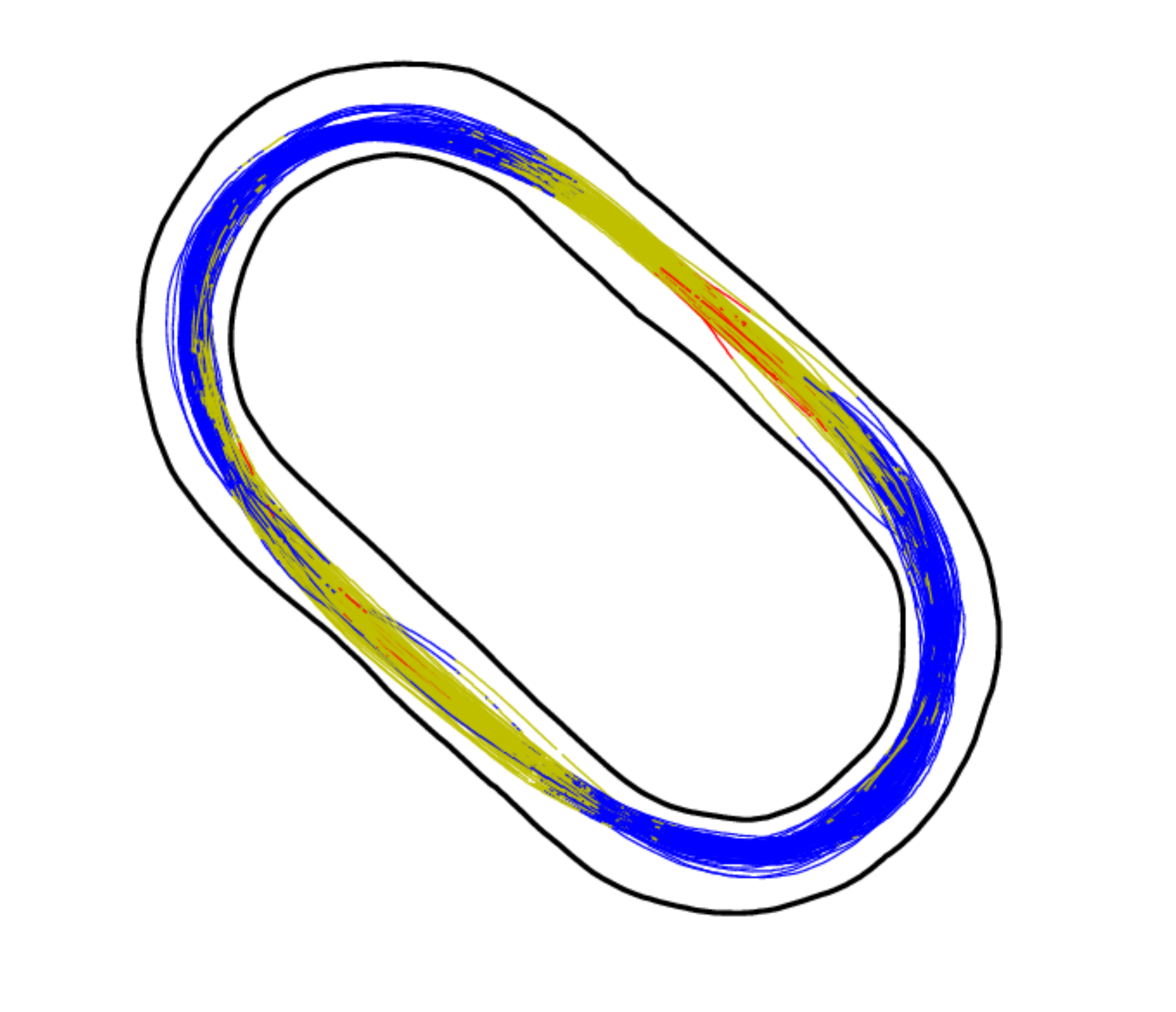} & 
\includegraphics[width=.2\textwidth]{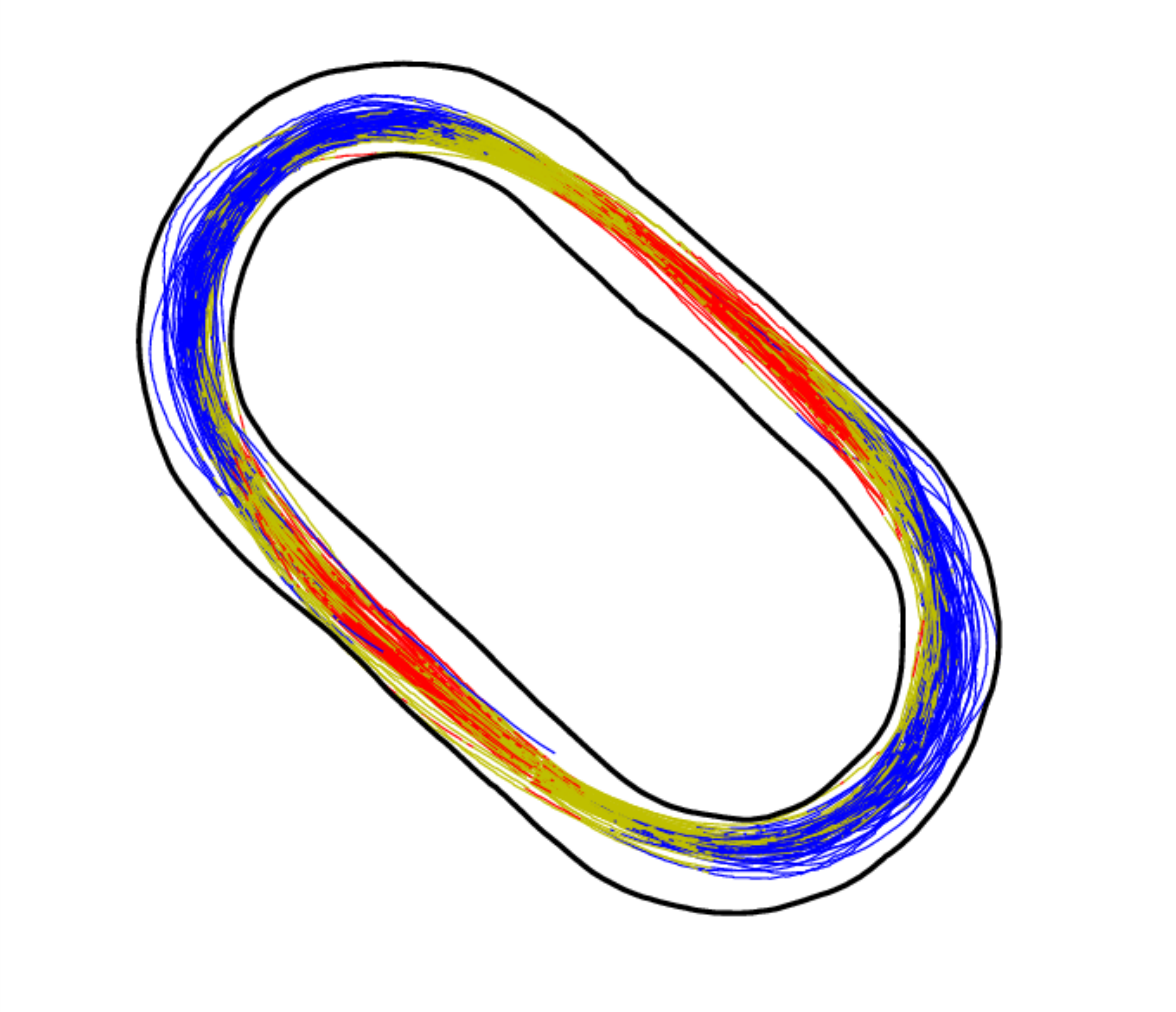} &
\includegraphics[width=.2\textwidth]{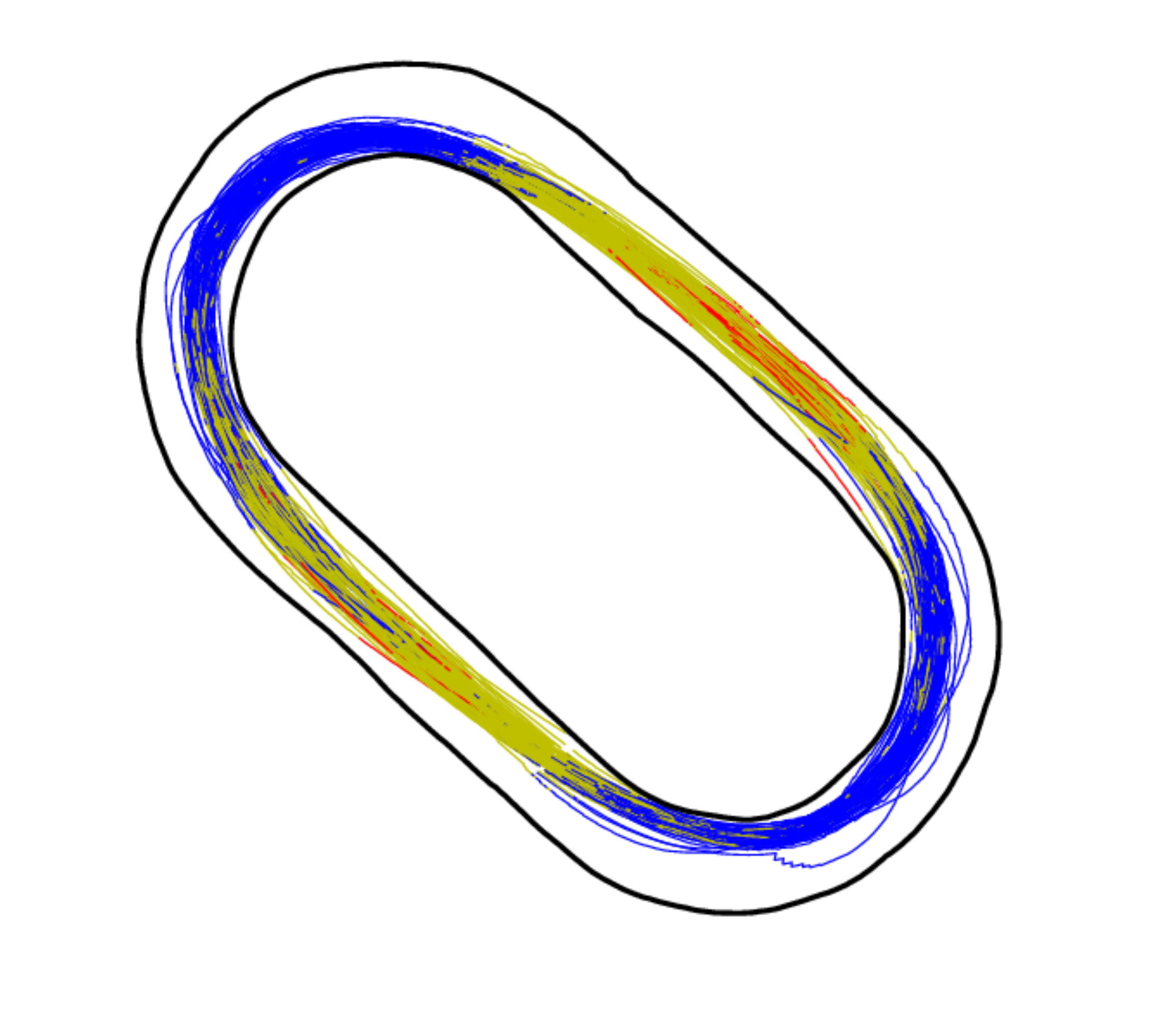} & \\
C-8.5ms & 
\includegraphics[width=.2\textwidth]{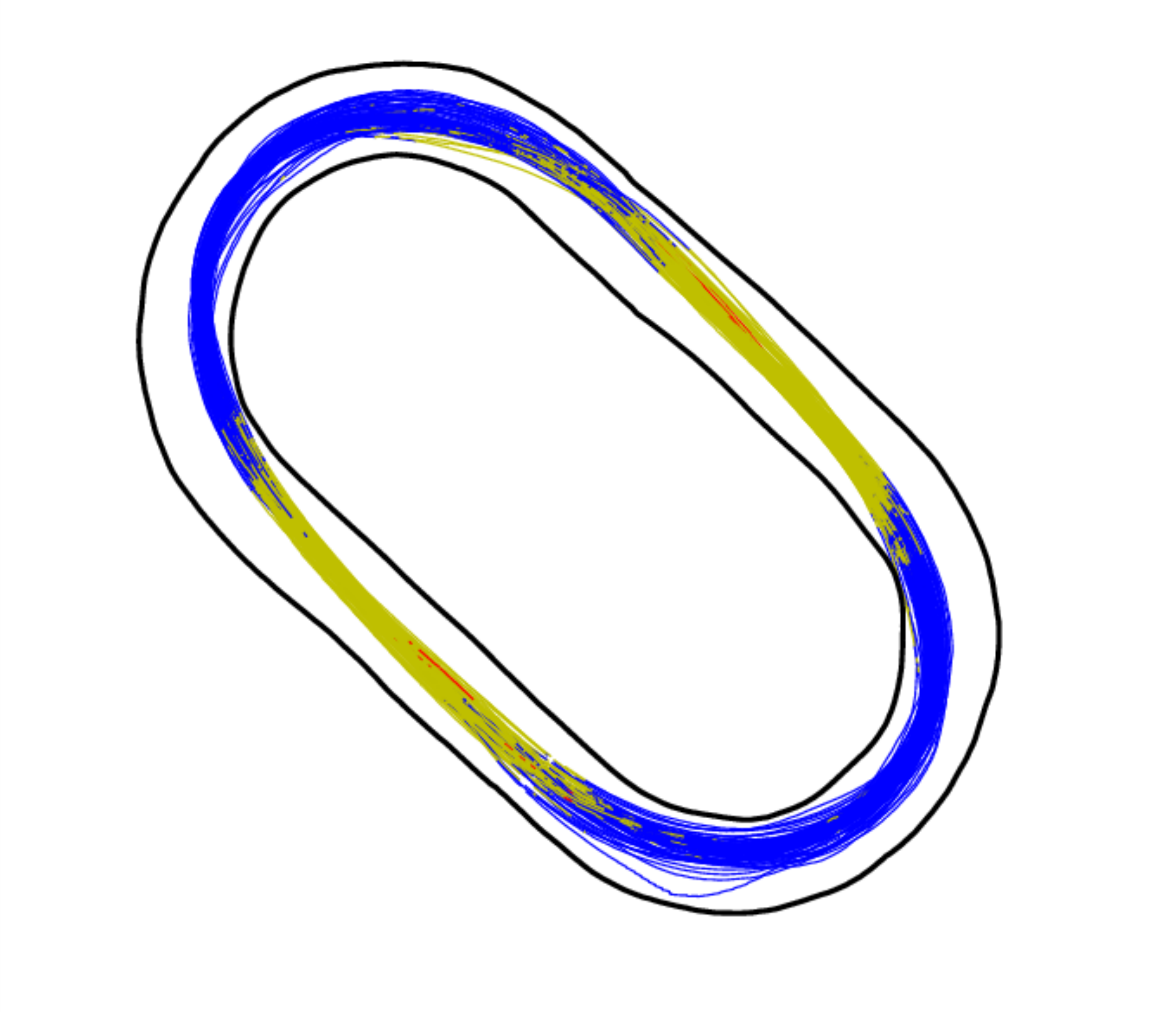} & 
\includegraphics[width=.2\textwidth]{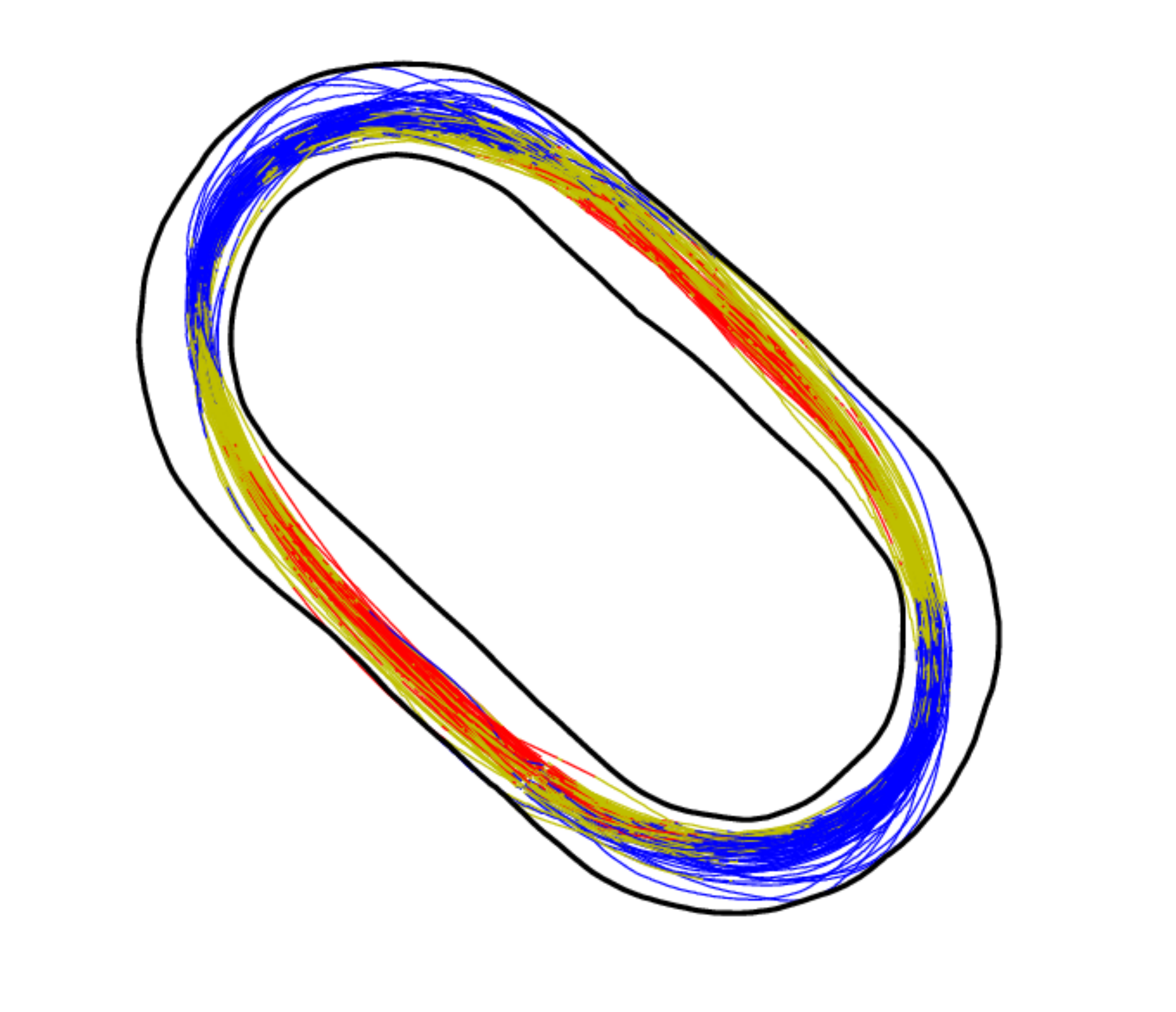} &
\includegraphics[width=.2\textwidth]{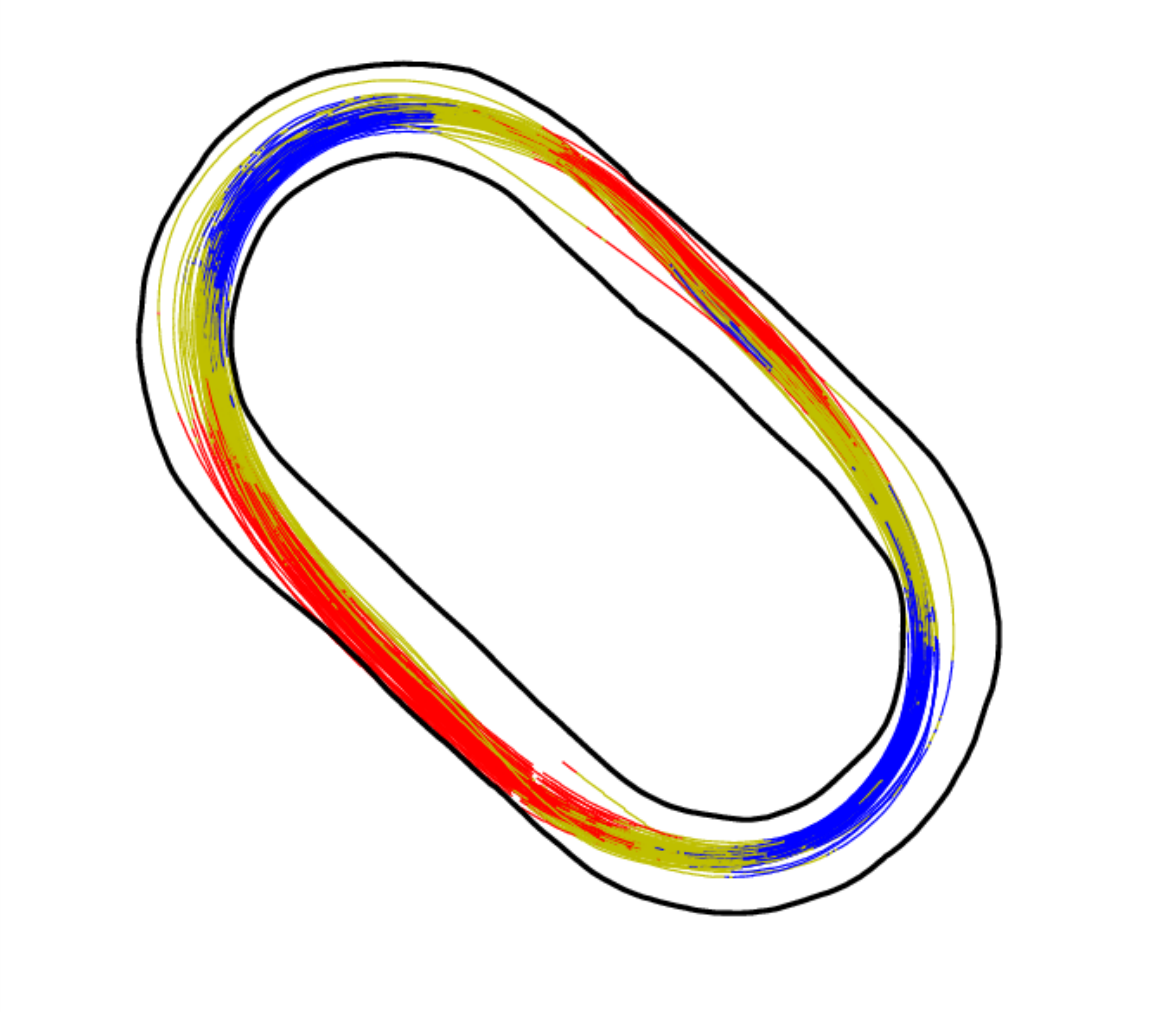} &  \\ \hline 
 & & & \\
CC 11ms & 
\includegraphics[width=.2\textwidth]{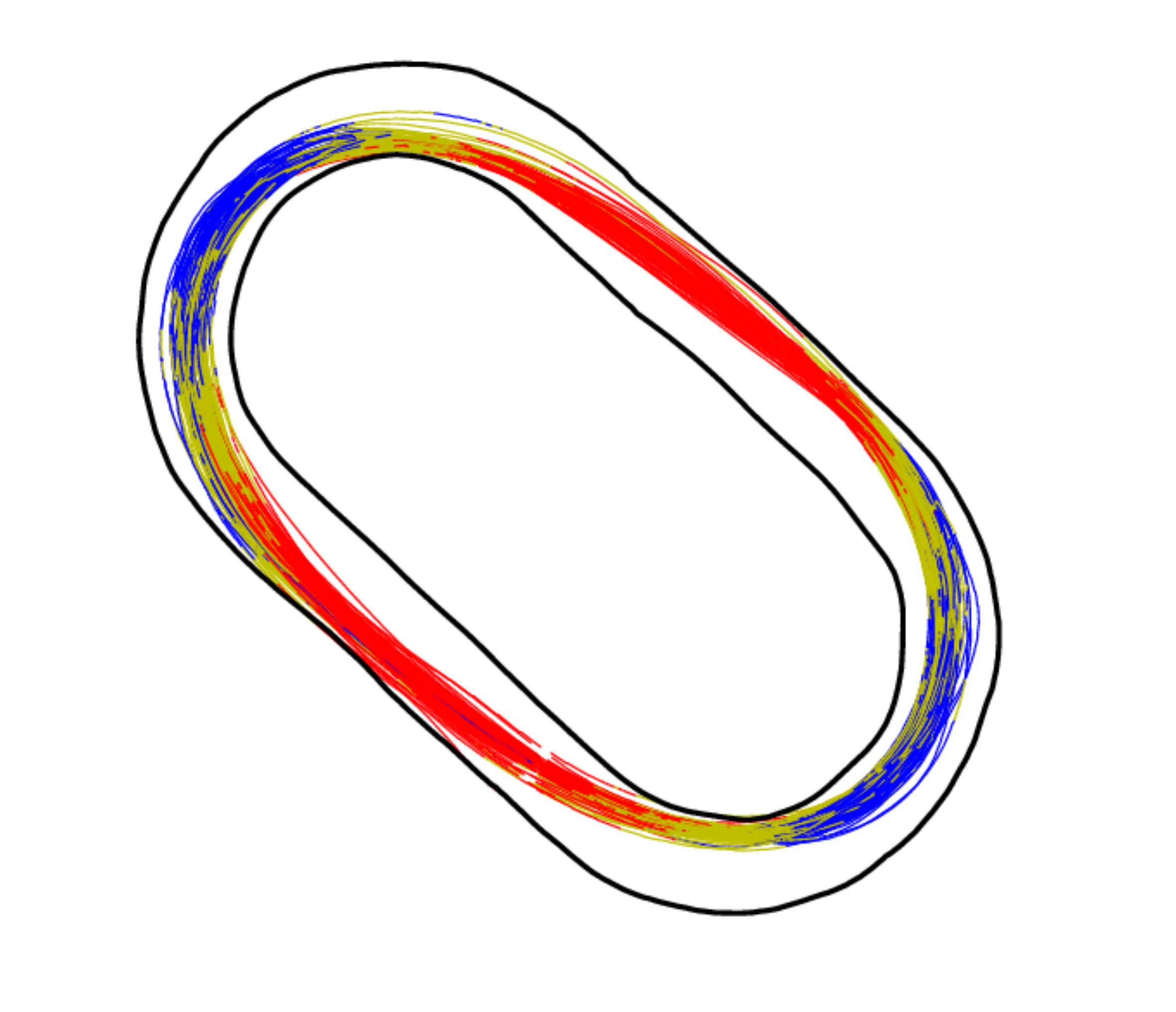} & 
\includegraphics[width=.2\textwidth]{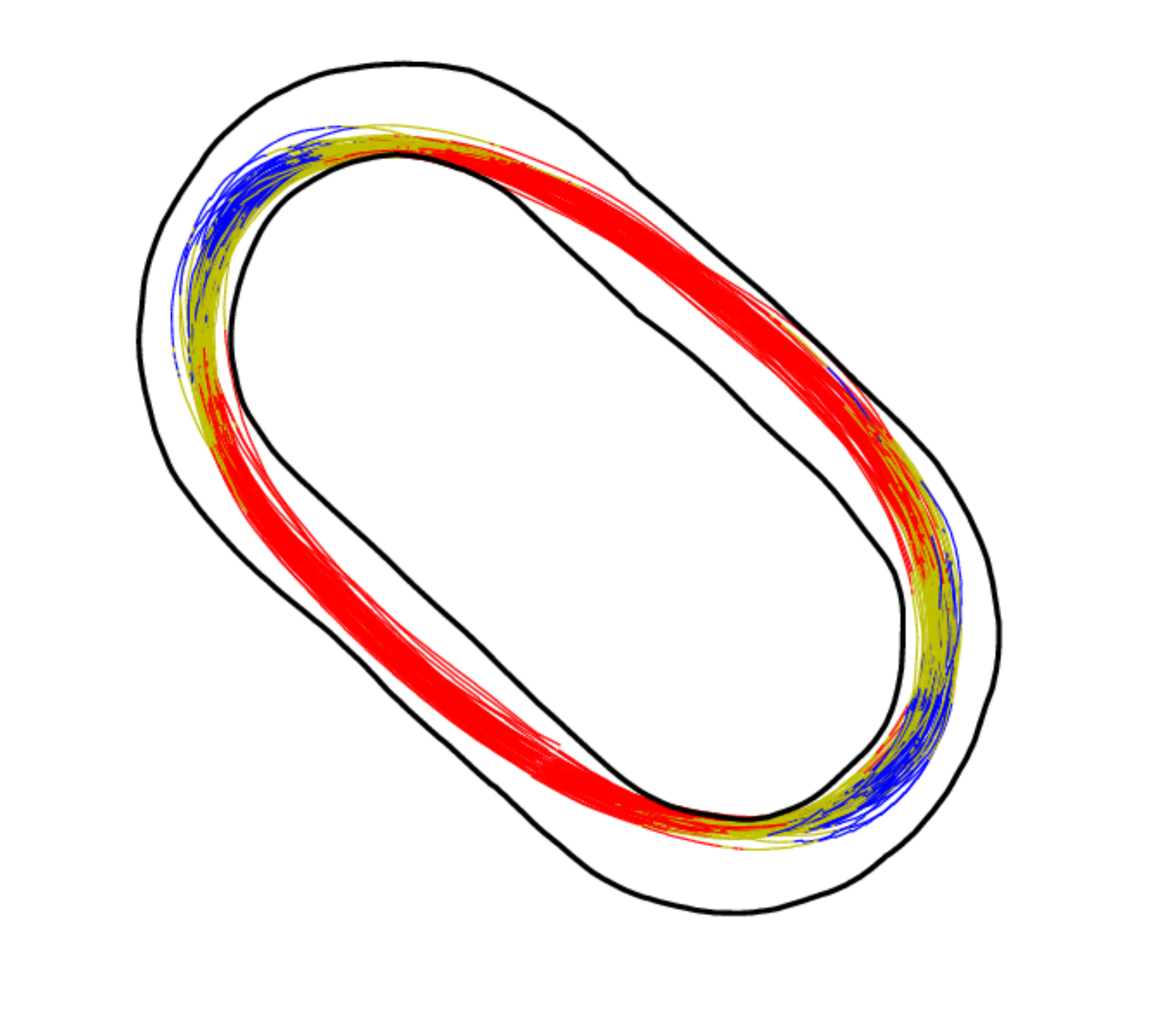} &   \\
C 11ms & 
\includegraphics[width=.2\textwidth]{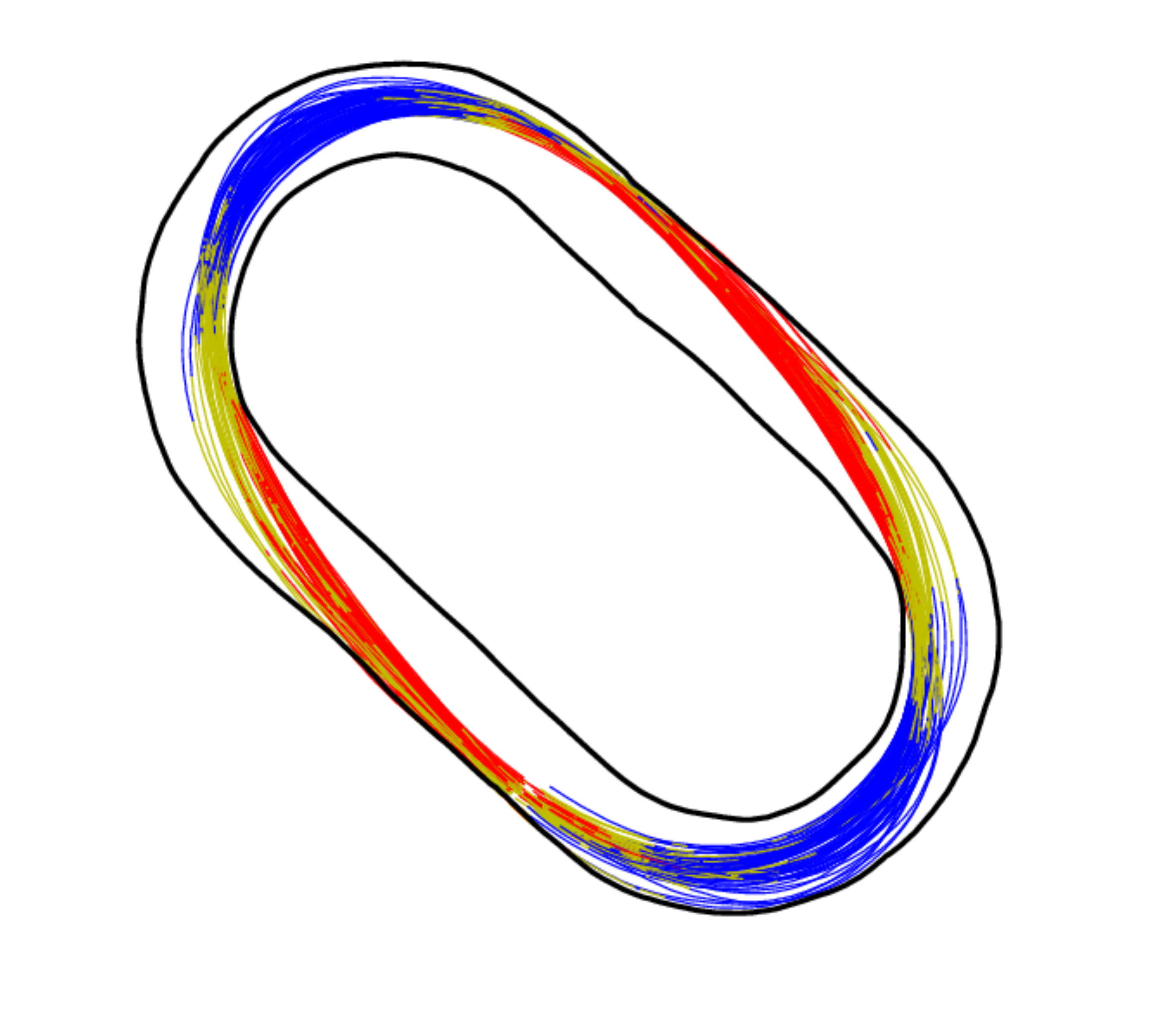} & &
\includegraphics[width=.33\textwidth]{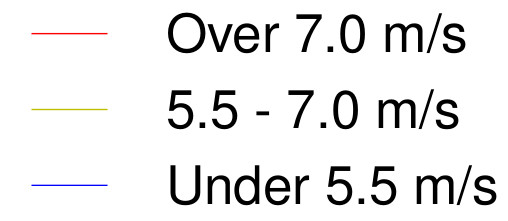}
 \\  \hline \\
\end{tabular}
\label{Table:TrajTraces}
\end{table*}

\subsubsection{6 m/s target} 
\begin{figure}
\centering
\includegraphics[width=.99\columnwidth]{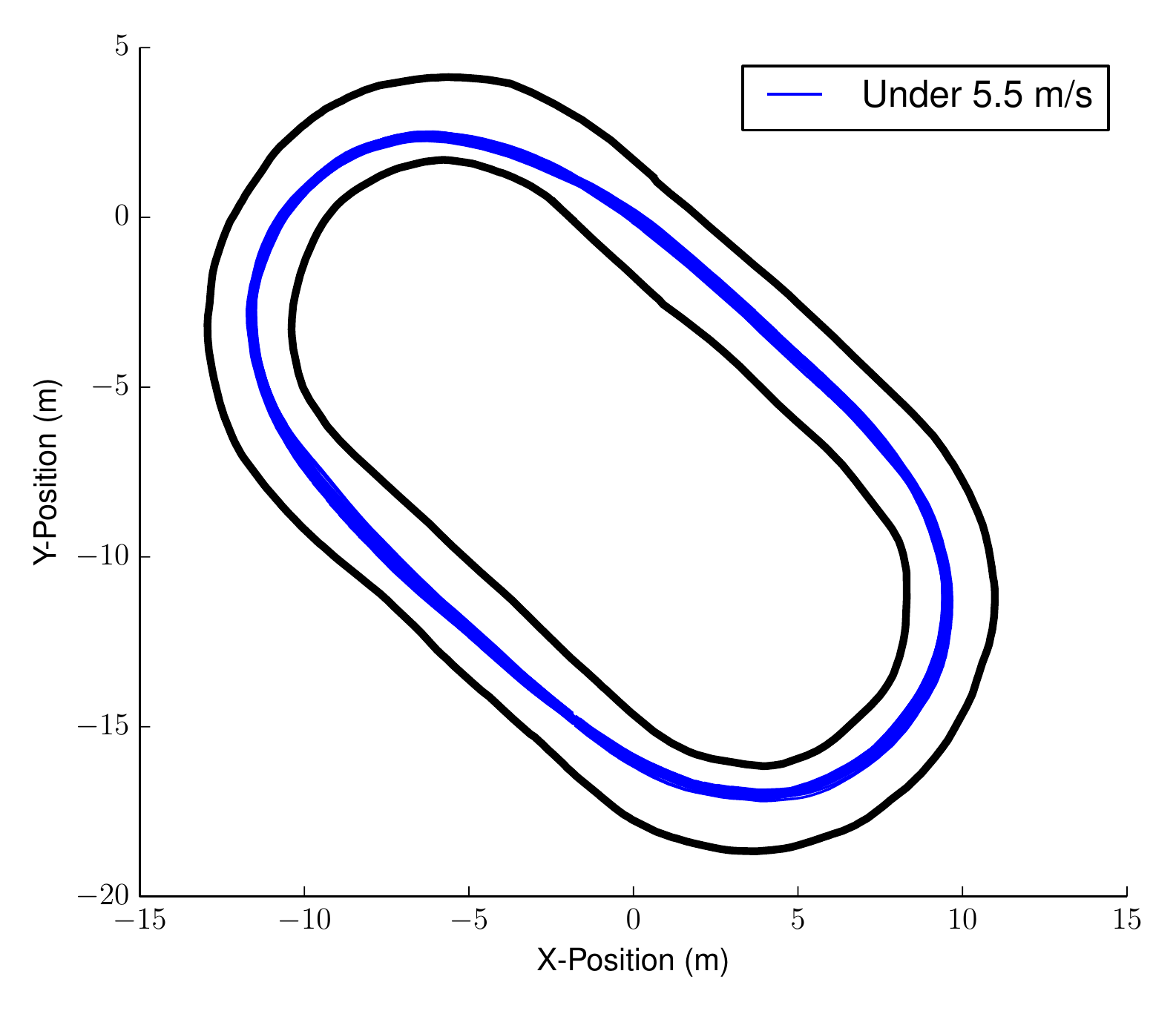}
\includegraphics[width=.99\columnwidth]{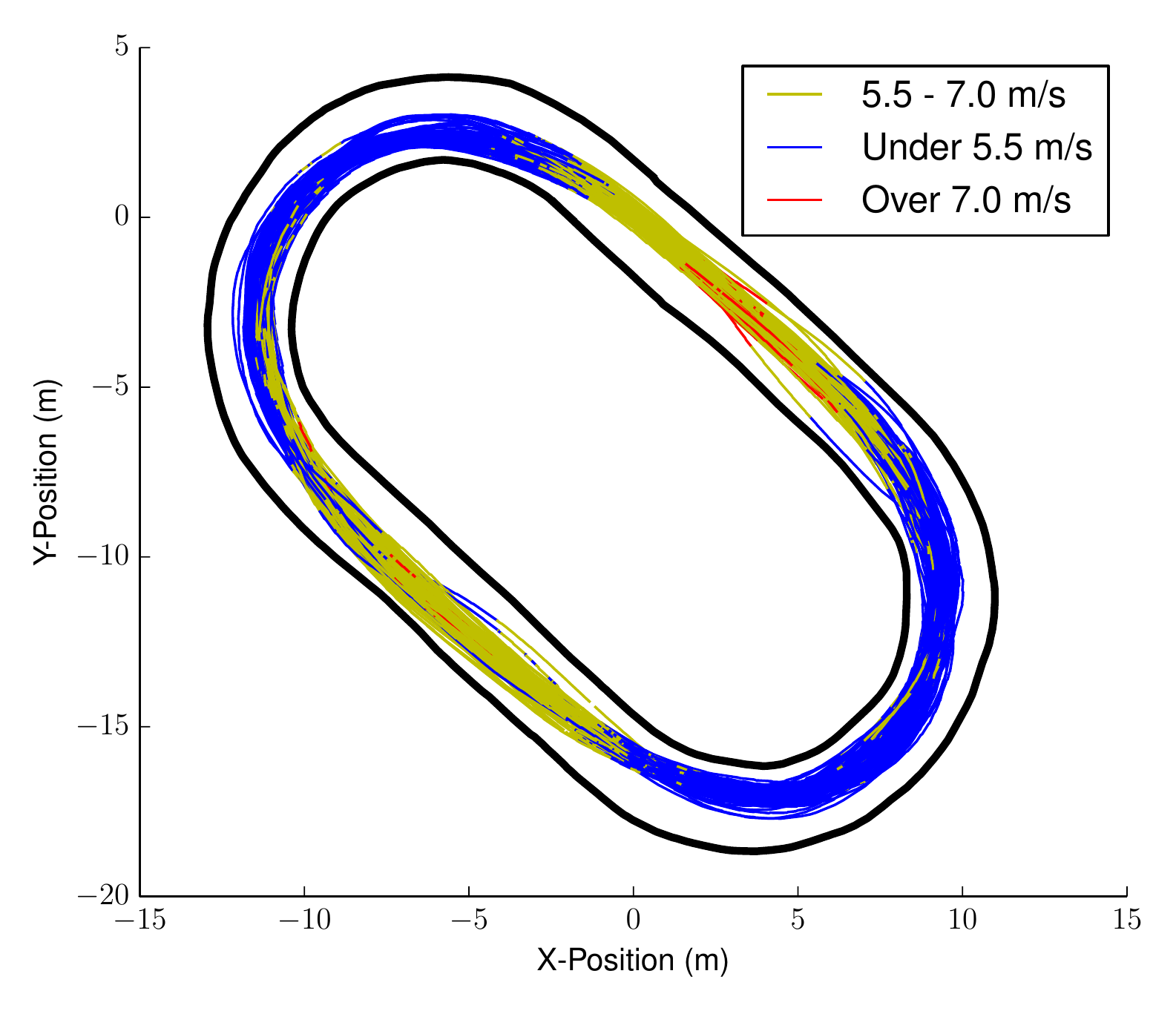}
\includegraphics[width=.99\columnwidth]{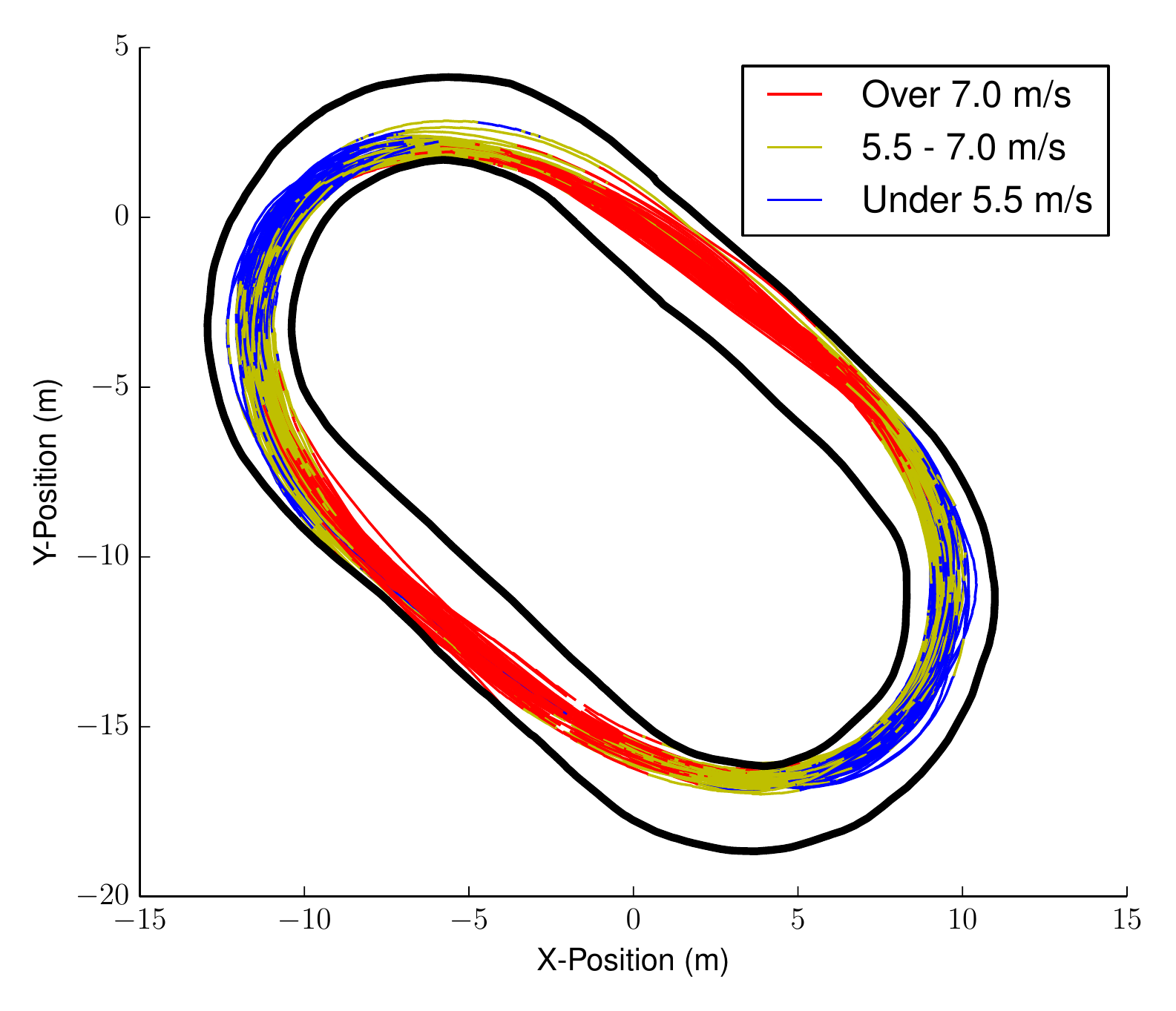}
\caption{Top: Trajectory traces of IT-MPC controller using the neural network model at the 6 m/s (Top), 8.5 m/s (Middle), and 11 m/s (Bottom) target velocity. Each figure represents 100 laps, or approximately 6 kilometers of driving. Direction of travel is counter-clockwise.}
\label{Fig:low_var}
\end{figure}

At the 6 meter per second target, the IT-MPC controllers all perform very consistently, albeit conservatively. Using both the basis function and neural network model the controller navigates the vehicles around the track at speeds varying from just over 1 m/s to a maximum of 5.77 m/s. This keeps the vehicle below the friction limits of the track and vehicle system, which means the car does not slide. The performance of IT-MPC with the neural network is remarkably consistent, especially from a stochastic controller, as the 100 laps in both counter-clockwise and clockwise have extremely low variance from lap to lap. Figure \ref{Fig:low_var} shows the 100 laps collected at the 6 m/s target with the IT-MPC algorithm and neural network model traveling counter-clockwise. Despite the fact that the control actions are generated on-the-fly using a stochastic algorithm, the resulting trajectories are smooth and consistent over the entire trial. 

The cross-entropy method using the neural network model performs perfectly at this settings as well, and actually achieves significantly faster speeds than the IT-MPC algorithm. However, the cross-entropy method cannot be as discriminative as the IT-MPC controller, since IT-MPC can discard, by assigning a low weight, any trajectories that leave the track. In contrast, the cross-entropy method must accept the top 20\% of trajectories into its solution. Even at the slow setting of 6 m/s the cross-entropy method has a failure with the basis function model, and only achieves an 83.16\% (79/95 successful laps) success rate going clockwise around the track using the basis function model. Additionally, the variance of the cross-entropy method at this setting is much higher than the variance of the IT-MPC controller. The trajectory traces for each of the different settings at the 6 m/s target are displayed in the first two rows of Table \ref{Table:TrajTraces}.

\subsubsection{8.5 m/s target} At the 8.5 meter per second target, differences between the algorithms and models become more apparent. IT-MPC using the neural network model is the only method which performs flawlessly going both clockwise and counter-clockwise at this setting. IT-MPC is still more cautious than the cross-entropy method, and achieves maximum speeds about 1 m/s slower than the target velocity. This is consistent with the performance at the 6 m/s target. The speed ranges also start to become dramatic at this setting, for instance IT-MPC with the neural network model (in the counter-clockwise direction) had speed ranges between 1.84 m/s and 7.5 m/s during the approximately 100 laps collected at that setting. 
The 1.84 m/s speed at this setting was not typical, but was the result of the vehicle encountering a large disturbance (due to a bump in the track), and it demonstrates the controller's ability to make drastic mode shifts in order to react to disturbances. Also, note that the IT-MPC controller no longer maintains the extremely tight variance that it did at the 6 m/s target, as the speed cost at 8.5 m/s reduces the relative importance of staying near the center of the track.
%
%

The cross-entropy method has significant difficult at this setting. At the 8.5 m/s setting using the basis function model, the algorithm was unable to complete the trials at a satisfactory rate, and was generally unsafe to run. The issue was that it disregarded the track boundaries, and collided with either the inside or outside track barrier on over 50\% of the trials. The cross-entropy method still maintained a high success rate using the neural network model. 

\subsubsection{11 m/s target} At the fast speed target of 11 m/s only the IT-MPC controller traveling in the counter-clockwise direction is able to complete all 100 laps without a significant violation of the track boundaries. The top speed achieved by IT-MPC at this setting is 9.06 m/s, or approximately 20 miles per hour. The cross-entropy method is still faster than IT-MPC, however the success rate of the algorithm for actually completing laps is very low (only 66.32\%). Figure \ref{Fig:low_var} shows the trajectory traces for the IT-MPC controller with the neural network traveling counter-clockwise. The trajectory traces come extremely close to the barrier, but do not collide with it indicating that the impulse term in the cost function is able to enforce behavior which avoids a collision. 

\begin{figure}
\centering
\includegraphics[width=.99\columnwidth]{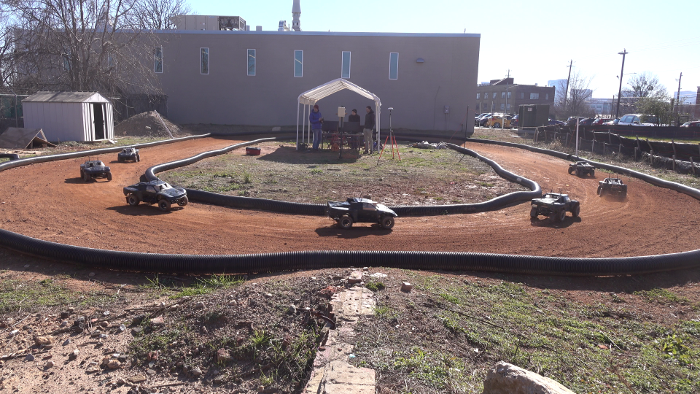}
\caption{Time-lapse image of the vehicle making a cornering maneuver. Notice how the front left wheel is off the ground as the vehicle enters the turn.}
\label{Fig:CornerTimeLapse}
\end{figure}

\subsection{Cornering Maneuvers}

\begin{figure}
\centering
\includegraphics[width=.99\columnwidth]{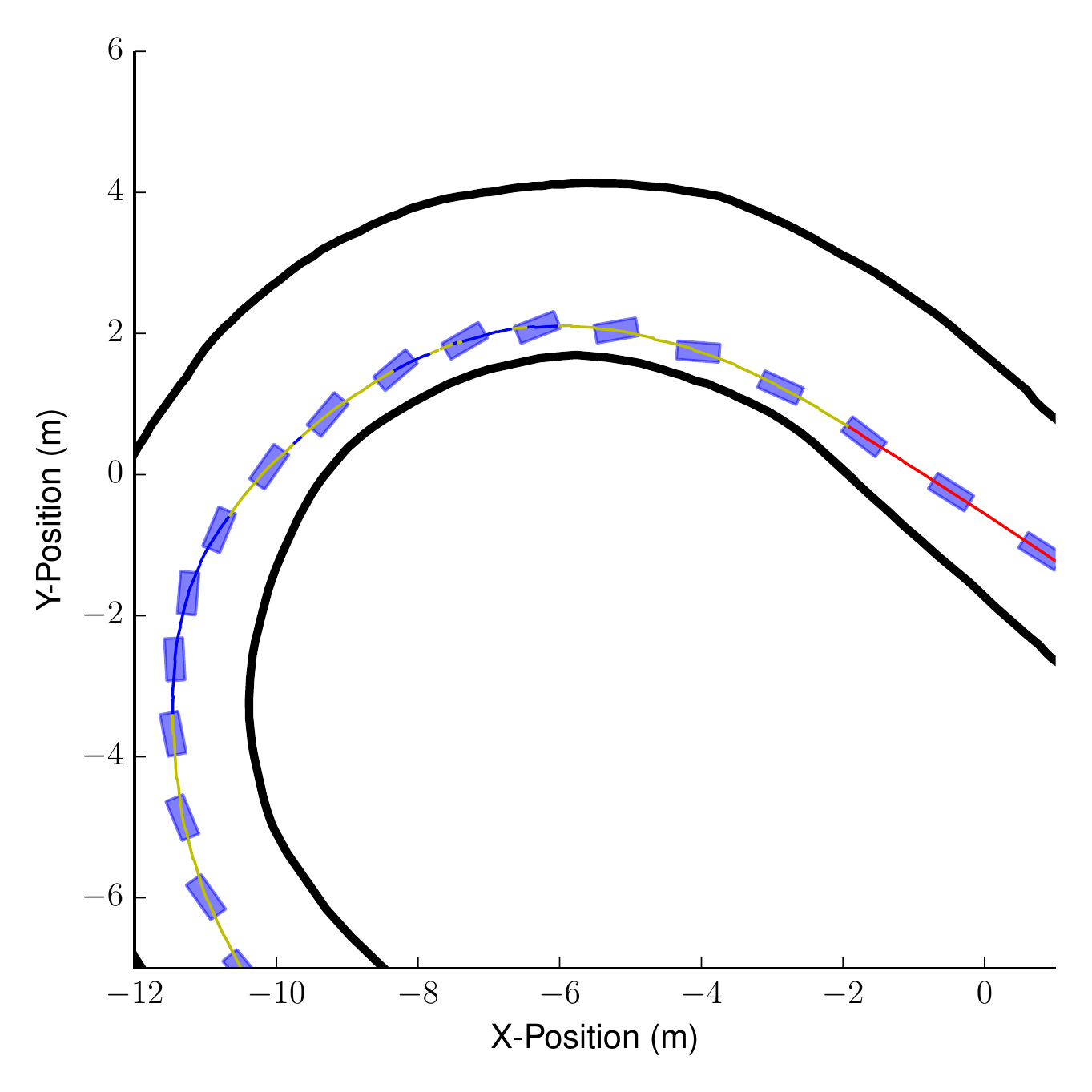}
\caption{Trajectory and heading trace of cornering maneuver. The direction of travel is counter-clockwise, heading indicator is not to scale.}
\label{Fig:OverheadCornerTimeLapse}
\end{figure}

The most difficult part of aggressive driving, from a control perspective, is cornering. Successful cornering requires significantly reducing speed, and then applying the throttle as the vehicle exits the turn. Failing to reduce speed or applying the throttle too soon can result in spin-outs (uncontrolled high heading rate). Using the neural network model, the IT-MPC controller decreases speed by performing a small slide into turns. This is a delicate maneuver that often results in the left front wheel momentarily lifting off the ground. Once the vehicle straightens out, the controller hits the throttle and resumes sliding slightly as it exits the turn and enters the straight. Figure \ref{Fig:CornerTimeLapse} shows a time lapse video of the vehicle entering the turn at the 11 m/s target, and fig. \ref{Fig:OverheadCornerTimeLapse} shows the same maneuver from an overhead perspective.

Another common behavior of the controller is counter-steering (steering right to turn left) when exiting turns. This is a behavior which requires taking advantage of the non-linear dynamics of vehicle, and is only effective at high speeds. Figure \ref{Fig:powerslide} shows this behavior as the car exits a turn on one of the 11 m/s trials.

\subsection{Robustness to Model Error}
%
%

\begin{figure}
\centering
\includegraphics[width=.45\columnwidth]{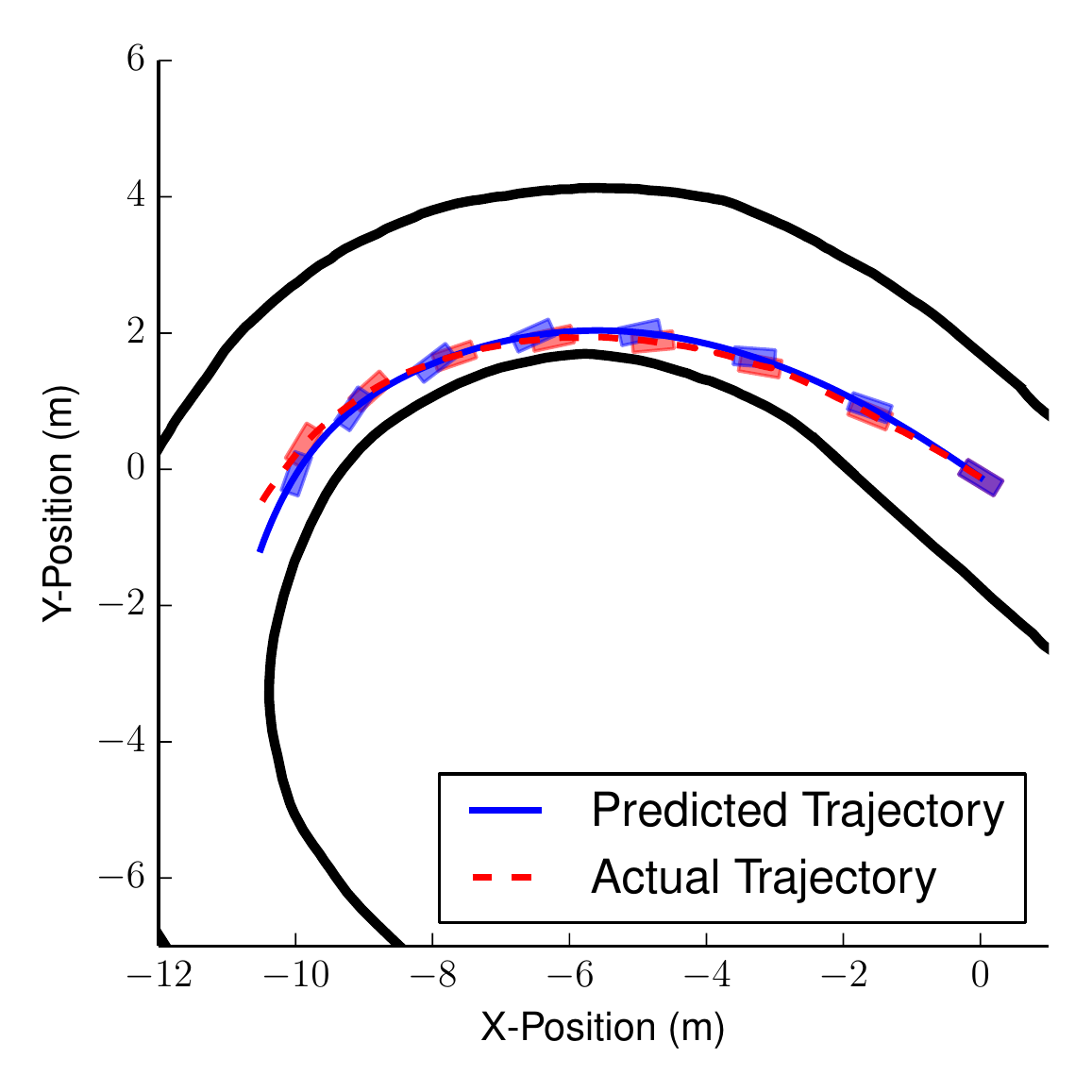}
\includegraphics[width=.45\columnwidth]{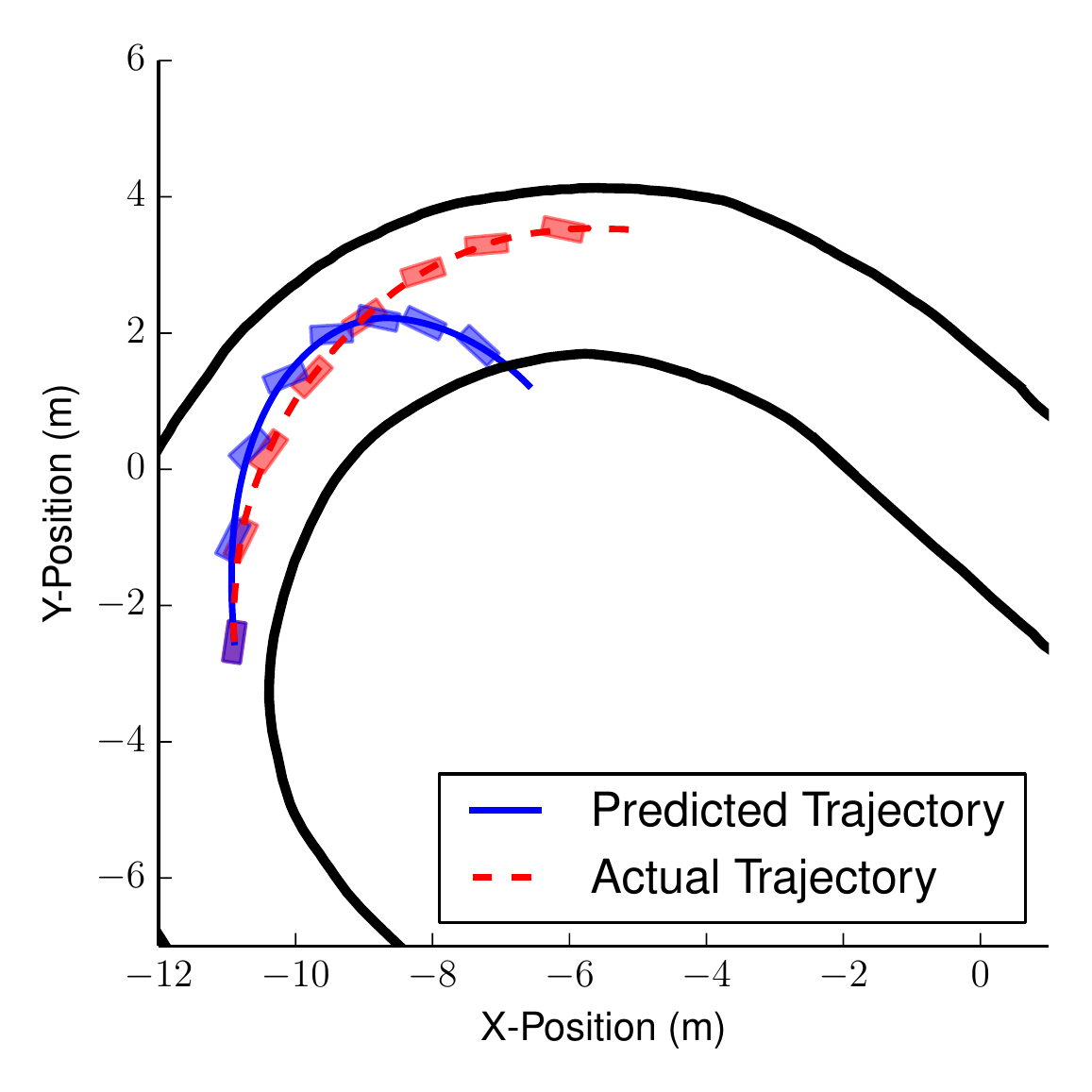}
\caption{Neural network modeling error at 11 m/s target. Going counter-clockwise the model
prediction is accurate, but clockwise the model predicts severe over-steer when it should have predicted under-steer. The predicted trajectory is generated by taking the applied
input sequence from the data recording and running it through the neural net model starting from the same initial condition.}
\label{Fig:ModelError}
\end{figure}

In order to navigate the vehicle around the track, the controller has to be robust to modeling error (Table \ref{Table:DynamicsPerformance}). Figure \ref{Fig:ModelError} shows how the predicted model differs from reality around a typical turn at the 11 m/s target with the neural network model. Going counter-clockwise the model is able to accurately predict out to the 2 second time horizon. However, in the clockwise direction the model incorrectly predicts over-steer when in fact the vehicle under-steers. 

This behavior is likely due to asymmetry in the training data. Even though the system identification data is collected in a symmetric and choreographed manner, the human pilot reacts in slightly different ways going clockwise and counter-clockwise. This is especially the case when agile and high speed maneuvers are generated by the human pilot. As a consequence, the dynamics are better identified going counter-clockwise, which results in a difference in performance when pushing the vehicle to its limits. Despite the large error in the clockwise direction, the vehicle is still able to successfully complete the task close to $80\%$ of the time, and achieves a top speed of over 9 meters per second.
%
%

\subsection{Disturbance Rejection}

\begin{figure*}
\centering
\includegraphics[width=.245\textwidth]{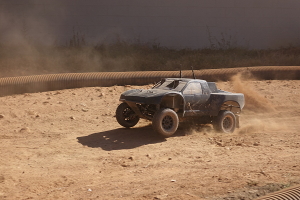}
\includegraphics[width=.245\textwidth]{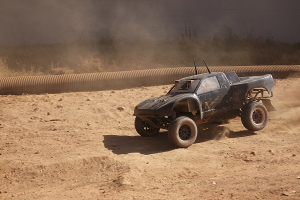}
\includegraphics[width=.245\textwidth]{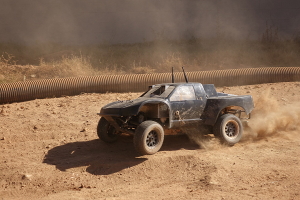}
\includegraphics[width=.245\textwidth]{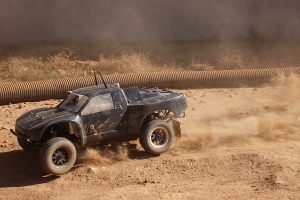}
\caption{Disturbance rejection by the IT-MPC controller. The car hits a large hole in the track and the front and rear wheels leave the ground in alternating fashion while the vehicle is attempting to steer around the corner.}
\label{Fig:Disturbances}
\end{figure*}

In addition to systemic modeling error, the dirt track provides a source of strong disturbances which cannot be modeled using our state representation. This includes environment effects like holes and lose patches of dirt. This became especially difficult during the 8.5 m/s test runs when dry weather\footnote{The Atlanta area experienced a severe drought in late 2016 which made it impossible to compact the dirt in order to repair the track. The drought relented soon after the completion of the 8.5 m/s runs, and was repaired before testing the 11 m/s target.} and hundreds of consecutive laps around the track made it very difficult to drive. Despite these effects, the neural network model with IT-MPC was able to successfully complete all 100 laps. Figure \ref{Fig:Disturbances} shows a series of images which demonstrate the effect of these disturbances on the vehicle. 

\subsection{Failure Modes}

\begin{figure}
\centering
\includegraphics[width=.49\columnwidth]{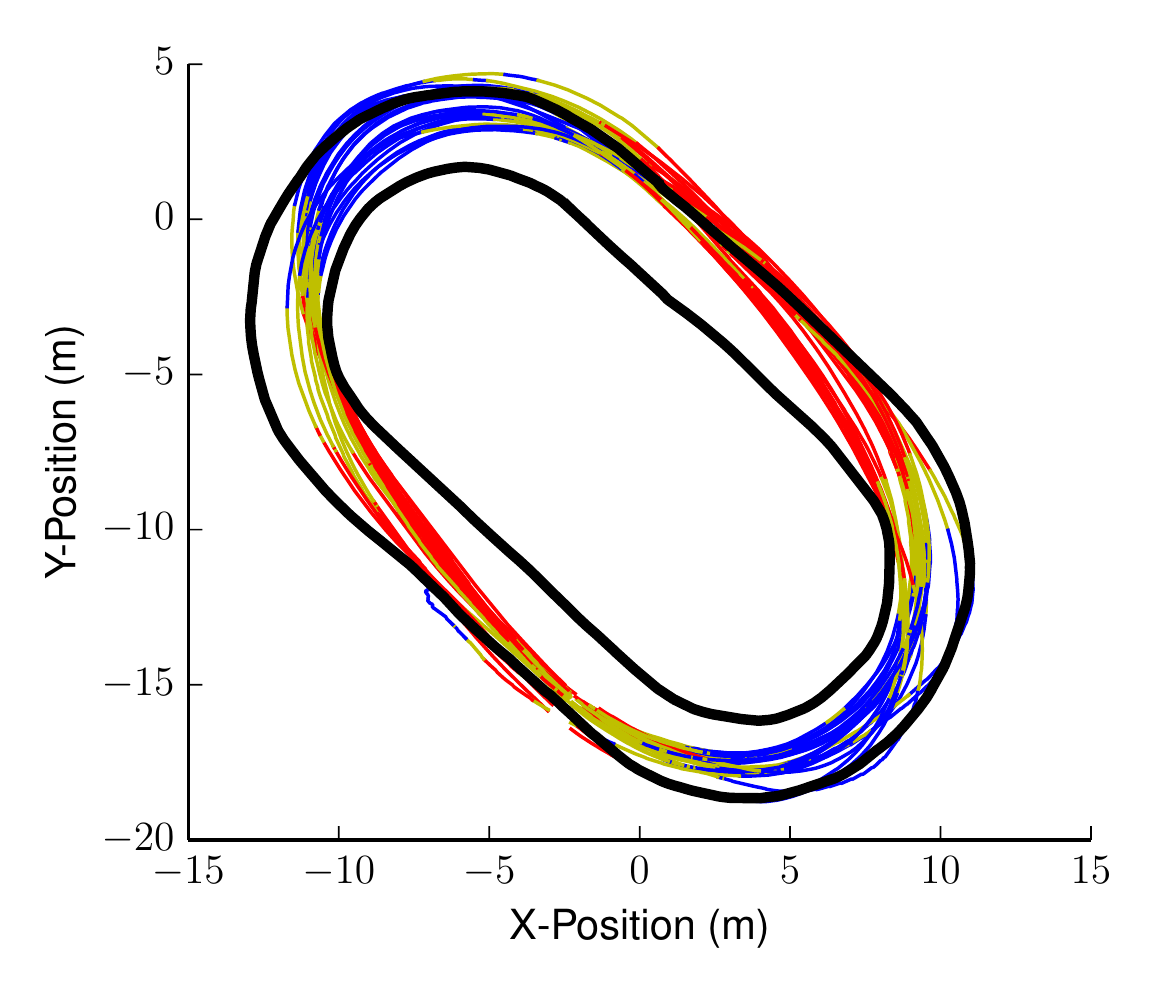}
\includegraphics[width=.49\columnwidth]{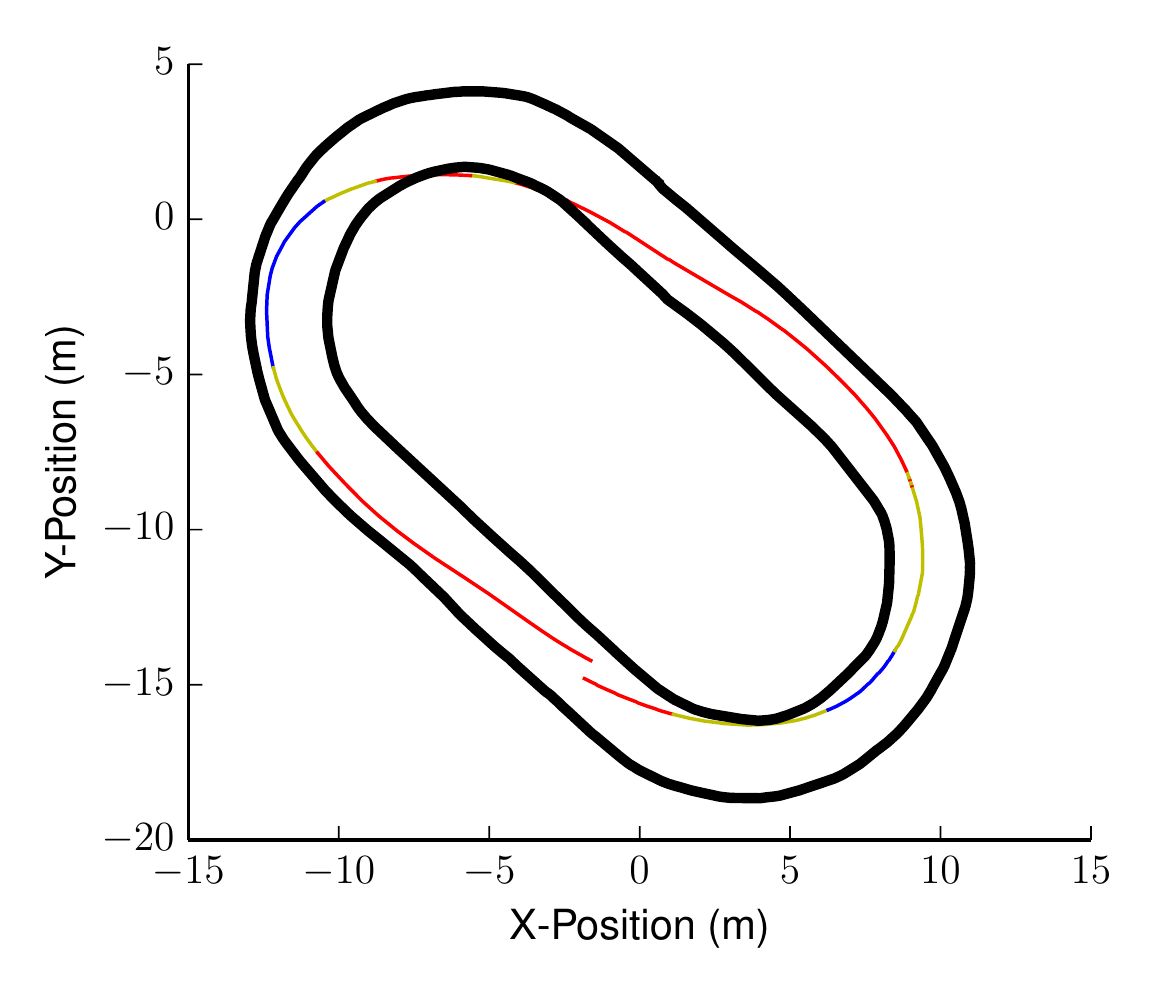}
\caption{Failure mode of the IT-MPC algorithm. Speed setting is 11 m/s and the direction of travel is clockwise. The model under-estimates the amount of steering input required, which results in under-steer and collision with the barrier.}
\label{Figure:mppi_fail}
\end{figure}

Although the neural network model generally outperformed the basis function model, and IT-MPC generally outperformed the CEM-MPC in terms of success rate. All of the methods suffered some failures. With IT-MPC and the neural network, the only failures came from attempting to navigate the track clockwise at the 11 m/s target. The problem in this case was systematic modeling error which caused the vehicle to under-steer around the corners. Figure \ref{Figure:mppi_fail}  shows all of the trajectories generated by the IT-MPC controller at this setting which failed. Notice that all of the trajectories fail in a similar manner, note that the track boundaries were pushed out a little bit so that we could continue collecting data even when the vehicle violated the boundary.
In the case of cross-entropy, the failure come from not respecting the track boundary. Even when going clockwise with the neural network model, which is very accurate, the cross-entropy method consistently violates the track boundary. This is due to the sampling method used by cross-entropy, which allows trajectories into the sampling even if they violate the track constraint.

\section{Discussion}
In this paper we have derived an information theoretic framework which provides the mathematical tools required to design sampling based optimization algorithms suited for controlling autonomous systems. We compared and contrasted the theoretical aspects of this new framework with traditional stochastic optimal control, and we demonstrated how the new framework can be used to derive a sampling based model predictive controller.

We applied this new model predictive control algorithm on an autonomous driving system, and showed that it was capable of consistent, smooth driving at low to medium speeds, as well as performing high speed maneuvers when the desired target speed is set high above the friction limits of the vehicle. Unlike the current approaches to autonomous driving, which split the control problem into planning and execution steps, our approach simultaneously plans high level behaviors through sampling directly in control space. This approach is made possible by massive parallel sampling on a GPU. 

Our experiments demonstrate that the costs and dynamics associated with the autonomous driving problem are well suited for a sampling based control scheme: the method naturally handles the non-linear dynamics and it is possible to use large impulse terms in the cost function to provide a strong incentive to avoid the track boundary, while still treating track boundary collisions as a soft constraint. This enables the vehicle to steer out of collision when it does contact the barrier. This approach compares favorably with the a model predictive control version of the cross-entropy method, which, although it is able to handle the non-linearity of the dynamics and plan aggressive trajectories, is unable to finely discriminate between trajectories which do and do not contact the barrier. This leads to a lower overall success rate than IT-MPC.

The type of approach that we have demonstrated is a promising new direction for solving the challenging problems that arise in autonomous driving tasks. The key tools in this approach are the information theoretic concepts of free energy and the KL divergence, and intensive parallel computation for online optimization.


%

\appendices

\section{Basis Function Model}\label{appendix:bfm}
We used an analytic model of vehicle dynamics derived in \cite{hindiyeh2013dynamics} as a source of physics based knowledge about vehicle dynamics. This model makes the simplifying assumption that the two front tires and two back tires are lumped into one tire at the front and the back (and is therefore known as a bicycle vehicle model). This makes the incorporation of lateral tire forces into the model tractable. The full state equations are for the model are given below:
\begin{align*}
\dot{x} &= v_x \cos(\psi) - v_y \sin(\psi),~~  \dot{y} = v_x \sin(\psi) + v_y \cos(\psi) \\
\dot{\psi} &= r, ~~\dot{\beta} = \frac{F_{yF} + F_{yR} }{M v_x} - r \\ 
\dot{v_x} &= \frac{u_F - F_{yF} \sin(u_\delta)}{M} + r v_x \beta, ~~\dot{v_y} = \frac{F_{yF} + F_{yR} }{M} - r v_x \\
\dot{r} &= \frac{aF_{yF} - bF_{yR} }{I_z}. 
\end{align*}
Here $(x,y)$ is position, $\psi$ is the heading, $\beta$ is the side-slip angle, $v_x, v_y$ are longitudinal and lateral velocity in the body frame of the vehicle, $r$ is the heading (yaw) rate, $u_\delta$ is the steering angle, and $u_F$ is the longitudinal force imparted by the rear wheels. The inputs to the model are $u_\delta$ and $u_F$. The terms $(a,b)$ are the distances from the center of mass to the front and rear axles. The terms $F_{yF}$ and $F_{yR}$ are the lateral tire forces imparted by the front and rear wheels respectively. This force is a function of the slip angle $\alpha$ which we compute based on a brush tire model.
\begin{align*}
\alpha_F &= \tan^{-1}(\beta + a\frac{r}{v_x}) - u_\delta, ~ \alpha_R = \tan^{-1}(\beta - b\frac{r}{v_x})
\end{align*}
and then the lateral forces are:
\begin{align}
F_{yi} &= \begin{cases} 
-\mu \xi F_z \frac{|\alpha_i|}{\alpha_i} \quad \text{if} \quad \alpha_i \ge \gamma_i \\
\zeta \quad \text{otherwise}\\
\end{cases} \\
\xi &= \left( \frac{\mu^2 F_z^2 - u_F^2}{\mu F_z}\right)^{1/2} \\
\gamma &= \left| \tan^{-1} \left( 3 \xi F_z \frac{|\alpha|}{\alpha} \right) \right| \\
\zeta &= -C \tan(\alpha_i) + \frac{C^2}{3 \xi \mu F_z} \frac{\tan(\alpha_i)^3}{|\tan(\alpha_i)|} - \frac{C^{3}}{27\mu^2 \xi^2 F_z^2}\tan(\alpha_i)^3.
\end{align}
Here $C$ is the cornering stiffness of the tire, and $\mu$ is the co-efficient of friction between the tire and the ground. Based on these equations we picked out the key non-linearities found in the previous model and used them to form 25 basis function. The equations of motion are then:
\begin{equation}
\label{Equation:EqMot}
\dot{\vx} = \theta^\rT \Phi(\vx),
\end{equation}
In the following we define:
\begin{equation*}
\alpha_f = 
\begin{cases}
\arctan \left( \frac{v_y}{v_x} + .45\frac{r}{v_x} - u_\delta \right) \quad &\text{if } v_x > 0.1 \\
-u_\delta \quad &\text{otherwise}
\end{cases}
\end{equation*}
\begin{equation*}
\alpha_r = 
\begin{cases}
\arctan \left( \frac{v_y}{v_x} - .35\frac{r}{v_x} \right) \quad &\text{if } v_x > 0.1 \\
0 \quad &\text{otherwise}
\end{cases}
\end{equation*}
The basis functions that we choose for the AutoRally model are then:
\begin{align*}
\phi_1 &= u_F,~ \phi_2 = v_x/10, ~\phi_3 = \sin(u_\delta)\tan(\alpha_f)/1200 \\
\phi_4 &= \sin(u_\delta)\tan(\alpha_f)\|\tan(\alpha_f)\|/1200^2 \\
\phi_5 &= \sin(u_\delta)\tan(\alpha_f)^3/1200^3 \\
\phi_6 &= r v_y / 25, ~\phi_7 = r/10,~ \phi_8 = v_y/10, ~\phi_9 = \sin(u_\delta) \\
\phi_{10} &= 
\begin{cases}
\frac{v_y}{v_x}/40 \quad &\text{if } v_x > 0.1 \\
0 &\text{otherwise} \\
\end{cases}
\\
\phi_{11} &= \tan(\alpha_f)/1400, ~\phi_{12} = \tan(\alpha_f)\|\tan(\alpha_f)\|/1400^2 \\
\phi_{13} &= \tan(\alpha_f)^3 / 1400^3,~\phi_{14} = \tan(\alpha_r)/40 \\
\phi_{15} &= \tan(\alpha_r)\|\tan(\alpha_r) \| / 40^2, ~\phi_{16} = \tan(\alpha_r)^3 / 40^3 \\
\phi_{17} &= r v_x / 50,~ \phi_{18} = \theta \\
\phi_{19} &= \theta r, ~\phi_{20} = \theta v_x / 3,~ \phi_{21} = \theta v_x r / 5, ~ \phi_{22} = v_x^2 / 100 \\
\phi_{23} &= v_x^3 / 1000, ~\phi_{24} = u_F^2, ~\phi_{25} = u_F^3.
\end{align*}

\section*{Acknowledgment}
This work was made possible by the ARO through MURI award W911NF-11-1-0046,  DURIP award W911NF-12-1-0377, NSF award NRI-1426945

\ifCLASSOPTIONcaptionsoff
  \newpage
\fi




\bibliographystyle{IEEEtran}
\bibliography{references}

%
%

%

\newpage

\begin{IEEEbiographynophoto}{Grady Williams}
Grady Williams received a B.S in mathematics from the University of Washington in Winter 2014, and is currently working towards a PhD in Robotics at the Georgia Institute of Technology in the Autonomous Control and Decisions Systems Laboratory. His research interests lie at the intersection of machine learning and control with applications to fast autonomous navigation.
\end{IEEEbiographynophoto}
\begin{IEEEbiographynophoto}{Paul Drews}
Paul Drews received a B.S. in Electrical and Computer Engineering from the Missouri University of Science and Technology in Fall 2008, and is currently working towards a PhD in Robotics a the Georgia Institute of Technology in the Wall Lab.  His research interests lie in the tight coupling of neural networks and computer vision to control problems, specifically applied to fast autonomous navigation.
\end{IEEEbiographynophoto}
\begin{IEEEbiographynophoto}{Brian Goldfain}
Brian Goldfain received a B.S in Electrical and Computer Engineering from
Carnegie Mellon University in Spring 2010, a M.S. in Computer Science from the Georgia Institute of Technology in 2013, and is currently working towards a PhD in Robotics at the Georgia Institute of Technology in the Computational Perception Laboratory. His research interests lie in creating affordable and robust autonomous vehicle testbeds to decrease development time for safe autonomous vehicles.
\end{IEEEbiographynophoto}
\begin{IEEEbiographynophoto}{James M. Rehg}
James M. Rehg (pronounced "ray") is a Professor in the School of Interactive Computing at the Georgia Institute of Technology, where he is Director of the Center for Behavioral Imaging and co-Director of the Computational Perception Lab (CPL). He is also affiliated with the Institute of Robotics and Intelligent Machines. He received his Ph.D. from CMU in 1995 and worked at the Cambridge Research Lab of DEC (and then Compaq) from 1995-2001, where he managed the computer vision research group. He received an NSF CAREER award in 2001 and a Raytheon Faculty Fellowship from Georgia Tech in 2005. He and his students have received best student paper awards at ICML 2005, BMVC 2010, Mobihealth 2014, and Face and Gesture 2015, and a 2013 Method of the Year Award from the journal Nature Methods. Dr. Rehg serves on the Editorial Board of the Intl. J. of Computer Vision, and he served as the Program co-Chair for ACCV 2012 and General co-Chair for CVPR 2009. His research interests include computer vision, machine learning, robot perception, autonomous vehicles, and mobile health. He has authored more than 100 peer-reviewed scientific papers and holds 25 issued U.S. patents. 
\end{IEEEbiographynophoto}
\begin{IEEEbiographynophoto}{Evangelos A. Theodorou}
Evangelos A. Theodorou is an assistant professor with the Guggenheim School of aerospace engineering at Georgia Institute of Technology. He is also affiliated with the Institute of Robotics and Intelligent Machines. Evangelos Theodorou earned his Diploma in Electronic and Computer Engineering from the Technical University
of Crete (TUC), Greece in 2001. He has also received a MSc in Production Engineering from TUC in 2003, a MSc in Computer Science and Engineering from University of Minnesota in spring of 2007 and a MSc in Electrical
Engineering on dynamics and controls from the University of Southern California(USC) in Spring 2010. In May of 2011 he graduated with his PhD, in Computer Science at USC. After his PhD, he became a Postdoctoral Research Associate with the department of computer science and engineering, University of Washington, Seattle.  Evangelos Theodorou is the recipient of the King-Sun Fu best paper award of the IEEE Transactions on Robotics for the year 2012 and recipient of the best paper award in cognitive robotics in International Conference of Robotics and Automation
2011. He was also the finalist for the best paper award in International Conference of Humanoid Robotics 2010 and International Conference of Robotics and Automation 2017. His theoretical research spans the areas of stochastic optimal  control theory, machine learning, information theory and statistical physics. Applications involve learning, planning and control in autonomous, robotics and aerospace
systems.
\end{IEEEbiographynophoto}





\end{document}